\documentclass[lettersize,journal]{IEEEtran}
\usepackage{amsmath,amsfonts}
\usepackage{algorithmic}
\usepackage{algorithm}
\usepackage{array}
\usepackage[caption=false,font=normalsize,labelfont=sf,textfont=sf]{subfig}
\usepackage{textcomp}
\usepackage{stfloats}
\usepackage{url}
\usepackage{verbatim}
\usepackage{graphicx}
\usepackage{cite}
\usepackage{tabularx}
\usepackage{multirow}
\usepackage{booktabs}
\usepackage{makecell}
\usepackage[shortlabels]{enumitem}
\usepackage{tikz}
\usepackage{pgfplots}
\pgfplotsset{compat=1.16}
\usetikzlibrary{positioning,arrows.meta,fit,backgrounds}
\usepackage[edges]{forest}
\usepackage{xcolor}
\begin{document}
\bstctlcite{IEEEbstctl}

\title{Why Is Video Still So Expensive? A Survey of Inference-Efficiency Mechanisms in Video and Audiovisual LLMs}

\author{Killian Steunou,~\IEEEmembership{} Yannis Tevissen,~\IEEEmembership{Member,~IEEE,} Mounîm A. El Yacoubi,~\IEEEmembership{Senior Member,~IEEE}
  % <-this % stops a space
  \thanks{K. Steunou is with the SAMOVAR Laboratory, T\'el\'ecom SudParis, Institut Polytechnique de Paris, Palaiseau, France, and also with Moments Lab, Paris, France (e-mail: killian.steunou@ip-paris.fr).}% <-this % stops a space
\thanks{Y. Tevissen is with Moments Lab, Paris, France (e-mail: yannis.tevissen@momentslab.com).}
\thanks{M. A. El Yacoubi is with the SAMOVAR Laboratory, T\'el\'ecom SudParis, Institut Polytechnique de Paris, Palaiseau, France (e-mail: mounim.el\_yacoubi@telecom-sudparis.eu).}
% \thanks{Manuscript received ; revised .}}
\thanks{This work has been submitted to the IEEE for possible publication. Copyright may be transferred without notice, after which this version may no longer be accessible.}}

% The paper headers
%\markboth{IEEE Transactions on Pattern Analysis and Machine Intelligence,Vol.~48, No.~9, September~2026}%
%{Steunou \MakeLowercase{\textit{et al.}}: Why Is Video Still So Expensive? A Survey}
\markboth{Under Review. September 2026}%
{Steunou \MakeLowercase{\textit{et al.}}: Why Is Video Still So Expensive? A Survey}

\IEEEpubid{\copyright~2026 The Authors}
% Remember, if you use this you must call \IEEEpubidadjcol in the second
% column for its text to clear the IEEEpubid mark.

\maketitle

\begin{abstract}

  Video understanding has rapidly evolved toward video large language models (VideoLLMs): systems that couple video representations with pretrained large language models and condition generation on a textual prompt. Their strong performance on captioning, question answering, retrieval and temporal grounding comes at a computation and memory cost that grows with frame count and context length, limiting deployment in real-time, mobile and resource-constrained settings. This survey covers inference-efficiency mechanisms for visual and audiovisual VideoLLMs that report concrete reductions in parameter count, FLOPs per input, latency, memory, or visual and audio token count. We analyze bottlenecks across frame sampling, modality encoding, connector-level token reduction, and LLM prefilling and decoding. We organize methods by the pipeline stage at which they act, covering VideoLLMs developed since late 2022 together with earlier frame-sampling and vision-encoder mechanisms that remain components of current pipelines. We assemble literature-reported accuracy--cost comparisons under shared host models and input protocols wherever available, distinguish them from heterogeneous cross-paper evidence, and identify gaps in audiovisual efficiency and standardized evaluation. We maintain a repository at \mbox{\url{https://github.com/momentslab/awesome-efficient-videollm}}.

\end{abstract}


\section{Introduction}
\label{sec:intro}

% AUTO-GENERATED by figures/gen_timeline.py -- edit the script, not this file.
% Timeline of every method in the survey taxonomy, colored by pipeline stage
% (same colors as Fig.~\ref{fig:efficiency_taxonomy}); gray = several stages.
\definecolor{color-frame}{HTML}{E4FF77}    % Moments Citron
\definecolor{color-backbone}{HTML}{FF8934} % Moments Lab Orange
\definecolor{color-context}{HTML}{6342E8}  % Moments Purple
\definecolor{color-llm}{HTML}{ADAAFF}      % Moments Lavender
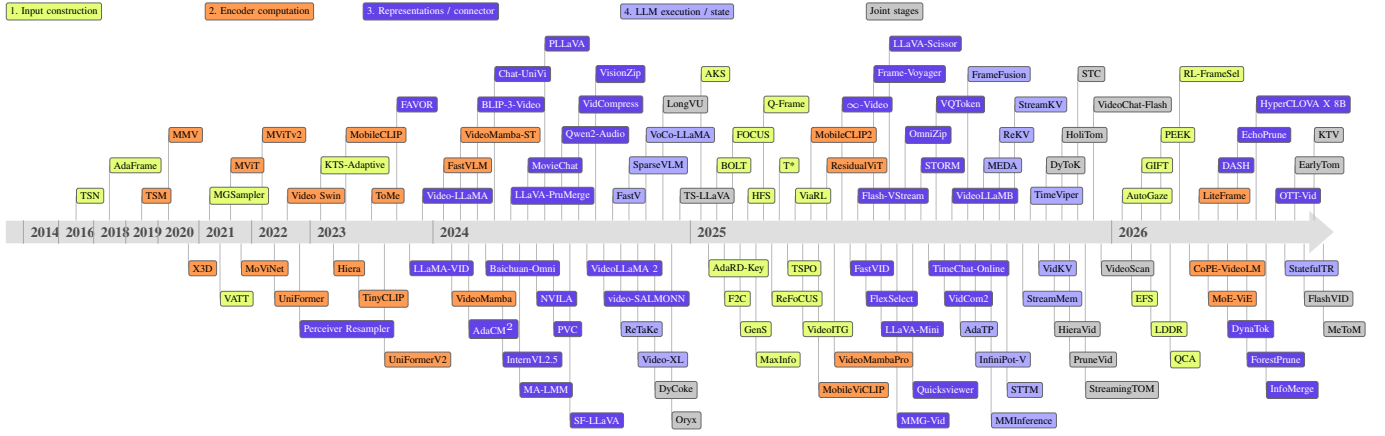
\begin{figure*}[t]
\centering
\resizebox{\textwidth}{!}{%
\begin{tikzpicture}[
  stem/.style={gray!55, line width=0.45pt},
  bx/.style={draw=black!60, line width=0.4pt, rounded corners=1pt, inner sep=1.9pt, font=\tiny},
  bF/.style={bx, fill=color-frame},
  bE/.style={bx, fill=color-backbone!85},
  bC/.style={bx, fill=color-context, text=white},
  bL/.style={bx, fill=color-llm},
  bM/.style={bx, fill=black!22},
]
% ---- timeline arrow ----
\draw[line width=11pt, gray!25, -{Triangle[length=12pt, width=20pt]}] (-0.3,0) -- (22.4,0);
\node[font=\scriptsize\bfseries, text=black!70, anchor=west] at (0.00,0) {2014};
\draw[black!35, line width=0.5pt] (0.00,-0.20) -- (0.00,0.20);
\node[font=\scriptsize\bfseries, text=black!70, anchor=west] at (0.60,0) {2016};
\draw[black!35, line width=0.5pt] (0.60,-0.20) -- (0.60,0.20);
\node[font=\scriptsize\bfseries, text=black!70, anchor=west] at (1.20,0) {2018};
\draw[black!35, line width=0.5pt] (1.20,-0.20) -- (1.20,0.20);
\node[font=\scriptsize\bfseries, text=black!70, anchor=west] at (1.75,0) {2019};
\draw[black!35, line width=0.5pt] (1.75,-0.20) -- (1.75,0.20);
\node[font=\scriptsize\bfseries, text=black!70, anchor=west] at (2.30,0) {2020};
\draw[black!35, line width=0.5pt] (2.30,-0.20) -- (2.30,0.20);
\node[font=\scriptsize\bfseries, text=black!70, anchor=west] at (3.00,0) {2021};
\draw[black!35, line width=0.5pt] (3.00,-0.20) -- (3.00,0.20);
\node[font=\scriptsize\bfseries, text=black!70, anchor=west] at (3.90,0) {2022};
\draw[black!35, line width=0.5pt] (3.90,-0.20) -- (3.90,0.20);
\node[font=\scriptsize\bfseries, text=black!70, anchor=west] at (4.90,0) {2023};
\draw[black!35, line width=0.5pt] (4.90,-0.20) -- (4.90,0.20);
\node[font=\scriptsize\bfseries, text=black!70, anchor=west] at (7.00,0) {2024};
\draw[black!35, line width=0.5pt] (7.00,-0.20) -- (7.00,0.20);
\node[font=\scriptsize\bfseries, text=black!70, anchor=west] at (11.40,0) {2025};
\draw[black!35, line width=0.5pt] (11.40,-0.20) -- (11.40,0.20);
\node[font=\scriptsize\bfseries, text=black!70, anchor=west] at (18.60,0) {2026};
\draw[black!35, line width=0.5pt] (18.60,-0.20) -- (18.60,0.20);
% ---- stems ----
\draw[stem] (0.90,0.2) -- (0.90,0.62);
\draw[stem] (1.47,0.2) -- (1.47,1.14);
\draw[stem] (2.02,0.2) -- (2.02,0.62);
\draw[stem] (2.48,0.2) -- (2.48,1.66);
\draw[stem] (3.18,0.2) -- (3.18,0.62);
\draw[stem] (3.54,0.2) -- (3.54,1.14);
\draw[stem] (4.08,0.2) -- (4.08,1.66);
\draw[stem] (4.51,0.2) -- (4.51,0.62);
\draw[stem] (5.08,0.2) -- (5.08,1.14);
\draw[stem] (5.51,0.2) -- (5.51,1.66);
\draw[stem] (5.95,0.2) -- (5.95,0.62);
\draw[stem] (6.38,0.2) -- (6.38,2.18);
\draw[stem] (6.82,0.2) -- (6.82,0.62);
\draw[stem] (7.18,0.2) -- (7.18,1.14);
\draw[stem] (7.47,0.2) -- (7.47,1.66);
\draw[stem] (7.76,0.2) -- (7.76,2.18);
\draw[stem] (8.05,0.2) -- (8.05,2.70);
\draw[stem] (8.33,0.2) -- (8.33,0.62);
\draw[stem] (8.62,0.2) -- (8.62,1.14);
\draw[stem] (8.91,0.2) -- (8.91,3.22);
\draw[stem] (9.20,0.2) -- (9.20,1.66);
\draw[stem] (9.49,0.2) -- (9.49,2.18);
\draw[stem] (9.78,0.2) -- (9.78,2.70);
\draw[stem] (10.07,0.2) -- (10.07,0.62);
\draw[stem] (10.35,0.2) -- (10.35,1.14);
\draw[stem] (10.64,0.2) -- (10.64,1.66);
\draw[stem] (10.93,0.2) -- (10.93,2.18);
\draw[stem] (11.22,0.2) -- (11.22,0.62);
\draw[stem] (11.58,0.2) -- (11.58,2.70);
\draw[stem] (11.85,0.2) -- (11.85,1.14);
\draw[stem] (12.12,0.2) -- (12.12,1.66);
\draw[stem] (12.38,0.2) -- (12.38,0.62);
\draw[stem] (12.65,0.2) -- (12.65,2.18);
\draw[stem] (12.92,0.2) -- (12.92,1.14);
\draw[stem] (13.19,0.2) -- (13.19,0.62);
\draw[stem] (13.46,0.2) -- (13.46,1.66);
\draw[stem] (13.73,0.2) -- (13.73,1.14);
\draw[stem] (13.99,0.2) -- (13.99,2.18);
\draw[stem] (14.26,0.2) -- (14.26,0.62);
\draw[stem] (14.53,0.2) -- (14.53,2.70);
\draw[stem] (14.80,0.2) -- (14.80,3.22);
\draw[stem] (15.07,0.2) -- (15.07,1.66);
\draw[stem] (15.34,0.2) -- (15.34,1.14);
\draw[stem] (15.60,0.2) -- (15.60,2.18);
\draw[stem] (15.87,0.2) -- (15.87,0.62);
\draw[stem] (16.14,0.2) -- (16.14,2.70);
\draw[stem] (16.41,0.2) -- (16.41,1.14);
\draw[stem] (16.68,0.2) -- (16.68,1.66);
\draw[stem] (16.94,0.2) -- (16.94,2.18);
\draw[stem] (17.21,0.2) -- (17.21,0.62);
\draw[stem] (17.48,0.2) -- (17.48,1.14);
\draw[stem] (17.75,0.2) -- (17.75,1.66);
\draw[stem] (18.02,0.2) -- (18.02,2.70);
\draw[stem] (18.29,0.2) -- (18.29,2.18);
\draw[stem] (18.78,0.2) -- (18.78,0.62);
\draw[stem] (19.11,0.2) -- (19.11,1.14);
\draw[stem] (19.44,0.2) -- (19.44,1.66);
\draw[stem] (19.76,0.2) -- (19.76,2.70);
\draw[stem] (20.09,0.2) -- (20.09,0.62);
\draw[stem] (20.42,0.2) -- (20.42,1.14);
\draw[stem] (20.75,0.2) -- (20.75,1.66);
\draw[stem] (21.07,0.2) -- (21.07,2.18);
\draw[stem] (21.40,0.2) -- (21.40,0.62);
\draw[stem] (21.73,0.2) -- (21.73,1.14);
\draw[stem] (22.06,0.2) -- (22.06,1.66);
\draw[stem] (2.82,-0.2) -- (2.82,-0.62);
\draw[stem] (3.36,-0.2) -- (3.36,-1.14);
\draw[stem] (3.72,-0.2) -- (3.72,-0.62);
\draw[stem] (4.29,-0.2) -- (4.29,-1.14);
\draw[stem] (4.72,-0.2) -- (4.72,-1.66);
\draw[stem] (5.30,-0.2) -- (5.30,-0.62);
\draw[stem] (5.73,-0.2) -- (5.73,-1.14);
\draw[stem] (6.17,-0.2) -- (6.17,-2.18);
\draw[stem] (6.60,-0.2) -- (6.60,-0.62);
\draw[stem] (7.32,-0.2) -- (7.32,-1.14);
\draw[stem] (7.61,-0.2) -- (7.61,-1.66);
\draw[stem] (7.90,-0.2) -- (7.90,-0.62);
\draw[stem] (8.19,-0.2) -- (8.19,-2.18);
\draw[stem] (8.48,-0.2) -- (8.48,-2.70);
\draw[stem] (8.77,-0.2) -- (8.77,-1.14);
\draw[stem] (9.06,-0.2) -- (9.06,-1.66);
\draw[stem] (9.34,-0.2) -- (9.34,-3.22);
\draw[stem] (9.63,-0.2) -- (9.63,-0.62);
\draw[stem] (9.92,-0.2) -- (9.92,-1.14);
\draw[stem] (10.21,-0.2) -- (10.21,-1.66);
\draw[stem] (10.50,-0.2) -- (10.50,-2.18);
\draw[stem] (10.79,-0.2) -- (10.79,-2.70);
\draw[stem] (11.08,-0.2) -- (11.08,-3.22);
\draw[stem] (11.71,-0.2) -- (11.71,-0.62);
\draw[stem] (11.98,-0.2) -- (11.98,-1.14);
\draw[stem] (12.25,-0.2) -- (12.25,-1.66);
\draw[stem] (12.52,-0.2) -- (12.52,-2.18);
\draw[stem] (12.79,-0.2) -- (12.79,-1.14);
\draw[stem] (13.06,-0.2) -- (13.06,-0.62);
\draw[stem] (13.32,-0.2) -- (13.32,-1.66);
\draw[stem] (13.59,-0.2) -- (13.59,-2.70);
\draw[stem] (13.86,-0.2) -- (13.86,-2.18);
\draw[stem] (14.13,-0.2) -- (14.13,-0.62);
\draw[stem] (14.40,-0.2) -- (14.40,-1.14);
\draw[stem] (14.66,-0.2) -- (14.66,-1.66);
\draw[stem] (14.93,-0.2) -- (14.93,-3.22);
\draw[stem] (15.20,-0.2) -- (15.20,-2.70);
\draw[stem] (15.47,-0.2) -- (15.47,-0.62);
\draw[stem] (15.74,-0.2) -- (15.74,-1.14);
\draw[stem] (16.01,-0.2) -- (16.01,-1.66);
\draw[stem] (16.27,-0.2) -- (16.27,-2.18);
\draw[stem] (16.54,-0.2) -- (16.54,-3.22);
\draw[stem] (16.81,-0.2) -- (16.81,-2.70);
\draw[stem] (17.08,-0.2) -- (17.08,-1.14);
\draw[stem] (17.35,-0.2) -- (17.35,-0.62);
\draw[stem] (17.62,-0.2) -- (17.62,-1.66);
\draw[stem] (17.88,-0.2) -- (17.88,-2.18);
\draw[stem] (18.15,-0.2) -- (18.15,-2.70);
\draw[stem] (18.42,-0.2) -- (18.42,-0.62);
\draw[stem] (18.94,-0.2) -- (18.94,-1.14);
\draw[stem] (19.27,-0.2) -- (19.27,-1.66);
\draw[stem] (19.60,-0.2) -- (19.60,-2.18);
\draw[stem] (19.93,-0.2) -- (19.93,-0.62);
\draw[stem] (20.25,-0.2) -- (20.25,-1.14);
\draw[stem] (20.58,-0.2) -- (20.58,-1.66);
\draw[stem] (20.91,-0.2) -- (20.91,-2.18);
\draw[stem] (21.24,-0.2) -- (21.24,-2.70);
\draw[stem] (21.56,-0.2) -- (21.56,-0.62);
\draw[stem] (21.89,-0.2) -- (21.89,-1.14);
\draw[stem] (22.22,-0.2) -- (22.22,-1.66);
% ---- boxes ----
\node[bF, anchor=west] at (0.90,0.62) {TSN};
\node[bF, anchor=west] at (1.47,1.14) {AdaFrame};
\node[bE, anchor=west] at (2.02,0.62) {TSM};
\node[bE, anchor=west] at (2.48,1.66) {MMV};
\node[bF, anchor=west] at (3.18,0.62) {MGSampler};
\node[bE, anchor=west] at (3.54,1.14) {MViT};
\node[bE, anchor=west] at (4.08,1.66) {MViTv2};
\node[bE, anchor=west] at (4.51,0.62) {Video Swin};
\node[bF, anchor=west] at (5.08,1.14) {KTS-Adaptive};
\node[bE, anchor=west] at (5.51,1.66) {MobileCLIP};
\node[bE, anchor=west] at (5.95,0.62) {ToMe};
\node[bC, anchor=west] at (6.38,2.18) {FAVOR};
\node[bC, anchor=west] at (6.82,0.62) {Video-LLaMA};
\node[bE, anchor=west] at (7.18,1.14) {FastVLM};
\node[bE, anchor=west] at (7.47,1.66) {VideoMamba-ST};
\node[bC, anchor=west] at (7.76,2.18) {BLIP-3-Video};
\node[bC, anchor=west] at (8.05,2.70) {Chat-UniVi};
\node[bC, anchor=west] at (8.33,0.62) {LLaVA-PruMerge};
\node[bC, anchor=west] at (8.62,1.14) {MovieChat};
\node[bC, anchor=west] at (8.91,3.22) {PLLaVA};
\node[bC, anchor=west] at (9.20,1.66) {Qwen2-Audio};
\node[bC, anchor=west] at (9.49,2.18) {VidCompress};
\node[bC, anchor=west] at (9.78,2.70) {VisionZip};
\node[bL, anchor=west] at (10.07,0.62) {FastV};
\node[bL, anchor=west] at (10.35,1.14) {SparseVLM};
\node[bL, anchor=west] at (10.64,1.66) {VoCo-LLaMA};
\node[bM, anchor=west] at (10.93,2.18) {LongVU};
\node[bM, anchor=west] at (11.22,0.62) {TS-LLaVA};
\node[bF, anchor=west] at (11.58,2.70) {AKS};
\node[bF, anchor=west] at (11.85,1.14) {BOLT};
\node[bF, anchor=west] at (12.12,1.66) {FOCUS};
\node[bF, anchor=west] at (12.38,0.62) {HFS};
\node[bF, anchor=west] at (12.65,2.18) {Q-Frame};
\node[bF, anchor=west] at (12.92,1.14) {T*};
\node[bF, anchor=west] at (13.19,0.62) {ViaRL};
\node[bE, anchor=west] at (13.46,1.66) {MobileCLIP2};
\node[bE, anchor=west] at (13.73,1.14) {ResidualViT};
\node[bC, anchor=west] at (13.99,2.18) {$\infty$-Video};
\node[bC, anchor=west] at (14.26,0.62) {Flash-VStream};
\node[bC, anchor=west] at (14.53,2.70) {Frame-Voyager};
\node[bC, anchor=west] at (14.80,3.22) {LLaVA-Scissor};
\node[bC, anchor=west] at (15.07,1.66) {OmniZip};
\node[bC, anchor=west] at (15.34,1.14) {STORM};
\node[bC, anchor=west] at (15.60,2.18) {VQToken};
\node[bC, anchor=west] at (15.87,0.62) {VideoLLaMB};
\node[bL, anchor=west] at (16.14,2.70) {FrameFusion};
\node[bL, anchor=west] at (16.41,1.14) {MEDA};
\node[bL, anchor=west] at (16.68,1.66) {ReKV};
\node[bL, anchor=west] at (16.94,2.18) {StreamKV};
\node[bL, anchor=west] at (17.21,0.62) {TimeViper};
\node[bM, anchor=west] at (17.48,1.14) {DyToK};
\node[bM, anchor=west] at (17.75,1.66) {HoliTom};
\node[bM, anchor=west] at (18.02,2.70) {STC};
\node[bM, anchor=west] at (18.29,2.18) {VideoChat-Flash};
\node[bF, anchor=west] at (18.78,0.62) {AutoGaze};
\node[bF, anchor=west] at (19.11,1.14) {GIFT};
\node[bF, anchor=west] at (19.44,1.66) {PEEK};
\node[bF, anchor=west] at (19.76,2.70) {RL-FrameSel};
\node[bE, anchor=west] at (20.09,0.62) {LiteFrame};
\node[bC, anchor=west] at (20.42,1.14) {DASH};
\node[bC, anchor=west] at (20.75,1.66) {EchoPrune};
\node[bC, anchor=west] at (21.07,2.18) {HyperCLOVA X 8B};
\node[bC, anchor=west] at (21.40,0.62) {OTT-Vid};
\node[bM, anchor=west] at (21.73,1.14) {EarlyTom};
\node[bM, anchor=west] at (22.06,1.66) {KTV};
\node[bE, anchor=west] at (2.82,-0.62) {X3D};
\node[bF, anchor=west] at (3.36,-1.14) {VATT};
\node[bE, anchor=west] at (3.72,-0.62) {MoViNet};
\node[bE, anchor=west] at (4.29,-1.14) {UniFormer};
\node[bC, anchor=west] at (4.72,-1.66) {Perceiver Resampler};
\node[bE, anchor=west] at (5.30,-0.62) {Hiera};
\node[bE, anchor=west] at (5.73,-1.14) {TinyCLIP};
\node[bE, anchor=west] at (6.17,-2.18) {UniFormerV2};
\node[bC, anchor=west] at (6.60,-0.62) {LLaMA-VID};
\node[bE, anchor=west] at (7.32,-1.14) {VideoMamba};
\node[bC, anchor=west] at (7.61,-1.66) {AdaCM$^2$};
\node[bC, anchor=west] at (7.90,-0.62) {Baichuan-Omni};
\node[bC, anchor=west] at (8.19,-2.18) {InternVL2.5};
\node[bC, anchor=west] at (8.48,-2.70) {MA-LMM};
\node[bC, anchor=west] at (8.77,-1.14) {NVILA};
\node[bC, anchor=west] at (9.06,-1.66) {PVC};
\node[bC, anchor=west] at (9.34,-3.22) {SF-LLaVA};
\node[bC, anchor=west] at (9.63,-0.62) {VideoLLaMA 2};
\node[bC, anchor=west] at (9.92,-1.14) {video-SALMONN};
\node[bL, anchor=west] at (10.21,-1.66) {ReTaKe};
\node[bL, anchor=west] at (10.50,-2.18) {Video-XL};
\node[bM, anchor=west] at (10.79,-2.70) {DyCoke};
\node[bM, anchor=west] at (11.08,-3.22) {Oryx};
\node[bF, anchor=west] at (11.71,-0.62) {AdaRD-Key};
\node[bF, anchor=west] at (11.98,-1.14) {F2C};
\node[bF, anchor=west] at (12.25,-1.66) {GenS};
\node[bF, anchor=west] at (12.52,-2.18) {MaxInfo};
\node[bF, anchor=west] at (12.79,-1.14) {ReFoCUS};
\node[bF, anchor=west] at (13.06,-0.62) {TSPO};
\node[bF, anchor=west] at (13.32,-1.66) {VideoITG};
\node[bE, anchor=west] at (13.59,-2.70) {MobileViCLIP};
\node[bE, anchor=west] at (13.86,-2.18) {VideoMambaPro};
\node[bC, anchor=west] at (14.13,-0.62) {FastVID};
\node[bC, anchor=west] at (14.40,-1.14) {FlexSelect};
\node[bC, anchor=west] at (14.66,-1.66) {LLaVA-Mini};
\node[bC, anchor=west] at (14.93,-3.22) {MMG-Vid};
\node[bC, anchor=west] at (15.20,-2.70) {Quicksviewer};
\node[bC, anchor=west] at (15.47,-0.62) {TimeChat-Online};
\node[bC, anchor=west] at (15.74,-1.14) {VidCom2};
\node[bL, anchor=west] at (16.01,-1.66) {AdaTP};
\node[bL, anchor=west] at (16.27,-2.18) {InfiniPot-V};
\node[bL, anchor=west] at (16.54,-3.22) {MMInference};
\node[bL, anchor=west] at (16.81,-2.70) {STTM};
\node[bL, anchor=west] at (17.08,-1.14) {StreamMem};
\node[bL, anchor=west] at (17.35,-0.62) {VidKV};
\node[bM, anchor=west] at (17.62,-1.66) {HieraVid};
\node[bM, anchor=west] at (17.88,-2.18) {PruneVid};
\node[bM, anchor=west] at (18.15,-2.70) {StreamingTOM};
\node[bM, anchor=west] at (18.42,-0.62) {VideoScan};
\node[bF, anchor=west] at (18.94,-1.14) {EFS};
\node[bF, anchor=west] at (19.27,-1.66) {LDDR};
\node[bF, anchor=west] at (19.60,-2.18) {QCA};
\node[bE, anchor=west] at (19.93,-0.62) {CoPE-VideoLM};
\node[bE, anchor=west] at (20.25,-1.14) {MoE-ViE};
\node[bC, anchor=west] at (20.58,-1.66) {DynaTok};
\node[bC, anchor=west] at (20.91,-2.18) {ForestPrune};
\node[bC, anchor=west] at (21.24,-2.70) {InfoMerge};
\node[bL, anchor=west] at (21.56,-0.62) {StatefulTR};
\node[bM, anchor=west] at (21.89,-1.14) {FlashVID};
\node[bM, anchor=west] at (22.22,-1.66) {MeToM};
% ---- legend ----
\node[bF, anchor=west] at (-0.3,3.77) {1. Input construction};
\node[bE, anchor=west] at (3.1,3.77) {2. Encoder computation};
\node[bC, anchor=west] at (5.8,3.77) {3. Representations / connector};
\node[bL, anchor=west] at (10.2,3.77) {4. LLM execution / state};
\node[bM, anchor=west] at (14.4,3.77) {Joint stages};
\end{tikzpicture}%
}
\caption{Chronology of the reviewed inventory. Colors denote the stage at which a method reduces cost and gray denotes methods acting at several stages. The year axis is non-linear. Methods evaluated outside a VideoLLM are identified in Figure~\ref{fig:efficiency_taxonomy}.}
\label{fig:method_timeline}
\end{figure*}

\IEEEPARstart{V}{ideo} content spans short social media clips, instructional videos, movies and long-form egocentric recordings, combining spatial, temporal and multimodal cues: frames, audio, speech, subtitles and overlaid text. Video understanding has shifted from task-specific architectures to large pre-trained foundation models, trained on video-text corpora to support captioning, question answering, retrieval, spatiotemporal grounding and dense summarization \cite{tongVideoMAEMaskedAutoencoders2022,tangVideoUnderstandingLarge2023,madanFoundationModelsVideo2024,nguyenVideoLanguageUnderstandingSurvey2024,faragVideoCaptioningUsing2026}. Video large language models (VideoLLMs) extend text-only LLMs with visual encoders and, in audiovisual systems, audio encoders \cite{tangVideoUnderstandingLarge2023,yinSurveyMultimodalLarge2024}, supporting open-ended reasoning and instruction following on video-centric tasks, often without task-specific fine-tuning.

However, these advances come with substantial computational and memory costs~\cite{wengLongVLMEfficientLong2024}: video encoders may process hundreds of high-resolution frames per clip across multiple modalities and long temporal contexts, and large language backbones add attention compute and inference-time memory overhead~\cite{chatterjeeMemoryefficientStreamingVideoLLMs2025,ningLiveVLM2025}. Efficient VideoLLMs aim to retain these semantic and reasoning capabilities while reducing parameter count, floating-point operations (FLOPs) per input, latency or memory.

Typical VideoLLMs share a pipeline made of four stages: (1) construct the visual input by selecting frames, patches and resolution, (2) encode it with a vision backbone, (3) reduce and map the encoded representations to the LLM input space, and (4) process them together with a textual prompt in the LLM.

Recent work explores the capability vs. efficiency trade-off throughout the pipeline: selecting fewer frames before encoding, lighter vision backbones, compressing connector outputs, pruning visual tokens inside the LLM, and reducing the visual key-value (KV) cache \cite{bhardwajEfficientVideoClassification2019,tangAdaptiveKeyframeSampling2025,huang-etal-2025-prunevid,li2026echoprune,tao2025vidkv}. Audiovisual systems additionally compress audio tokens or use sound to guide visual selection \cite{taoOmniZipAudioGuidedDynamic2025,taoOmniAgentAudioGuidedActive2025}. Figure~\ref{fig:method_timeline} traces these mechanisms across the four pipeline stages. Because these ideas are often proposed in isolation, tied to particular tasks such as captioning, question answering (QA) or temporal localization, and evaluated with heterogeneous metrics, it is difficult to determine where computational cost actually goes and which strategy is most effective under a given constraint. This fragmentation motivates a video-specific synthesis that relates reported efficiency gains to their pipeline stage, input coverage and evaluation conditions.\IEEEpubidadjcol

Recent surveys approach video understanding from complementary perspectives. Madan \textit{et al.}~\cite{madanFoundationModelsVideo2024}, Nguyen \textit{et al.}~\cite{nguyenVideoLanguageUnderstandingSurvey2024}, and Tang \textit{et al.}~\cite{tangVideoUnderstandingLarge2023} review video foundation models, video-language learning, and VideoLLM architectures, respectively, emphasizing capabilities, tasks and benchmarks. Other surveys focus on long video understanding~\cite{zouSecondsHoursReviewing2024}, temporal grounding~\cite{wuSurveyVideoTemporal2025}, evaluation protocols~\cite{kumarVideoLLMBenchmarksEvaluation2025}, and omni-modal language models~\cite{chenSurveyOmnimodalLanguage2025}. General multimodal LLM (MLLM) surveys place video within a broader landscape of modalities and architectures~\cite{yinSurveyMultimodalLarge2024,baiSurveyMultimodalLarge2024,caffagniRevolutionMultimodalLarge2024}.

Efficiency-focused surveys overlap more directly with our scope. Jin \textit{et al.}~\cite{jinEfficientMultimodalLarge2024} cover efficient MLLM architectures, vision and language components, and training strategies, with video discussed as an application. Shao \textit{et al.}~\cite{shaoWhenTokensTalk2025} organize token compression by its underlying mechanisms across images, videos and audio; they also compare video compression methods under specified host models and token budgets. Their treatment provides a mechanism-centered account of token reduction, while our scope additionally includes frame-selection strategies and efficient video-encoder architectures, including mechanisms evaluated before the emergence of VideoLLMs.

Two recent surveys explicitly adopt a pipeline perspective. Zhang \textit{et al.}~\cite{zhang2026efficientinference} organize Large Vision-Language Models (LVLM) inference around encoding, prefilling and decoding, including keyframe selection, and analyze how optimization at one stage affects downstream bottlenecks. Wu \textit{et al.}~\cite{wu2026compressionlifecycle} organize MLLM compression by input, encoder, projector and LLM intervention points, crossed with five compression operations. These works establish pipeline structure and cross-stage cost interactions as shared foundations for efficiency analysis.

Our contribution is a video-centric synthesis built on these foundations. We connect frame selection and video-encoder design to connector compression and LLM-side inference, and examine how audio-token reduction and audio-guided visual selection affect the joint audiovisual workload. We assemble literature-reported comparisons within shared host models, input settings and token budgets wherever available, and distinguish these from comparisons across heterogeneous systems. Our emphasis is on how temporal coverage, encoder cost and multimodal token budgets jointly determine the benefits and limits of video inference-efficiency mechanisms.

In our survey, a \emph{VideoLLM} is an \emph{encoder--connector--LLM} system (illustrated in Figure~\ref{fig:vid_llm_overview}) that provides video representations and a textual prompt to a pretrained LLM; visual-only systems encode frames, while audiovisual VideoLLMs additionally encode synchronized audio (Section~\ref{sec:methodo} details how the surveyed methods were selected). We use ``efficient'' in a system-level sense: for a given task and hardware regime, an efficient method preserves or improves semantic performance while reducing parameter count, FLOPs per input, wall-clock latency, or memory; power and energy are also relevant but remain rarely reported \cite{kumarVideoLLMBenchmarksEvaluation2025}. Sections~\ref{subsec:bottlenecks} and~\ref{sec:efficiency-mechanisms} make this definition concrete through pipeline costs and the metrics reported in the literature.

We make the following contributions:
\begin{itemize}
  \item We synthesize video-specific inference-efficiency mechanisms across frame sampling, vision-encoder design, connector-level reduction and LLM-side processing, connecting upstream temporal coverage and encoding cost to downstream token and memory budgets.
  \item We assemble literature-reported accuracy--cost comparisons and identify which methods can be compared under shared hosts and evaluation settings. We separate these comparisons from heterogeneous cross-paper results and make differences in input protocols and FLOP-accounting boundaries explicit.
  \item We examine audiovisual efficiency through audio-token compression, audio-guided visual selection and joint token budgets, and use the evidence across stages to identify evaluation gaps and priorities for efficient VideoLLMs.
\end{itemize}

The remainder of this survey is structured as follows. Section~\ref{sec:data_eval} presents our paper-selection protocol, and defines the tasks and evaluation protocols; Section~\ref{sec:architectures} reviews representative VideoLLM architectures and their computational bottlenecks; Section~\ref{sec:efficiency-mechanisms} introduces the taxonomy and compares methods on shared benchmarks; Sections~\ref{sec:discussion} and~\ref{sec:conclusion} discuss trends and open challenges, and conclude.

\section{Preliminaries: Survey Scope, Tasks and Evaluation Protocols}
\label{sec:data_eval}

\subsection{Survey Scope and Paper Selection}
\label{sec:methodo}

We survey efficiency mechanisms along the inference pipeline, from frame selection and modality encoding to connector-level token reduction, LLM prefilling (the forward pass over the full prompt, before any token is generated), decoding, and KV-cache use. We identified candidate methods through keyword searches on arXiv and Google Scholar, combining VideoLLM terms with efficiency terms such as token pruning, token merging, frame selection and KV-cache compression, and through backward and forward citation snowballing from the surveys discussed in the introduction and from each retained method. We cover papers published or posted as preprints up to August 2026.

A method enters the taxonomy when it contributes or evaluates a targeted mechanism and reports a concrete effect on parameter count, FLOPs, retained-token count, latency, or memory. We focus on VideoLLMs developed since late 2022. Earlier frame-sampling and vision-encoder methods are included when they remain components or direct antecedents of current pipelines. Audiovisual methods are included when they reduce the audio-token stream, use audio to reduce visual processing, or bound the joint audiovisual token stream. Training-only methods, generic LLM optimizations, and image-only techniques are cited as adjacent context when they establish or directly supply a mechanism adopted by VideoLLMs. The taxonomy has no model-size limit, but our quantitative tables emphasize language backbones around 7B--8B parameters, so a mechanism demonstrated only on a larger host appears in the taxonomy but not in the comparisons. Section~\ref{subsec:efficiency-metrics} details the model-size and reporting conventions behind our comparisons.

These searches surfaced several hundred candidate papers. We screened titles and abstracts against the criteria above, and read the remaining papers in full, retaining 125. Figure~\ref{fig:efficiency_taxonomy} shows all of them, marking the encoder and sampling methods that predate VideoLLMs. A paper appears in several families when it reduces cost at several stages, so family sizes add up to more than the number of papers. The selection is representative: new efficiency methods appear every month, and many recent methods apply an established lever at a different stage or granularity. When several papers instantiate the same mechanism, we keep those with the most complete efficiency reporting and the clearest evaluation protocol, cite close variants as context, and favor methods whose input and measurement settings support the controlled comparisons of Section~\ref{sec:efficiency-mechanisms}.

\subsection{Tasks, Benchmarks and Evaluation}
\label{subsec:videollm_benchmarks}

Video understanding spans classification, grounding, captioning, retrieval, question answering (QA) and dialogue, operating on RGB (Red Green Blue) frames with optional synchronized audio and derived text such as subtitles, automatic speech recognition (ASR) transcripts or optical character recognition (OCR) tokens. Each task family has standard datasets and metrics: action recognition and temporal localization (top-1/top-5 accuracy; mAP at temporal IoU thresholds) on Kinetics \cite{kayKineticsHumanAction2017}, Something-Something V2 \cite{goyalSomethingSomethingVideo2017} and Ego4D \cite{graumanEgo4DWorld30002022}; clip-level and dense captioning (BLEU, METEOR, ROUGE-L, CIDEr) on MSR-VTT \cite{xuMSRVTTLargeVideo2016} and ActivityNet Captions \cite{krishnaDenseCaptioningEventsVideos2017}; video QA (accuracy) on ActivityNet-QA \cite{yuActivityNetQADatasetUnderstanding2019}, NExT-QA \cite{xiaoNExTQANextPhase2021} and EgoSchema \cite{mangalamEgoSchemaDiagnosticBenchmark2023}; text--video retrieval (R@K, median rank) on caption datasets and narrated corpora such as HowTo100M \cite{miechHowTo100MLearningTextVideo2019}; and temporal grounding (R@K at temporal IoU) on Charades-STA \cite{gaoTALLTemporalActivity2017} and Ego4D NLQ \cite{graumanEgo4DWorld30002022}. Efficiency-specific protocols are discussed in Section~\ref{subsec:efficiency-metrics}.

VideoLLM benchmarks complement these task-specific datasets by evaluating multiple capabilities under standardized protocols, most commonly through multiple-choice or structured QA, with emphasis on temporal reasoning beyond single-frame cues, long-context comprehension, and modality ablations. The efficiency comparisons later in this survey concentrate on MVBench \cite{liMVBenchComprehensiveMultimodal2024}, Video-MME \cite{fuVideoMMEFirstEverComprehensive2025}, EgoSchema \cite{mangalamEgoSchemaDiagnosticBenchmark2023} and LongVideoBench \cite{wuLongVideoBenchBenchmarkLongcontext2024} because they are the benchmarks most often shared by the methods we survey (Tables~\ref{tab:frame_sampling_isobackbone}--\ref{tab:llmside_hieravid}). Table~\ref{tab:videollm_benchmarks} summarizes the benchmarks that appear in our comparisons and discussion; a full inventory of recent VideoLLM benchmarks is provided in the supplementary material.

\begin{table}[t]\scriptsize
  \centering
  \caption{VideoLLM benchmarks used in the comparisons of this survey. Mod.\ denotes the modalities provided beyond the question text (V=video, A=audio, T=transcript/subtitles). Dur.: short ($<$1~min), medium (1--10~min), long ($>$10~min). \#V = Number of videos, \#Q = Number of questions. Fmt.: MCQ = Multiple Choice Question, OE = Open Ended}
  \label{tab:videollm_benchmarks}
  \setlength{\tabcolsep}{3.5pt}
  \renewcommand{\arraystretch}{1.1}
  \begin{tabular}{l c c c r r}
    \toprule
    \textbf{Benchmark} & \textbf{Mod.} & \textbf{Fmt.} & \textbf{Dur.} & \textbf{\#V} & \textbf{\#Q} \\
    \midrule
    MVBench \cite{liMVBenchComprehensiveMultimodal2024} & V & MCQ & S & 3,641 & 4,000 \\
    Video-MME \cite{fuVideoMMEFirstEverComprehensive2025} & V+A+T & MCQ & S/M/L & 900 & 2,700 \\
    EgoSchema \cite{mangalamEgoSchemaDiagnosticBenchmark2023} & V & MCQ & M & 5,063 & 5,063 \\
    LongVideoBench \cite{wuLongVideoBenchBenchmarkLongcontext2024} & V+T & MCQ & M/L & 3,763 & 6,678 \\
    MLVU \cite{zhouMLVUBenchmarkingMultitask2025} & V & MCQ & M/L & 1,730 & 3,102 \\
    RVS-Ego / RVS-Movie \cite{zhang2025flashvstream} & V & OE & L & 32 & 3,500 \\
    \bottomrule
  \end{tabular}
\end{table}

\section{VideoLLM Architectures and Computational Bottlenecks}
\label{sec:architectures}

We first review representative VideoLLMs, grouped by four families (short-video chat systems, unified image-video models, long-video and streaming systems, and audiovisual models) which determine where tokens are produced and how many. We then formalize the compute and memory costs of the resulting encoder--connector--LLM pipeline, which Section~\ref{sec:efficiency-mechanisms} uses as its common basis for comparison.

\subsection{Representative VideoLLM Architectures}

\begin{figure}[!t]
  \centering
  \includegraphics[width=\columnwidth]{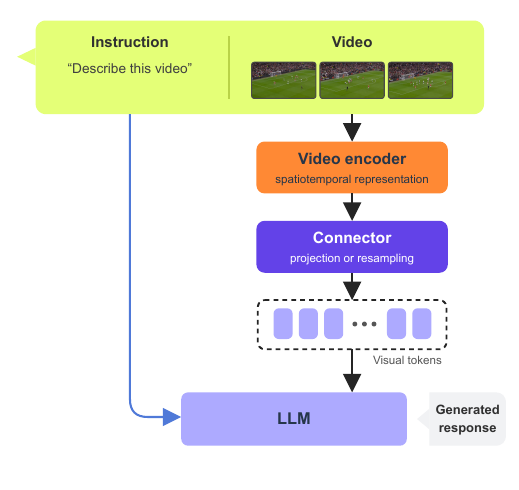}
  \caption{Video Embedder $\times$ LLM paradigm. A video encoder transforms sampled frames into continuous features; a connector projects or compresses them into the LLM token space, where they are combined with a textual prompt for task-conditioned generation. To additionally process audio, audio features (obtained with a separate audio encoder) can also be projected to the input space of the LLM through a different connector.}
  \label{fig:vid_llm_overview}
\end{figure}

Tang \textit{et al.} \cite{tangVideoUnderstandingLarge2023} distinguish three VideoLLM families by how video information reaches the LLM: \textit{Video Analyzer $\times$ LLM} systems convert the video into textual evidence (captions, timestamped events, serialized object tracks, ASR or OCR) before LLM processing; \textit{Video Embedder $\times$ LLM} systems map continuous encoder representations into the LLM input space through a connector; and hybrid \textit{(Analyzer + Embedder) $\times$ LLM} systems provide both. We restrict this survey to the Embedder family (the largest, comprising 79 of the 127 systems Tang \textit{et al.} catalog) because its encoder--connector--LLM structure matches the system boundary of our efficiency analysis: frame sampling, encoder cost, connector compression, multimodal token counts, LLM prefilling and KV-cache behavior. Analyzer-centric and hybrid systems would require accounting for the upstream expert models that produce textual analyses, and fall outside this pipeline-based scope. Figure~\ref{fig:vid_llm_overview} summarizes this framework.

Prompting lets the same backbone serve captioning, question answering, retrieval, temporal grounding and summarization without task-specific heads, so pipeline-level efficiency gains apply across all of them. Within this template, the most representative VideoLLMs differ mainly in their choice of encoders, connectors and language backbones, their target video length, and whether they use audio.

\noindent
\textbf{Short-video VideoLLMs and chat-centric systems.}
A first generation of VideoLLMs extends image-based VLMs (Vision Language Models) to short clips. Video-LLaMA \cite{zhangVideoLLaMAInstructiontunedAudioVisual2023} establishes the canonical pattern: CLIP \cite{radfordLearningTransferableVisual2021} or ViT \cite{dosovitskiyImageWorth16x162021} vision encoders, ImageBind audio features \cite{girdharImageBindOneEmbedding2023}, and a Q-Former connector \cite{liBLIP2BootstrappingLanguageImage2023} mapping both streams into Vicuna tokens \cite{VicunaOpenSourceChatbot}. VideoChat \cite{liVideoChatChatCentricVideo2024} and Valley \cite{luoValleyVideoAssistant2025} add chat-centric instruction tuning, Video-ChatGPT \cite{maazVideoChatGPTDetailedVideo2024} popularizes GPT-based self-instruct training data, and mPLUG/mPLUG-2 \cite{xuYoukumPLUG10Million2023,xuMPLUG2ModularizedMultimodal2023} apply dual-encoder contrastive pretraining to short video QA.

\noindent
\textbf{Unified image-video LLMs.}
A second wave moves to unified image-video models reusing image encoders with sparse frame sampling. The LLaVA family \cite{zhangLLaVAVideoVideoInstruction2025,linVideoLLaVALearningUnited2023} adds temporal pooling, LLaMA-VID \cite{liLLaMAVIDImageWorth2023} compresses each frame to two visual tokens, making long VideoQA feasible, and MiniGPT4-Video \cite{ataallahMiniGPT4VideoAdvancingMultimodal2024} interleaves visual and textual tokens, later serving as the backbone of Goldfish \cite{ataallahGoldfishVisionLanguageUnderstanding2024}. General-purpose VLMs such as Qwen2-VL \cite{wangQwen2VLEnhancingVisionLanguage2024} and InternVL \cite{chenExpandingPerformanceBoundaries2025} adopt the same unified pipeline, and InternVideo2.x \cite{wangInternVideo2ScalingFoundation2024,wangInternVideo25EmpoweringVideo2025} shows that high-capacity video encoders with lightweight connectors compete favorably on MVBench \cite{liMVBenchComprehensiveMultimodal2024} and Video-MME \cite{fuVideoMMEFirstEverComprehensive2025}.

\noindent\textbf{Long-video and streaming VideoLLMs.}
As long-video benchmarks emerged (EgoSchema \cite{mangalamEgoSchemaDiagnosticBenchmark2023}, LongVideoBench \cite{wuLongVideoBenchBenchmarkLongcontext2024}, TVQA-long \cite{ataallahGoldfishVisionLanguageUnderstanding2024}), a third line targeted minute-to-hour contexts under strict limits: hierarchical memory approaches (MovieChat \cite{songMovieChatDenseToken2024}, LongVLM \cite{wengLongVLMEfficientLong2024}, MA-LMM \cite{heMALMMMemoryAugmented2024}) compress visual tokens into multi-scale representations or explicit memory modules; streaming and retrieval methods (VideoStreaming \cite{qianStreamingLongVideo2024}, VideoLLM-online \cite{chenVideoLLMonlineOnlineVideo2024}, VideoLLM-MoD \cite{wuVideoLLMMoDEfficientVideoLanguage2024}, Goldfish \cite{ataallahGoldfishVisionLanguageUnderstanding2024}) maintain constant token budgets; $\infty$-Video \cite{santos$infty$VideoTrainingFreeApproach2025} adds training-free long-term memory and frame selection around existing VideoLLMs \cite{zhangVideoLLaMAInstructiontunedAudioVisual2023,liMVBenchComprehensiveMultimodal2024}; and the VideoChat family refines temporal encoding, reinforcement tuning for grounding, and multi-agent planning (VideoChat-T \cite{zengTimeSuiteImprovingMLLMs2024}, VideoChat-R1 \cite{liVideoChatR1EnhancingSpatioTemporal2025}, VideoChat-M1 \cite{chenVideoChatM1CollaborativePolicy2025}).

\noindent\textbf{Audiovisual VideoLLMs.}
Audiovisual VideoLLMs keep the same encoder--connector--LLM template while adding synchronized audio. Video-LLaMA maps ImageBind audio and ViT video features into Vicuna through separate Q-Formers; VideoLLaMA~2 \cite{chengVideoLLaMA2Advancing2024} replaces this interface with spatial-temporal convolution connectors; and recent systems such as Qwen2.5-Omni \cite{xuQwen25OmniTechnicalReport2025} and OmniVinci \cite{yeOmniVinciEnhancingArchitecture2025} use dedicated visual and audio encoders with learned temporal alignment before a shared language core. These architectures add an audiovisual dimension to the taxonomy: audio adds an encoder and token stream, but the efficiency question remains how much encoded evidence reaches the LLM and at what cost.

\subsection{Sources of Computational Cost and Architectural Bottlenecks}
\label{subsec:bottlenecks}

We now formalize the dominant compute and memory scaling factors of the encoder--connector--LLM pipeline. Frame count and resolution determine encoder cost and the number of modality tokens produced; connector compression controls how many of those tokens enter the LLM; and the resulting context length determines LLM prefilling cost and KV-cache memory during decoding.

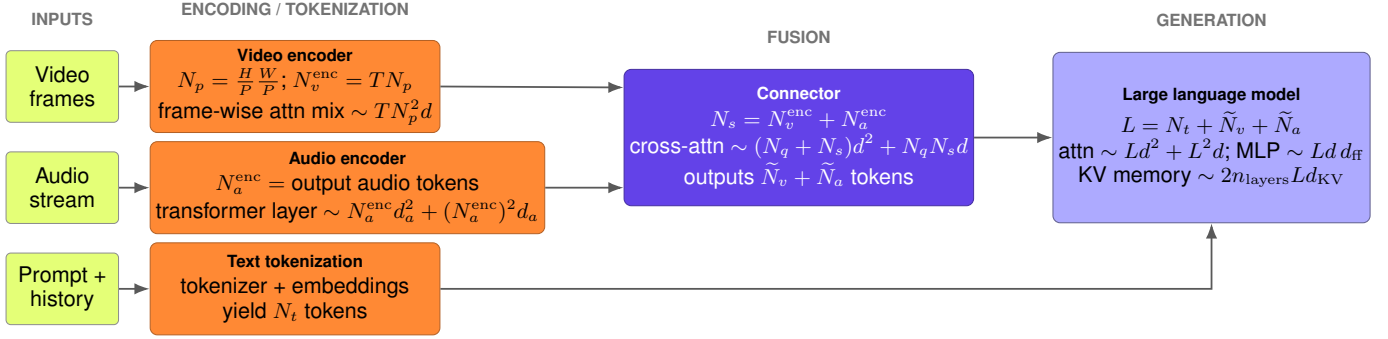
\begin{figure*}[!ht]
  \centering
  \scriptsize
  \definecolor{costInput}{HTML}{E4FF77}
  \definecolor{costEncoder}{HTML}{FF8934}
  \definecolor{costConnector}{HTML}{6342E8}
  \definecolor{costLLM}{HTML}{ADAAFF}
  \resizebox{\textwidth}{!}{%
  \begin{tikzpicture}[
      font=\sffamily,
      >=Latex,
      input/.style={rounded corners=2pt, draw=black!45, fill=costInput,
        minimum width=1.65cm, minimum height=1.05cm, align=center, font=\sffamily\small},
      encoder/.style={rounded corners=3pt, draw=costEncoder!70!black, fill=costEncoder,
        minimum width=4.25cm, minimum height=1.35cm, align=center},
      connector/.style={rounded corners=3pt, draw=costConnector!70!black, fill=costConnector,
        text=white, minimum width=4.0cm, minimum height=2.0cm, align=center},
      llm/.style={rounded corners=3pt, draw=costLLM!55!black, fill=costLLM,
        minimum width=4.1cm, minimum height=2.5cm, align=center},
      flow/.style={->, line width=0.75pt, draw=black!65},
      heading/.style={font=\sffamily\scriptsize\bfseries, text=black!55}
    ]
    \node[input] (video-in) {Video\\frames};
    \node[input, below=0.42cm of video-in] (audio-in) {Audio\\stream};
    \node[input, below=0.42cm of audio-in] (text-in) {Prompt +\\history};

    \node[encoder, right=0.45cm of video-in] (video) {\textbf{Video encoder}\\[3pt]
      \small $N_p=\frac{H}{P}\frac{W}{P}$; $N_v^{\mathrm{enc}}=TN_p$\\[2pt]
      \small frame-wise attn mix $\sim TN_p^2d$};
    \node[encoder, right=0.45cm of audio-in] (audio) {\textbf{Audio encoder}\\[3pt]
      \small $N_a^{\mathrm{enc}}=\text{output audio tokens}$\\[2pt]
      \small transformer layer $\sim N_a^{\mathrm{enc}}d_a^2+(N_a^{\mathrm{enc}})^2d_a$};
    \node[encoder, right=0.45cm of text-in] (text) {\textbf{Text tokenization}\\[3pt]
      \small tokenizer + embeddings\\[2pt]
      \small yield $N_t$ tokens};

    \node[connector, right=1.15cm of audio, yshift=0.735cm] (connector) {\textbf{Connector}\\[3pt]
      \small $N_s=N_v^{\mathrm{enc}}+N_a^{\mathrm{enc}}$\\[2pt]
      \small cross-attn $\sim (N_q+N_s)d^2+N_qN_sd$\\[2pt]
      \small outputs $\widetilde N_v+\widetilde N_a$ tokens};
    \node[llm, right=1.15cm of connector] (llm) {\textbf{Large language model}\\[3pt]
      \small $L=N_t+\widetilde N_v+\widetilde N_a$\\[2pt]
      \small attn $\sim Ld^2+L^2d$; MLP $\sim Ld\,d_{\mathrm{ff}}$\\[2pt]
      \small KV memory $\sim 2n_{\mathrm{layers}}Ld_{\mathrm{KV}}$};

    \node[heading, above=0.28cm of video-in] {INPUTS};
    \node[heading, above=0.28cm of video] {ENCODING / TOKENIZATION};
    \node[heading, above=0.28cm of connector] {FUSION};
    \node[heading, above=0.28cm of llm] {GENERATION};

    \draw[flow] (video-in) -- (video);
    \draw[flow] (audio-in) -- (audio);
    \draw[flow] (text-in) -- (text);
    \draw[flow] (video.east) -- ([yshift=0.735cm]connector.west);
    \draw[flow] (audio.east) -- ([yshift=-0.735cm]connector.west);
    \draw[flow] (connector) -- (llm);
    \draw[flow] (text.east) -- ++(0.55cm,0) -| (llm.south);

  \end{tikzpicture}
  }
  \caption{Token and compute scaling across the encoder--connector--LLM pipeline. Video and audio encoders produce $N_v^{\mathrm{enc}}$ and $N_a^{\mathrm{enc}}$ tokens; the connector retains $\widetilde N_v$ and $\widetilde N_a$, while tokenized prompts and history contribute $N_t$. LLM compute and KV memory scale with the resulting context $L=N_t+\widetilde N_v+\widetilde N_a$.}
  \label{fig:efficiency_bottlenecks}
\end{figure*}

We denote by $T$ the number of video frames fed to the encoder, by $H\times W$ the spatial resolution of each frame, and by $P\times P$ the patch size used by a frame-wise ViT encoder. The number of spatial patches per frame is
$
N_p \;=\; \frac{H}{P}\cdot\frac{W}{P},
$
so the encoder initially produces
$
N_v^{\mathrm{enc}} \;=\; T \cdot N_p \;=\; T \cdot \frac{H}{P}\cdot\frac{W}{P}
$
(up to special tokens). For video transformers using temporal tubelets of length $\tau$, $T$ is replaced by $T/\tau$. We write $N_a^{\mathrm{enc}}$ for the audio-encoder output length and $\widetilde N_v,\widetilde N_a$ for the visual and audio token counts retained after connector-side pooling, projection or resampling. With $N_t$ text tokens (prompt, history and any previously generated tokens), the LLM context length is
$
L \;=\; N_t + \widetilde N_v + \widetilde N_a.
$
For a joint Q-Former that replaces both modality streams with $N_q$ query outputs, the corresponding context is $L=N_t+N_q$.
A transformer block's hidden width is its per-token embedding dimension: $d_v$ for the video encoder, $d_a$ for the audio encoder and $d$ for the LLM; $d_{\mathrm{ff}}$ and $d_{\mathrm{ff},a}$ are the corresponding feed-forward widths, and $d_{\mathrm{KV}}$ the total key/value width stored per token. The generic transformer-layer expressions below use $N$ and $d$ for the token count and width of the block in question.
Figure~\ref{fig:efficiency_bottlenecks} summarizes these bottlenecks visually.

\noindent
\textbf{Video and modality encoders.}
For a fixed 2D CNN (Convolutional Neural Network) applied frame-wise, encoder cost scales as
$
\text{FLOPs}_\text{enc}^{\text{2D}} = T \cdot C_{\mathrm{frame}}(H,W),
$
where $C_{\mathrm{frame}}(H,W)$ is the cost of one pass through the chosen backbone; thus cost is linear in $T$ at fixed resolution and architecture. 3D CNNs and video transformers add temporal interactions. For a transformer layer processing a sequence of $N$ tokens, the attention and MLP (Multi-Layer Perceptron) costs scale as
$
\text{FLOPs}_\text{attn} \propto N \cdot d^2 + N^2 \cdot d,
\qquad
\text{FLOPs}_\text{mlp} \propto N \cdot d \cdot d_{\mathrm{ff}}.
$

For a frame-wise ViT, the attention-mixing term summed across frames is $T N_p^2d$; only full joint space--time attention incurs $(N_v^{\mathrm{enc}})^2d$, while factorized architectures lie between these regimes. Increasing the frame count or spatial resolution nevertheless inflates encoder cost. Long-video VideoLLMs often process hundreds of frames or minute-long clips via sliding windows or dense sampling, so the encoder alone can dominate total cost unless frames are subsampled or pooled.

Audio encoders usually begin from a denser temporal signal than sparsely sampled video, but their output length and cost depend strongly on convolutional stride, pooling and architecture. For a transformer layer operating on $N_a^{\mathrm{enc}}$ audio tokens,
$
\text{FLOPs}_{\text{audio,attn}} \propto N_a^{\mathrm{enc}} \cdot d_a^2 + (N_a^{\mathrm{enc}})^2 \cdot d_a,
$
with a further $N_a^{\mathrm{enc}} d_a d_{\mathrm{ff},a}$ contribution from the MLP; convolutional front ends have architecture-specific costs. Audio may be negligible after aggressive downsampling or material in long-form audiovisual inputs; it cannot be ranked against the visual stream from sampling rates alone because each video frame produces many spatial patch tokens. Additional ASR, OCR or subtitle-processing modules likewise add costs that should be reported separately \cite{nguyenVideoLanguageUnderstandingSurvey2024,zouSecondsHoursReviewing2024}.

\noindent
\textbf{Connectors and cross-modal fusion.}
Connectors project high-dimensional spatiotemporal features (visual and audio tokens) into the LLM token space. In the simplest case, visual and audio tokens are flattened and passed through linear layers or small MLPs, yielding a cost
$
\text{FLOPs}_\text{proj} \propto N_v^{\mathrm{enc}}d_v d + N_a^{\mathrm{enc}}d_a d.
$
More sophisticated connectors, such as Q-Former \cite{liBLIP2BootstrappingLanguageImage2023} or cross-attention modules, use a set of $N_q$ learnable query tokens attending over $N_s=N_v^{\mathrm{enc}}+N_a^{\mathrm{enc}}$ source tokens. Including query, key, value and output projections, the cross-attention cost per layer scales as
$
\text{FLOPs}_\text{cross-attn} \propto (N_q+N_s)d^2 + N_qN_sd.
$
Although $N_q$ is usually small, the source sequence $N_s$ can still be large. Many VideoLLMs therefore apply temporal or spatial pooling, audio downsampling, or selective token fusion before cross-attention, often enforcing a fixed joint token budget \cite{tangVideoUnderstandingLarge2023,chenSurveyOmnimodalLanguage2025}.

\noindent
\textbf{LLM context length and KV cache.}
Once projected, the retained visual and audio tokens are concatenated (or interleaved) with textual tokens and processed by the LLM. In a standard transformer layer, self-attention over $L=N_t+\widetilde N_v+\widetilde N_a$ tokens has cost
$
\text{FLOPs}_\text{LLM,attn} \propto Ld^2 + L^2d,
$
while feed-forward blocks contribute
$
\text{FLOPs}_\text{LLM,mlp} \propto Ld\,d_{\mathrm{ff}}.
$
Whether the quadratic attention term or the linear feed-forward term dominates depends on the hidden width and the retained token counts. Among those tokens, visual patches usually outnumber the rest before compression, while audio can become substantial in long audiovisual inputs. Autoregressive decoding also stores key-value (KV) caches for each layer, with memory scaling
$
\text{Mem}_\text{KV} \propto 2B \cdot n_{\mathrm{layers}} \cdot L \cdot d_{\mathrm{KV}} \cdot b,
$
where $B$ is the batch size, $n_{\mathrm{layers}}$ the number of decoder layers and $b$ the bytes per stored element; grouped- and multi-query attention reduce $d_{\mathrm{KV}}$. This limits feasible context length for multi-turn dialogue grounded in long videos, especially when audio, OCR or subtitles share the same context window \cite{zouSecondsHoursReviewing2024,hanLongInsightBenchComprehensiveBenchmark2025}.

The dominant regime also changes between prefilling and autoregressive decoding. Attention-score computation during prefilling is quadratic in $L$, although kernels, hardware and the linear-in-$L$ projection and feed-forward terms determine whether execution is actually compute-bound. At each decoding step, attention mixing over the cached prefix costs $O(Ld)$ per layer, alongside $O(d^2)$ projection and feed-forward work, and is often constrained by memory traffic~\cite{daoFlashAttentionFastMemoryEfficient2022}. Reducing input tokens therefore benefits both stages, while KV-cache compression primarily targets decoding memory; their relative impact varies across interactive, batch and offline workloads.

Because the relative importance of encoding, connector token count and LLM prefill, decoding and cache growth is architecture- and workload-dependent, the next section organizes methods by the pipeline stage at which they reduce cost.

\section{Taxonomy of Efficiency Mechanisms in VideoLLMs}
\label{sec:efficiency-mechanisms}

% AUTO-GENERATED by figures/gen_taxonomy.py from tables/taxonomy_families.tex -- edit those, not this file.
% Cards are
% positioned relative to each other (rows below rows, bands fitted to cards);
% card boxes are drawn on the background layer after each row is measured so
% every card in a row shares the row height. Requires tikzlibrary fit,backgrounds.
% Markers: $^a$ evaluated outside a VideoLLM, $^+$ reduces cost at several stages.

\definecolor{taxonomy-frame}{HTML}{E4FF77}
\definecolor{taxonomy-encoder}{HTML}{FF8934}
\definecolor{taxonomy-connector}{HTML}{6342E8}
\definecolor{taxonomy-llm}{HTML}{ADAAFF}
\definecolor{taxonomy-border}{RGB}{70,70,70}

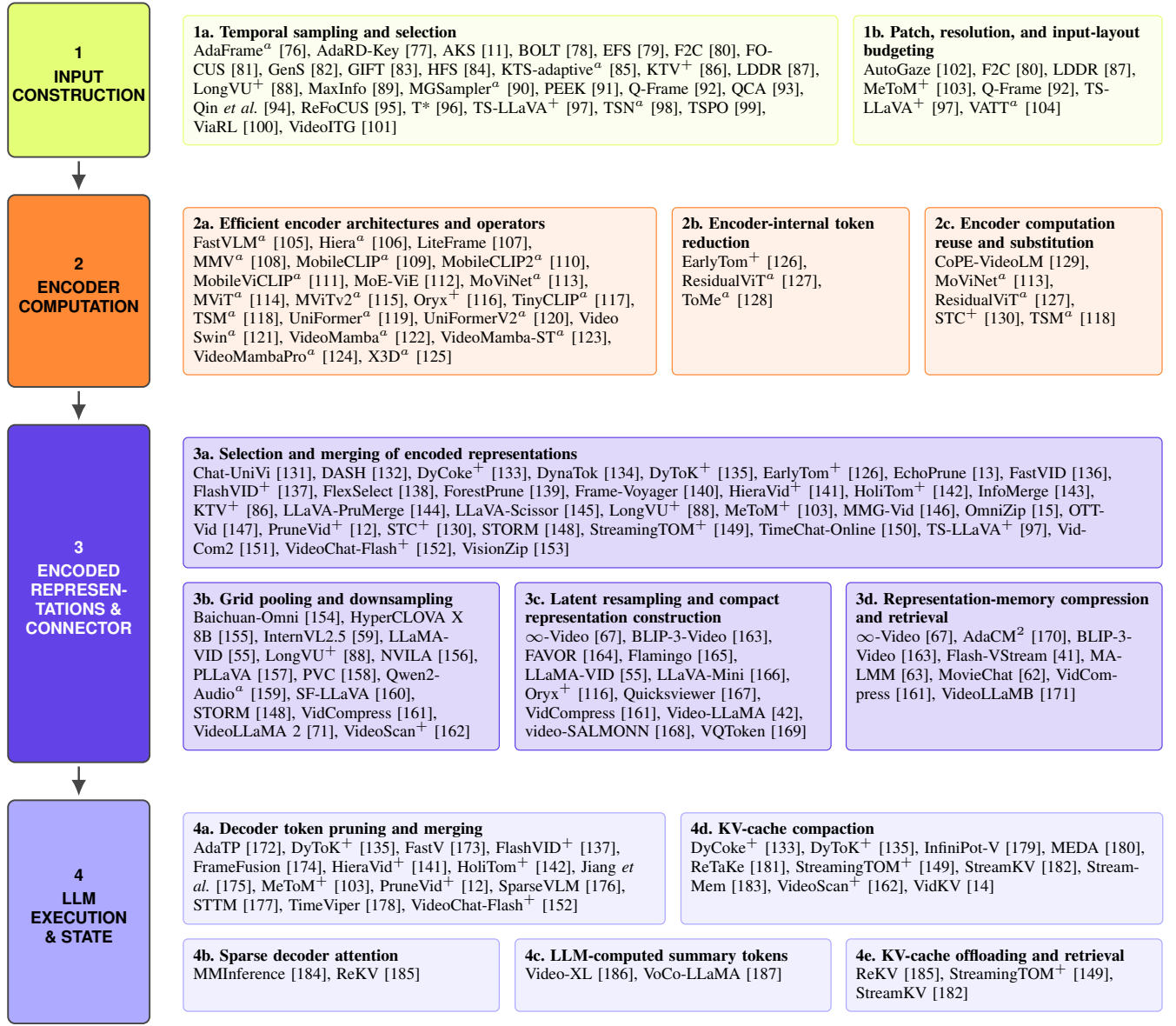
\begin{figure*}[!tp]
  \centering
  \resizebox{0.98\textwidth}{!}{%
    \begin{tikzpicture}[
      stage/.style={draw=taxonomy-border, rounded corners=3pt, line width=0.7pt, inner sep=0pt,
        text width=2.1cm, align=center, font=\sffamily\bfseries\scriptsize},
      card/.style={anchor=north west, inner sep=4pt, align=left, font=\scriptsize},
      cardbox/.style={draw=taxonomy-border, rounded corners=2pt, line width=0.45pt, inner sep=0pt}
    ]
      % Stage 1.
      \node[card, text width=9.41cm] (1a) at (2.60,0)
        {\textbf{1a. Temporal sampling and selection}\\
         AdaFrame$^{a}$~\cite{wuAdaFrameAdaptiveFrame2018}, AdaRD-Key~\cite{zhangAdaRDkeyAdaptiveRelevanceDiversity2025}, AKS~\cite{tangAdaptiveKeyframeSampling2025}, BOLT~\cite{liu2025bolt}, EFS~\cite{chen2026efs}, F2C~\cite{sunFramesClipsTrainingfree2025}, FOCUS~\cite{zhuFOCUSEfficientKeyframe2025}, GenS~\cite{yaoGenerativeFrameSampler2025}, GIFT~\cite{ma2026gift}, HFS~\cite{yangHFSHolisticQueryAware2025}, KTS-adaptive$^{a}$~\cite{afhamRevisitingKernelTemporal2023}, KTV$^{+}$~\cite{song2026ktv}, LDDR~\cite{chen2026lddr}, LongVU$^{+}$~\cite{shenLongVUSpatiotemporalAdaptive2024}, MaxInfo~\cite{liMaxInfoTrainingFreeKeyFrame2025}, MGSampler$^{a}$~\cite{zhiMGSamplerExplainableSampling2021}, PEEK~\cite{steunou2026peek}, Q-Frame~\cite{zhangQFrameQueryawareFrame2025}, QCA~\cite{peng2026qca}, Qin \textit{et al.}~\cite{qin2026efficient}, ReFoCUS~\cite{lee2025refocus}, T*~\cite{ye2025tstar}, TS-LLaVA$^{+}$~\cite{qu2024tsllava}, TSN$^{a}$~\cite{wangTemporalSegmentNetworks2016}, TSPO~\cite{tang2025tspo}, ViaRL~\cite{xu2025viarl}, VideoITG~\cite{wang2025videoitg}};
      \node[card, text width=4.28cm] (1b) at (12.54,0)
        {\textbf{1b. Patch, resolution, and input-layout budgeting}\\
         AutoGaze~\cite{shi2026autogaze}, F2C~\cite{sunFramesClipsTrainingfree2025}, LDDR~\cite{chen2026lddr}, MeToM$^{+}$~\cite{wu2026metom}, Q-Frame~\cite{zhangQFrameQueryawareFrame2025}, TS-LLaVA$^{+}$~\cite{qu2024tsllava}, VATT$^{a}$~\cite{akbariVATTTransformersMultimodal2021}};
      \node[fit=(1a)(1b), inner sep=0] (row10) {};
      \begin{scope}[on background layer]
        \node[cardbox, draw=taxonomy-frame, fill=taxonomy-frame!12, fit={(1a.north west) (1a.south east |- row10.south)}] {};
        \node[cardbox, draw=taxonomy-frame, fill=taxonomy-frame!12, fit={(1b.north west) (1b.south east |- row10.south)}] {};
      \end{scope}
      \node[fit=(1a)(1b), inner sep=0.18cm] (band1) {};
      \node[stage, fill=taxonomy-frame, fit={(0,0 |- band1.north) (2.1cm,0 |- band1.south)}] (stage1)
        {1\\[2pt]INPUT\\CONSTRUCTION};
      % Stage 2.
      \node[card, text width=6.72cm] (2a) at ([yshift=-0.75cm]band1.south -| 2.60,0)
        {\textbf{2a. Efficient encoder architectures and operators}\\
         FastVLM$^{a}$~\cite{vasuFastVLMEfficient2024}, Hiera$^{a}$~\cite{ryaliHieraHierarchicalVision2023}, LiteFrame~\cite{kimLiteFrameEfficient2026}, MMV$^{a}$~\cite{alayracSelfSupervisedMultiModalVersatile2020}, MobileCLIP$^{a}$~\cite{vasuMobileCLIPFast2023}, MobileCLIP2$^{a}$~\cite{faghri2025mobileclip2}, MobileViCLIP$^{a}$~\cite{yangMobileViCLIPEfficientVideoText2025}, MoE-ViE~\cite{zhang2026moevie}, MoViNet$^{a}$~\cite{kondratyukMoViNetsMobileVideo2021}, MViT$^{a}$~\cite{fanMultiscaleVisionTransformers2021}, MViTv2$^{a}$~\cite{liMViTv2ImprovedMultiscale2022}, Oryx$^{+}$~\cite{liu2024oryx}, TinyCLIP$^{a}$~\cite{wuTinyCLIPCLIP2023}, TSM$^{a}$~\cite{linTSMTemporalShift2019}, UniFormer$^{a}$~\cite{liUniFormerUnifiedTransformer2022}, UniFormerV2$^{a}$~\cite{liUniFormerV2Spatiotemporal2023}, Video Swin$^{a}$~\cite{liuVideoSwinTransformer2021}, VideoMamba$^{a}$~\cite{liVideoMambaState2024}, VideoMamba-ST$^{a}$~\cite{parkVideoMambaSpatio2024}, VideoMambaPro$^{a}$~\cite{luSnakesLadders2025}, X3D$^{a}$~\cite{feichtenhoferX3DExpandingArchitectures2020}};
      \node[card, text width=3.22cm] (2b) at ([yshift=-0.75cm]band1.south -| 9.85,0)
        {\textbf{2b. Encoder-internal token reduction}\\
         EarlyTom$^{+}$~\cite{wang2026earlytom}, ResidualViT$^{a}$~\cite{soldanResidualViT2025}, ToMe$^{a}$~\cite{bolyaTokenMergingYour2023}};
      \node[card, text width=3.22cm] (2c) at ([yshift=-0.75cm]band1.south -| 13.60,0)
        {\textbf{2c. Encoder computation reuse and substitution}\\
         CoPE-VideoLM~\cite{sarkar2026cope}, MoViNet$^{a}$~\cite{kondratyukMoViNetsMobileVideo2021}, ResidualViT$^{a}$~\cite{soldanResidualViT2025}, STC$^{+}$~\cite{wang2025stc}, TSM$^{a}$~\cite{linTSMTemporalShift2019}};
      \node[fit=(2a)(2b)(2c), inner sep=0] (row20) {};
      \begin{scope}[on background layer]
        \node[cardbox, draw=taxonomy-encoder, fill=taxonomy-encoder!12, fit={(2a.north west) (2a.south east |- row20.south)}] {};
        \node[cardbox, draw=taxonomy-encoder, fill=taxonomy-encoder!12, fit={(2b.north west) (2b.south east |- row20.south)}] {};
        \node[cardbox, draw=taxonomy-encoder, fill=taxonomy-encoder!12, fit={(2c.north west) (2c.south east |- row20.south)}] {};
      \end{scope}
      \node[fit=(2a)(2b)(2c), inner sep=0.18cm] (band2) {};
      \node[stage, fill=taxonomy-encoder, fit={(0,0 |- band2.north) (2.1cm,0 |- band2.south)}] (stage2)
        {2\\[2pt]ENCODER\\COMPUTATION};
      % Stage 3.
      \node[card, text width=14.22cm] (3a) at ([yshift=-0.75cm]band2.south -| 2.60,0)
        {\textbf{3a. Selection and merging of encoded representations}\\
         Chat-UniVi~\cite{jinChatUniViUnifiedVisual2024}, DASH~\cite{li2026dash}, DyCoke$^{+}$~\cite{tao2024dycoke}, DynaTok~\cite{park2026dynatok}, DyToK$^{+}$~\cite{li2025dytok}, EarlyTom$^{+}$~\cite{wang2026earlytom}, EchoPrune~\cite{li2026echoprune}, FastVID~\cite{shen2025fastvid}, FlashVID$^{+}$~\cite{fan2026flashvid}, FlexSelect~\cite{zhang2025flexselect}, ForestPrune~\cite{ju2026forestprune}, Frame-Voyager~\cite{yuFrameVoyagerLearningQuery2025}, HieraVid$^{+}$~\cite{hieravid2025}, HoliTom$^{+}$~\cite{shao2025holitom}, InfoMerge~\cite{liu2026infomerge}, KTV$^{+}$~\cite{song2026ktv}, LLaVA-PruMerge~\cite{shang2024prumerge}, LLaVA-Scissor~\cite{sun2025llavascissor}, LongVU$^{+}$~\cite{shenLongVUSpatiotemporalAdaptive2024}, MeToM$^{+}$~\cite{wu2026metom}, MMG-Vid~\cite{ma2025mmgvid}, OmniZip~\cite{taoOmniZipAudioGuidedDynamic2025}, OTT-Vid~\cite{kang2026ottvid}, PruneVid$^{+}$~\cite{huang-etal-2025-prunevid}, STC$^{+}$~\cite{wang2025stc}, STORM~\cite{jiangSTORMTokenEfficientLong2025}, StreamingTOM$^{+}$~\cite{streamingTOM2025}, TimeChat-Online~\cite{yao2025timechatonline}, TS-LLaVA$^{+}$~\cite{qu2024tsllava}, VidCom2~\cite{liu2025vidcom2}, VideoChat-Flash$^{+}$~\cite{li2025videochatflash}, VisionZip~\cite{yang2024visionzip}};
      \node[fit=(3a), inner sep=0] (row30) {};
      \begin{scope}[on background layer]
        \node[cardbox, draw=taxonomy-connector, fill=taxonomy-connector!20, fit={(3a.north west) (3a.south east |- row30.south)}] {};
      \end{scope}
      \node[card, text width=4.39cm] (3b) at ([yshift=-0.22cm]row30.south -| 2.60,0)
        {\textbf{3b. Grid pooling and downsampling}\\
         Baichuan-Omni~\cite{liBaichuanOmniTechnicalReport2024}, HyperCLOVA X 8B~\cite{hyperclovaxteamHyperCLOVA8BOmni2026}, InternVL2.5~\cite{chenExpandingPerformanceBoundaries2025}, LLaMA-VID~\cite{liLLaMAVIDImageWorth2023}, LongVU$^{+}$~\cite{shenLongVUSpatiotemporalAdaptive2024}, NVILA~\cite{liu2024nvila}, PLLaVA~\cite{xuPLLaVAParameterfreeLLaVA2024}, PVC~\cite{yang2024pvc}, Qwen2-Audio$^{a}$~\cite{chuQwen2AudioTechnicalReport2024}, SF-LLaVA~\cite{xu2024slowfastllava}, STORM~\cite{jiangSTORMTokenEfficientLong2025}, VidCompress~\cite{lan2024vidcompress}, VideoLLaMA~2~\cite{chengVideoLLaMA2Advancing2024}, VideoScan$^{+}$~\cite{li2025videoscan}};
      \node[card, text width=4.39cm] (3c) at ([yshift=-0.22cm]row30.south -| 7.52,0)
        {\textbf{3c. Latent resampling and compact representation construction}\\
         $\infty$-Video~\cite{santos$infty$VideoTrainingFreeApproach2025}, BLIP-3-Video~\cite{ryoo2024blip3video}, FAVOR~\cite{sunFinegrainedAudioVisualJoint2023}, Flamingo~\cite{alayracFlamingoVisualLanguage2022}, LLaMA-VID~\cite{liLLaMAVIDImageWorth2023}, LLaVA-Mini~\cite{zhangLLaVAMiniEfficientImage2025}, Oryx$^{+}$~\cite{liu2024oryx}, Quicksviewer~\cite{qi2025quicksviewer}, VidCompress~\cite{lan2024vidcompress}, Video-LLaMA~\cite{zhangVideoLLaMAInstructiontunedAudioVisual2023}, video-SALMONN~\cite{sunVideoSALMONNSpeechEnhancedAudioVisual2024}, VQToken~\cite{zhang2025vqtoken}};
      \node[card, text width=4.39cm] (3d) at ([yshift=-0.22cm]row30.south -| 12.43,0)
        {\textbf{3d. Representation-memory compression and retrieval}\\
         $\infty$-Video~\cite{santos$infty$VideoTrainingFreeApproach2025}, AdaCM$^2$~\cite{man2025adacm2}, BLIP-3-Video~\cite{ryoo2024blip3video}, Flash-VStream~\cite{zhang2025flashvstream}, MA-LMM~\cite{heMALMMMemoryAugmented2024}, MovieChat~\cite{songMovieChatDenseToken2024}, VidCompress~\cite{lan2024vidcompress}, VideoLLaMB~\cite{wang2025videollamb}};
      \node[fit=(3b)(3c)(3d), inner sep=0] (row31) {};
      \begin{scope}[on background layer]
        \node[cardbox, draw=taxonomy-connector, fill=taxonomy-connector!20, fit={(3b.north west) (3b.south east |- row31.south)}] {};
        \node[cardbox, draw=taxonomy-connector, fill=taxonomy-connector!20, fit={(3c.north west) (3c.south east |- row31.south)}] {};
        \node[cardbox, draw=taxonomy-connector, fill=taxonomy-connector!20, fit={(3d.north west) (3d.south east |- row31.south)}] {};
      \end{scope}
      \node[fit=(3a)(3b)(3c)(3d), inner sep=0.18cm] (band3) {};
      \node[stage, fill=taxonomy-connector, text=white, fit={(0,0 |- band3.north) (2.1cm,0 |- band3.south)}] (stage3)
        {3\\[2pt]ENCODED\\REPRESEN-\\TATIONS \&\\CONNECTOR};
      % Stage 4.
      \node[card, text width=6.85cm] (4a) at ([yshift=-0.75cm]band3.south -| 2.60,0)
        {\textbf{4a. Decoder token pruning and merging}\\
         AdaTP~\cite{sun-etal-2025-adatp}, DyToK$^{+}$~\cite{li2025dytok}, FastV~\cite{chen2024fastv}, FlashVID$^{+}$~\cite{fan2026flashvid}, FrameFusion~\cite{fu2025framefusion}, HieraVid$^{+}$~\cite{hieravid2025}, HoliTom$^{+}$~\cite{shao2025holitom}, Jiang \textit{et al.}~\cite{jiang2026stateful}, MeToM$^{+}$~\cite{wu2026metom}, PruneVid$^{+}$~\cite{huang-etal-2025-prunevid}, SparseVLM~\cite{zhang2024sparsevlm}, STTM~\cite{hyun2025sttm}, TimeViper~\cite{xu2025timeviper}, VideoChat-Flash$^{+}$~\cite{li2025videochatflash}};
      \node[card, text width=6.85cm] (4d) at ([yshift=-0.75cm]band3.south -| 9.98,0)
        {\textbf{4d. KV-cache compaction}\\
         DyCoke$^{+}$~\cite{tao2024dycoke}, DyToK$^{+}$~\cite{li2025dytok}, InfiniPot-V~\cite{kim2025infinipotv}, MEDA~\cite{wan2025meda}, ReTaKe~\cite{wang2024retake}, StreamingTOM$^{+}$~\cite{streamingTOM2025}, StreamKV~\cite{chen2025streamkv}, StreamMem~\cite{yang2025streammem}, VideoScan$^{+}$~\cite{li2025videoscan}, VidKV~\cite{tao2025vidkv}};
      \node[fit=(4a)(4d), inner sep=0] (row40) {};
      \begin{scope}[on background layer]
        \node[cardbox, draw=taxonomy-llm, fill=taxonomy-llm!18, fit={(4a.north west) (4a.south east |- row40.south)}] {};
        \node[cardbox, draw=taxonomy-llm, fill=taxonomy-llm!18, fit={(4d.north west) (4d.south east |- row40.south)}] {};
      \end{scope}
      \node[card, text width=4.39cm] (4b) at ([yshift=-0.22cm]row40.south -| 2.60,0)
        {\textbf{4b. Sparse decoder attention}\\
         MMInference~\cite{li2025mminference}, ReKV~\cite{di2025rekv}};
      \node[card, text width=4.39cm] (4c) at ([yshift=-0.22cm]row40.south -| 7.52,0)
        {\textbf{4c. LLM-computed summary tokens}\\
         Video-XL~\cite{shu2024videoxl}, VoCo-LLaMA~\cite{yeVoCoLLaMAVisionCompression2024}};
      \node[card, text width=4.39cm] (4e) at ([yshift=-0.22cm]row40.south -| 12.43,0)
        {\textbf{4e. KV-cache offloading and retrieval}\\
         ReKV~\cite{di2025rekv}, StreamingTOM$^{+}$~\cite{streamingTOM2025}, StreamKV~\cite{chen2025streamkv}};
      \node[fit=(4b)(4c)(4e), inner sep=0] (row41) {};
      \begin{scope}[on background layer]
        \node[cardbox, draw=taxonomy-llm, fill=taxonomy-llm!18, fit={(4b.north west) (4b.south east |- row41.south)}] {};
        \node[cardbox, draw=taxonomy-llm, fill=taxonomy-llm!18, fit={(4c.north west) (4c.south east |- row41.south)}] {};
        \node[cardbox, draw=taxonomy-llm, fill=taxonomy-llm!18, fit={(4e.north west) (4e.south east |- row41.south)}] {};
      \end{scope}
      \node[fit=(4a)(4d)(4b)(4c)(4e), inner sep=0.18cm] (band4) {};
      \node[stage, fill=taxonomy-llm, fit={(0,0 |- band4.north) (2.1cm,0 |- band4.south)}] (stage4)
        {4\\[2pt]LLM\\EXECUTION\\\& STATE};
      % Pipeline direction.
      \draw[-{Latex[length=2.4mm, width=2.2mm]}, taxonomy-border, line width=0.9pt]
        ([yshift=-0.05cm]stage1.south -| 1.050cm,0) -- ([yshift=0.05cm]stage2.north -| 1.050cm,0);
      \draw[-{Latex[length=2.4mm, width=2.2mm]}, taxonomy-border, line width=0.9pt]
        ([yshift=-0.05cm]stage2.south -| 1.050cm,0) -- ([yshift=0.05cm]stage3.north -| 1.050cm,0);
      \draw[-{Latex[length=2.4mm, width=2.2mm]}, taxonomy-border, line width=0.9pt]
        ([yshift=-0.05cm]stage3.south -| 1.050cm,0) -- ([yshift=0.05cm]stage4.north -| 1.050cm,0);
    \end{tikzpicture}%
  }
  \caption{Taxonomy of efficiency mechanisms, organized by where each acts in the encoder--connector--LLM pipeline. $^a$ marks encoder or sampling methods evaluated outside a VideoLLM, on recognition or retrieval tasks; $^+$ marks methods that reduce cost at several stages and appear at each.}
  \label{fig:efficiency_taxonomy}
\end{figure*}

We analyze efficiency using the encoder--connector--LLM decomposition introduced in Section~\ref{sec:architectures}. After defining the reporting conventions used in this survey, we organize mechanisms by the pipeline stage at which they act: input construction and selection, encoder computation, encoded representations and connector, and LLM execution and state. Figure~\ref{fig:efficiency_taxonomy} summarizes the taxonomy; audiovisual methods are included when audio compression or audio-guided selection directly reduces VideoLLM inference cost.

\subsection{Comparison Protocol}
\label{subsec:efficiency-metrics}

As reviewed in Section~\ref{subsec:bottlenecks}, VideoLLM papers mix analytical indicators and system-level measurements under heterogeneous assumptions about video length, resolution, modality coverage and hardware; we compare results only when their measurement scope and input protocol are explicit. Analytical, hardware-independent indicators include parameter count, FLOPs per input and retained-token count or ratio (we reserve FLOP/s for rate-based throughput). Their scope depends on the pipeline stage: vision-encoder comparisons report encoder size and GFLOPs together with the clip configuration \cite{feichtenhoferX3DExpandingArchitectures2020,fanMultiscaleVisionTransformers2021}, whereas connector and token-reduction comparisons report host-LLM size, and accounting boundaries differ even then: HoliTom~\cite{shao2025holitom} and HieraVid~\cite{hieravid2025} report LLM prefilling FLOPs while EarlyTom includes vision-encoder FLOPs \cite{wang2026earlytom}. For audiovisual systems, whose video and audio encoders can have very different profiles and whose connectors range from linear projections to Q-Formers \cite{liBLIP2BootstrappingLanguageImage2023} or Perceiver resamplers \cite{alayracFlamingoVisualLanguage2022}, we distinguish encoder, connector and LLM costs whenever the source provides them \cite{jinEfficientMultimodalLarge2024}.

Analytical indicators do not necessarily predict runtime performance, because operator mix, parallelism, memory access and implementation determine measured speed \cite{dehghani2022efficiencymisnomer,reddiMLPerfInferenceBenchmark2020}. We therefore distinguish them from system-level measurements (wall-clock latency, throughput, peak memory), whose interpretation depends on batch size, sequence length, precision, device and software stack \cite{reddiMLPerfInferenceBenchmark2020}. Offline methods report end-to-end or stage-specific latency and memory; streaming systems additionally report processing rate or response latency together with bounded memory as the stream grows \cite{chatterjeeMemoryefficientStreamingVideoLLMs2025,streamingTOM2025}. Energy is a relevant system metric \cite{tschandMLPerfPowerBenchmarking2024}, but none of the surveyed methods reports it, so we do not compare it.

Models are also rarely evaluated under identical input and modality conditions: two systems may claim the same GFLOPs per video while processing different frame counts, resolutions and modalities. EgoSchema's intrinsic temporal length quantifies how much of a video must be processed to answer a question \cite{mangalamEgoSchemaDiagnosticBenchmark2023}, and frame-sampling and streaming methods report accuracy against frame or time budgets \cite{liuTrainingfreeLongVideo2025}, but these budgets are not standardized across papers.

We therefore use only values explicitly reported by each paper, record the corresponding model variant and input setting, and neither infer FLOPs or latency from architecture alone nor convert token-retention budgets into FLOPs; cross-paper comparisons serve only as indicative evidence. The quantitative comparisons emphasize 7B language backbones and approximately-8B configurations when the source reports a targeted efficiency mechanism. Parameter columns follow the source's accounting boundary, which can include the full model or only the language backbone. The taxonomy also includes transferable mechanisms evaluated on larger hosts and standalone encoder methods; these do not enter a shared-host comparison unless their evaluation setting matches it.

Following the pipeline perspective of prior efficiency surveys~\cite{zhang2026efficientinference,wu2026compressionlifecycle}, we classify each reduction by its position in the forward pass:
\begin{enumerate}
  \item \textbf{Input construction and selection}: selecting frames, patches, resolution or layouts before the encoder;
  \item \textbf{Encoder computation}: changing feature-extraction operators, reducing intermediate tokens, or reusing and substituting encoder computation;
  \item \textbf{Encoded representations and connector}: reducing encoder outputs or connector representations before the LLM;
  \item \textbf{LLM execution and state}: reducing token processing or attention within the LLM, constructing summary tokens with its layers, or managing its KV cache.
\end{enumerate}

We assign each mechanism to the stage whose computation it removes. Pooling applied after an encoder's final block therefore counts as stage 3 even when it is implemented in the encoder, and whole-frame selection can occur after encoding, as in Frame-Voyager~\cite{yuFrameVoyagerLearningQuery2025}. A method that reduces cost at several stages appears at each of them in Figure~\ref{fig:efficiency_taxonomy}, and its reported end-to-end gain is not attributed to a single stage.

\subsection{Input Construction and Selection}
\label{subsubsec:frame_sampling}

Frame sampling reduces the number of processed frames $T$ before the encoder runs, directly lowering encoder-side cost and the number of visual tokens $\widetilde N_v$ injected into the language model.
In encoder--connector--LLM pipelines, this upstream decision impacts both the cost of feature extraction and the LLM prefilling cost through the total context length $L=N_t+\widetilde N_v(+\widetilde N_a)$ (Section~\ref{subsec:bottlenecks}); frames discarded at this stage cannot be recovered downstream.
The two input families control temporal coverage and the spatial input budget, respectively. Temporal selectors may be query-free, using only the video signal, or query-aware, conditioning on the question or instruction. A proxy encoder or selector may process candidates that the target model never sees; its cost remains part of the selection pipeline.

\subsubsection{Temporal Sampling and Selection}

\paragraph{Fixed coverage sampling}
Uniform or strided sampling remains the simplest query-free baseline: it is deterministic, model-free, and often strong.
Recent controlled evaluation \cite{brkic2025framesamplingstrategiesmatter} confirms that frame-sampling choices alone can change video-QA results, and that uniform sampling can be the strongest strategy on Video-MME for some small VLMs \cite{fuVideoMMEFirstEverComprehensive2025}.
Temporal Segment Networks (TSN) \cite{wangTemporalSegmentNetworks2016} introduced a stronger fixed-budget pattern by splitting the video into $K$ segments and sampling one snippet per segment for constant-cost temporal coverage.
This ``coverage under fixed $K$'' idea remains a useful reference in later recognition and video-language pipelines \cite{wuAdaFrameAdaptiveFrame2018,leiLessMoreClipBERT2021}.

\paragraph{Content-based coverage}
For minute-to-hour videos, temporal redundancy makes fixed windows particularly inefficient.
Kernel Temporal Segmentation (KTS) \cite{potapovCategorySpecificVideoSummarization2014} partitions a sequence of frame descriptors into segments. Their KVS summarizer adds trained category-specific SVM scoring to select summary segments. Later work \cite{afhamRevisitingKernelTemporal2023} uses KTS to allocate samples before a downstream backbone for long-form classification and temporal localization.
MaxInfo \cite{liMaxInfoTrainingFreeKeyFrame2025} uses proxy frame embeddings to maximize the geometric volume spanned by the selected subset. MGSampler \cite{zhiMGSamplerExplainableSampling2021} uses motion saliency and motion-uniform temporal coverage without a learned sampling policy.

\paragraph{Learned query-free selection}
Learned query-free samplers use trained visual policies or scorers to adapt frame selection to each video without requiring a user query at inference time.
AdaFrame \cite{wuAdaFrameAdaptiveFrame2018} selects frames adaptively and performs early stopping using predicted future utilities.
PEEK \cite{steunou2026peek} distills caption-conditioned teacher rankings into a small visual temporal scorer, scoring frames from video embeddings alone.
Earlier adaptive methods similarly learned to concentrate computation on informative video regions \cite{yeungEndtoendLearningAction2017,fanWatchingSmallPortion2018,wuMultiAgentReinforcementLearning2019,wuLiteEvalCoarsetoFineFramework2019,mengARNetAdaptiveFrame2020}.

\paragraph{Query-conditioned relevance and diversity}
Query-aware methods condition selection on the question or instruction, usually by scoring frame--text alignment and then enforcing diversity or coverage.
Adaptive Keyframe Sampling (AKS) \cite{tangAdaptiveKeyframeSampling2025} combines prompt--frame relevance with temporal coverage under a fixed token budget.
Q-Frame \cite{zhangQFrameQueryawareFrame2025} uses a text-image matching model such as CLIP \cite{radfordLearningTransferableVisual2021} to score frames and also adapts per-frame resolution to process more frames within the same budget.
FOCUS \cite{zhuFOCUSEfficientKeyframe2025} formulates keyframe selection as pure exploration in a multi-armed bandit, identifying informative temporal regions while processing only a small fraction of candidate frames.
AdaRD-Key \cite{zhangAdaRDkeyAdaptiveRelevanceDiversity2025} maximizes a relevance--diversity objective and falls back to diversity-only selection when the query alignment is weak.
Several 2025--2026 methods extend this training-free line. BOLT~\cite{liu2025bolt} samples frames by inverse-transform sampling over CLIP frame--query similarity. F2C~\cite{sunFramesClipsTrainingfree2025} scores frame--query relevance to select anchor frames and extends them into coherent clips. T*~\cite{ye2025tstar} recasts temporal search as object-guided spatial search over frame mosaics, and reports that 8 selected frames outperform 32 uniform frames. QCA~\cite{peng2026qca} allocates the frame budget across segments by query relevance and content variation, EFS~\cite{chen2026efs} partitions the video into events and anchors selection on the most query-relevant frame per event, and GIFT~\cite{ma2026gift} scores each frame's global irreplaceability under the query, matching 64-frame uniform accuracy with 32 frames. KTV~\cite{song2026ktv} combines query-free keyframe clustering at this stage with post-encoder token pruning (Section~\ref{subsec:connector}).
LDDR \cite{chen2026lddr} relaxes the binary keep/drop decision itself: it linearizes determinantal-point-process selection (from quadratic to linear complexity in the frame count) while jointly allocating per-frame resolution under an explicit token budget, applicable even to closed-source hosts.
Related training-free methods explore scalable text-video similarity \cite{liangKeyVideoLLMLargescaleVideo2024}, sequential relevance-diversity allocation \cite{sunMDP3TrainingfreeApproach2025}, semantic query decomposition \cite{guoLogicinFramesDynamicKeyframe2025}, and lightweight moment retrieval for long-form VideoQA \cite{chasmaiMomentSamplingVideo2025}.

\paragraph{Learned and generative selectors}
Other query-aware methods train explicit selectors.
GenS \cite{yaoGenerativeFrameSampler2025} uses a separate VideoLLM to generate question-relevant frame selections for minute-to-hour videos, while HFS \cite{yangHFSHolisticQueryAware2025} optimizes a differentiable set-level objective combining relevance, coverage, and redundancy through Gumbel-Softmax \cite{janggumbelsoftmax2016} and student--teacher mutual learning.
Qin \textit{et al.} \cite{qin2026efficient} train a 0.4B plug-in selector with reinforcement learning that transfers across seven LLM hosts and selects 8 of 128 candidate frames at less than half the selection latency of AKS \cite{tangAdaptiveKeyframeSampling2025}.
Several recent selectors use reinforcement learning. TSPO~\cite{tang2025tspo} trains a temporal sampling policy with only 3.5M trainable parameters over frozen CLIP features using a GRPO-style (Group Relative Policy Optimization) objective and transfers it across hosts; ReFoCUS~\cite{lee2025refocus} optimizes a 1.3B policy with a logit-gap reward from the answering model, at a reported selection cost of 428~TFLOPs, 9~s and 5.3~GB over 512-frame inputs that its accuracy gains must amortize; and ViaRL~\cite{xu2025viarl} trains a 3B selector through iterated amplification to pick 8 of 128 candidate frames, though it reports no selector-overhead measurements. VideoITG~\cite{wang2025videoitg} trains an 8B instructed temporal-grounding selector by supervised fine-tuning on automatically annotated data; its 32 selected frames match 64-frame uniform sampling while scanning 512 candidates and adding only 0.61~s, but the selector's own size dominates any parameter-based efficiency accounting.
Related learned methods include M-LLM-based frame selection \cite{huMLLMBasedVideo2025}, which trains a lightweight selector from pseudo-labels; K-frames \cite{yaoKframesSceneDrivenAnyk2025}, which predicts query-relevant coherent clips under arbitrary frame budgets; and FrameOracle \cite{liFrameOracleLearningWhat2025}, which predicts both which frames to retain and how many are needed.

\subsubsection{Patch, Resolution, and Input-Layout Budgeting}
Temporal selection leaves another choice: how much spatial detail to encode in each retained frame. Q-Frame~\cite{zhangQFrameQueryawareFrame2025} and LDDR~\cite{chen2026lddr} allocate per-frame resolution by relevance, while F2C~\cite{sunFramesClipsTrainingfree2025} trades spatial resolution for longer clips under a fixed token budget. TS-LLaVA~\cite{qu2024tsllava} combines several downsampled frames into a thumbnail grid before encoding, then samples additional encoded tokens in stage~3 (Section~\ref{subsec:connector}).

AutoGaze~\cite{shi2026autogaze} selects multi-scale patches before the ViT using a 3M-parameter autoregressive selector. MeToM's residual-guided patch merging~\cite{wu2026metom} uses codec residual energy to identify connected low-information regions and average their patch embeddings before the heavy encoder blocks. Both reduce the input sequence those blocks process. MeToM additionally merges tokens after projection and inside the LLM; its reported end-to-end gain belongs to the combined pipeline.
The idea has been explored in VATT~\cite{akbariVATTTransformersMultimodal2021}, which randomly discards a fraction of the input patches and audio tokens before the transformer, and its ablation shows encoder GFLOPs falling with the drop rate at a growing accuracy cost on recognition benchmarks. It is a mechanism from before VideoLLMs, but it established that a video transformer tolerates a sparse input, which is the premise of the learned patch selection above.

\subsubsection{Discussion and Synthesis}
Frame sampling should be evaluated as a performance--budget trade-off, not as a single accuracy number. Clean comparisons fix the downstream model, frame budget, benchmark and split; otherwise the sampler, vision representation, connector and LLM capacity are confounded, an issue KFS-Bench~\cite{liKFSBenchComprehensiveEvaluation2025} makes explicit by scoring coverage of the disjoint evidence scenes required for long-video QA alongside answer accuracy. Table~\ref{tab:frame_sampling_isobackbone} therefore reports only methods sharing a LLaVA-Video-7B~\cite{zhangLLaVAVideoVideoInstruction2025}, 64-frame protocol; selectors evaluated under other protocols (such as PEEK's captioning setting~\cite{steunou2026peek}) are discussed in the text and excluded from the comparison. Query-free selections can be reused across questions. Query-aware methods gain up to 5 points on LongVideoBench~\cite{wuLongVideoBenchBenchmarkLongcontext2024} over uniform sampling when the question identifies sparse evidence, and TSPO~\cite{tang2025tspo} leads the table with only a 3.5M-parameter selector. Net savings still depend on whether avoided downstream work exceeds scorer cost, and the advantage largely disappears on Video-MME. Selectors evaluated at reduced budgets (VideoITG~\cite{wang2025videoitg}, GIFT~\cite{ma2026gift}) match the 64-frame uniform reference with 32 selected frames.

\begin{table*}[t]
\centering
\caption{\colorbox{color-frame}{Frame samplers} on \textbf{LLaVA-Video-7B} at a $\sim$64-frame budget. All rows share the same uniform baseline (LongVideoBench 58.9 / Video-MME 64.4) unless marked. $^\dagger$FOCUS reports a 32--64 frame budget, not a fixed 64. $^\ddagger$Own uniform-baseline reproduction differs from the shared one (EFS: 58.8/64.6).}
\label{tab:frame_sampling_isobackbone}
\scriptsize
\setlength{\tabcolsep}{4pt}
\begin{tabular}{lccccc}
\toprule
\textbf{Method} & \textbf{Query-aware} & \textbf{Train-free} & \textbf{Frames} & \textbf{LongVideoBench} & \textbf{V-MME} \\
\midrule
Uniform baseline & -- & -- & 64 & 58.9 & 64.4 \\
MaxInfo \cite{liMaxInfoTrainingFreeKeyFrame2025} & no & yes & 64 & 61.5 & 64.2 \\
AKS \cite{tangAdaptiveKeyframeSampling2025} & yes & yes & 64 & 62.7 & 65.3 \\
AdaRD-Key \cite{zhangAdaRDkeyAdaptiveRelevanceDiversity2025} & yes & yes & 64 & 62.9 & -- \\
FOCUS \cite{zhuFOCUSEfficientKeyframe2025} & yes & yes & 32--64$^\dagger$ & 63.5 & 65.4 \\
EFS \cite{chen2026efs} & yes & yes & 64 & 62.1$^\ddagger$ & 65.6$^\ddagger$ \\
QCA \cite{peng2026qca} & yes & yes & 64 & 62.9 & 66.1 \\
TSPO \cite{tang2025tspo} & yes & no & 64 & 63.9 & 65.5 \\
\bottomrule
\end{tabular}
\end{table*}

\subsection{Encoder Computation}
\label{subsec:encoder}
Encoder efficiency targets feature extraction before connector or LLM processing. We distinguish efficient architectures and operators, intermediate-token reduction, and computation reuse or substitution. Many architectural antecedents were evaluated on recognition or retrieval; their costs and accuracies must be kept separate from integrated VideoLLM results.

\subsubsection{Efficient Encoder Architectures and Operators}
\paragraph{Spatiotemporal backbones} Lightweight convolutional networks and hierarchical pooling-attention transformers reduce the cost of extracting video features. Among the convolutional backbones, TSM~\cite{linTSMTemporalShift2019} inserts a parameter- and FLOP-free channel shift into a 2D CNN to capture temporal structure at roughly the cost of a 2D network; X3D~\cite{feichtenhoferX3DExpandingArchitectures2020} progressively expands a small 2D image architecture along its temporal, spatial, channel-width, and depth dimensions, selecting efficient configurations under increasing compute budgets; and MoViNet~\cite{kondratyukMoViNetsMobileVideo2021} pairs neural architecture search (NAS) with a stream-buffer that decouples memory from clip length for constant-memory streaming inference. The transformer-based group reduces token resolution inside the encoder: MViT~\cite{fanMultiscaleVisionTransformers2021} and MViTv2~\cite{liMViTv2ImprovedMultiscale2022} use multi-head pooling attention to progressively pool spatiotemporal tokens while widening channels, MViTv2 adding decomposed relative position and residual pooling for 82.9 vs. 82.7 on Kinetics-400 at 51M vs. 88M parameters and a third of the inference compute of Video Swin (Table~\ref{tab:backbone_efficiency}); Hiera~\cite{ryaliHieraHierarchicalVision2023} strips MViTv2 of its specialized components and leans on strong masked auto-encoder pretraining, yielding a simpler backbone about $2\times$ faster on video (40.8 vs. 20.5 clips/s) at 5.0 points higher video accuracy than MViTv2-L; Video Swin~\cite{liuVideoSwinTransformer2021} restricts self-attention to shifted local 3D windows; and UniFormer~\cite{liUniFormerUnifiedTransformer2022} couples convolution-like local aggregation in shallow layers with global attention in deeper layers, with UniFormerV2~\cite{liUniFormerV2Spatiotemporal2023} equipping a frozen pretrained image ViT with lightweight video-specific UniBlocks.

\paragraph{State-space operators} State-space encoders replace quadratic self-attention with selective state-space models for linear-time video encoding. VideoMamba~\cite{liVideoMambaState2024} uses bidirectional Mamba blocks~\cite{guMambaLinearTimeSequence2023} to process spatiotemporal tokens, reporting $6\times$ higher throughput and $40\times$ lower GPU memory than TimeSformer-Ti at 64 frames (A100-80G, batch size 128). VideoMamba-ST~\cite{parkVideoMambaSpatio2024} adapts the scan order to video structure (we use the -ST suffix because Park et al.\ also name their model VideoMamba). VideoMambaPro~\cite{luSnakesLadders2025} addresses information leakage in the backward scan through masked backward computation and residual connections, improving Kinetics-400 accuracy by 1.6 points over VideoMamba-M at matched $32\times224^2$ input (84.0 vs. 82.4) with slightly fewer parameters and FLOPs.

\paragraph{Compact and sparse encoders} Compact encoders reduce the cost of feature extraction, either by distilling CLIP-style encoders \cite{radfordLearningTransferableVisual2021} or by designing the encoder to emit fewer tokens. On the distillation side, TinyCLIP~\cite{wuTinyCLIPCLIP2023} combines affinity-mimicking distillation with weight inheritance to shrink CLIP encoders, MobileCLIP~\cite{vasuMobileCLIPFast2023} uses multi-modal reinforced training to produce fast image--text encoders (MobileCLIP2~\cite{faghri2025mobileclip2} strengthens the teacher ensembles at 1.5--20~ms on-device encoder latencies), and MobileViCLIP~\cite{yangMobileViCLIPEfficientVideoText2025} carries this to video with a compact mobile video--text encoder: MobileViCLIP-Small runs $55.4\times$ faster than InternVideo2-L14 on mobile hardware at similar zero-shot retrieval performance. On the token-budget side, FastVLM~\cite{vasuFastVLMEfficient2024} introduces a hybrid convolution--transformer encoder that downsamples aggressively to emit far fewer high-resolution visual tokens, reporting $85\times$ faster time-to-first-token with a $3.4\times$ smaller vision encoder than LLaVA-OneVision-0.5B~\cite{li2024llavaonevision} at $1152^2$ input, while LiteFrame~\cite{kimLiteFrameEfficient2026} distills a compact VideoLLM vision encoder that emits compressed tokens and cuts end-to-end latency by $35\%$ relative to InternVL3-8B while processing $8\times$ more frames. MoE-ViE~\cite{zhang2026moevie} scales the encoder sparsely, activating 1.1B of 3.5B parameters per token through a fine-grained mixture of experts to match a dense encoder $1.7\times$ its size at roughly three quarters of its latency, with video-benchmark evidence on an 8B host.

Oryx~\cite{liu2024oryx} also changes the encoder: native-resolution processing avoids fixed-resolution tiling, followed by a dynamic compressor in stage~3 (Section~\ref{subsec:connector}). Finally MMV~\cite{alayracSelfSupervisedMultiModalVersatile2020} adopts TSM for inexpensive temporal modeling.

\subsubsection{Encoder-Internal Token Reduction}
ToMe~\cite{bolyaTokenMergingYour2023} merges similar tokens through bipartite matching between encoder blocks, raising ViT-L video throughput by $2.2\times$ for a $0.2$\%--$0.3$\% accuracy drop. EarlyTom~\cite{wang2026earlytom} merges frame features between encoder blocks, then selects spatial tokens after encoding. Its encoder reduction and post-encoder selection therefore occupy stages~2 and~3 (Section \ref{subsec:encoder} and~\ref{subsec:connector}). ResidualViT~\cite{soldanResidualViT2025} combines intermediate-token reduction with temporal reuse, discussed below. Because these methods act between blocks, the early blocks still process the full sequence; the patch selection of Section~\ref{subsubsec:frame_sampling} instead removes tokens before the first block, so the encoder never sees them.

\subsubsection{Encoder Computation Reuse and Substitution}
ResidualViT~\cite{soldanResidualViT2025} propagates a residual subset of tokens across frames for temporally dense encoding, reducing per-frame encoding cost by $53$\%--$56$\% within 1.7 points of CLIP R@1 on Charades-STA. STC~\cite{wang2025stc} combines STC-Cacher, which reuses cached ViT features for temporally similar content, with STC-Pruner, which compresses the encoded sequence before LLM input. With cached features reused for $75\%$ of tokens, the combined method reduces ViT-encoding latency by $24.5\%$ and LLM prefilling latency by $45.3\%$. MoViNet's stream buffer~\cite{kondratyukMoViNetsMobileVideo2021} and the online variant of TSM~\cite{linTSMTemporalShift2019} retain temporal features between successive inputs.

CoPE-VideoLM~\cite{sarkar2026cope} runs the image encoder only on the few frames a video codec stores in full, and encodes the remaining frames from the motion and residual information the codec already provides, using a delta-encoder of under 15M parameters in place of dense RGB encoding.

\begin{table*}[t]
\centering
\caption{\colorbox{color-backbone}{Vision Encoder} efficiency methods. GFLOPs$\times$v gives inference GFLOPs per view times the number of temporal$\times$spatial views used for the reported accuracy, as stated by each paper; $^t$ marks papers reporting only the total across views (per-view cost not separately stated). Abbreviations: K-400/600 = Kinetics-400/600 \cite{kayKineticsHumanAction2017}, MiT = Moments in Time \cite{monfortMomentsTimeDataset2019}, AS = AudioSet \cite{gemmeke2017audioset}, UCF = UCF101 \cite{soomroUCF101Dataset101}, HMDB = HMDB51 \cite{kuehne2011hmdb}.}
\label{tab:backbone_efficiency}
\scriptsize
\setlength{\tabcolsep}{3pt}
\begin{tabular}{lccccccccc}
\toprule
\textbf{Method} & \textbf{Year} & \textbf{Params (B)} & \textbf{GFLOPs$\times$v} & \textbf{K-400} & \textbf{K-600} & \textbf{MiT} & \textbf{AS} & \textbf{UCF} & \textbf{HMDB} \\
\midrule
TSM~\cite{linTSMTemporalShift2019} & 2019 & 0.024 & 65$\times$1 & 74.7 & —    & —    & —    & 95.9 & 73.5 \\
MMV~\cite{alayracSelfSupervisedMultiModalVersatile2020} & 2020 & 0.094 & — & — & 70.5 & — & 30.9 & 95.2 & 75.0 \\
X3D~\cite{feichtenhoferX3DExpandingArchitectures2020} & 2020 & 0.011 & 35.84$\times$10 & 78.4 & 81.9 & —    & —    & —    & —    \\
MoViNet~\cite{kondratyukMoViNetsMobileVideo2021} & 2021 & 0.031 & 386$\times$1 & —    & 84.8 & 39.9 & —    & —    & —    \\
MViT~\cite{fanMultiscaleVisionTransformers2021} & 2021 & 0.037 & 170$\times$5 & 80.2 & 83.4 & —    & —    & —    & —    \\
VATT~\cite{akbariVATTTransformersMultimodal2021} & 2021 & 0.155 & 15\,020$^t$ & 79.9 & 80.8 & 37.8 & 39.3 & —    & —    \\
MViTv2~\cite{liMViTv2ImprovedMultiscale2022} & 2022 & 0.051 & 225$\times$5 & 82.9 & 85.5 & —    & —    & —    & —    \\
Video Swin\cite{liuVideoSwinTransformer2021} & 2022 & 0.088 & 282$\times$12 & 82.7 & —    & —    & —    & —    & —    \\
UniFormer~\cite{liUniFormerUnifiedTransformer2022} & 2022 & 0.050 & 3108$^t$ & 83.0 & 84.9 & —    & —    & —    & —    \\
Hiera~\cite{ryaliHieraHierarchicalVision2023} & 2023 & 0.213 & 413$\times$15 & 87.3 & —    & —    & —    & —    & —    \\
UniFormerV2~\cite{liUniFormerV2Spatiotemporal2023} & 2023 & 0.354 & 75\,300$^t$ & 90.0 & 90.1 & 47.8 & —    & —    & —    \\
VideoMamba~\cite{liVideoMambaState2024} & 2024 & 0.074 & 403$\times$12 & 82.4 & —    & —    & —    & 88.2 & 60.8 \\
VideoMamba-ST~\cite{parkVideoMambaSpatio2024} & 2024 & 0.027 & 68$\times$15 & 77.7 & —    & —    & —    & —    & 75.7 \\
VideoMambaPro~\cite{luSnakesLadders2025} & 2025 & 0.072 & 4700$^t$ & 84.0 & —    & —    & —    & 91.6 & 63.2 \\
\bottomrule
\end{tabular}
\end{table*}

% Accuracy--compute trade-off for the vision-encoder backbones of
% Table~\ref{tab:backbone_efficiency}. Total inference compute = per-view
% GFLOPs x number of views (Table's GFLOPs x v column, product taken);
% $^t$ entries enter as the paper-stated totals. Values card-verified
% (col-verify 2026-07-20); MMV and MoViNet omitted (no K-400 result in
% our table). VideoMambaPro's R9 per-clip-vs-aggregate ambiguity was
% resolved 2026-07-29: its 4.7T is already the 12-view total, so it
% enters at 4700, not 4700x12 = 56400. See tables/comparison_backbone.tex.
\definecolor{bbCnn}{HTML}{0072B2}
\definecolor{bbPool}{HTML}{E69F00}
\definecolor{bbTfm}{HTML}{009E73}
\definecolor{bbSsm}{HTML}{CC79A7}
\begin{figure}[t]
\centering
\begin{tikzpicture}
\begin{semilogxaxis}[
  width=\columnwidth, height=0.82\columnwidth,
  xlabel={Total inference compute (GFLOPs, log scale)},
  ylabel={Kinetics-400 top-1 (\%)},
  xmin=40, xmax=300000, ymin=73.5, ymax=91.5,
  grid=major, grid style={gray!20},
  axis line style={gray!70},
  tick label style={font=\scriptsize},
  label style={font=\footnotesize},
  legend style={font=\scriptsize, at={(0.03,0.97)}, anchor=north west,
    draw=gray!40, fill=white, fill opacity=0.85, text opacity=1},
  legend cell align=left,
]
\addplot[only marks, mark=*, mark size=2.4pt, color=bbCnn] coordinates {(65,74.7) (358.4,78.4)};
\addlegendentry{Convolutional}
\addplot[only marks, mark=square*, mark size=2.2pt, color=bbPool] coordinates {(850,80.2) (1125,82.9) (6195,87.3) (75300,90.0)};
\addlegendentry{Pooling attention}
\addplot[only marks, mark=triangle*, mark size=2.9pt, color=bbTfm] coordinates {(15020,79.9) (3384,82.7) (3108,83.0)};
\addlegendentry{Global attention}
\addplot[only marks, mark=diamond*, mark size=2.9pt, color=bbSsm] coordinates {(1020,77.7) (4836,82.4) (4700,84.0)};
\addlegendentry{State-space}
\begin{scope}[every node/.style={font=\tiny, text=black!75, inner sep=1.5pt}]
\node[anchor=west]  at (axis cs:65,74.7)    {TSM};
\node[anchor=west]  at (axis cs:358.4,78.4) {X3D};
\node[anchor=west]  at (axis cs:850,80.2)   {MViT};
\node[anchor=east]  at (axis cs:1125,82.9)  {MViTv2};
\node[anchor=south] at (axis cs:6195,87.3)  {Hiera};
\node[anchor=east]  at (axis cs:75300,90.0) {UniFormerV2};
\node[anchor=north] at (axis cs:15020,79.9) {VATT};
\node[anchor=north] at (axis cs:3384,82.7)  {Video Swin};
\node[anchor=south] at (axis cs:3108,83.0)  {UniFormer};
\node[anchor=west]  at (axis cs:1020,77.7)  {VideoMamba-ST};
\node[anchor=north] at (axis cs:4836,82.4) {VideoMamba};
\node[anchor=south] at (axis cs:4700,84.0) {VideoMambaPro};
\end{scope}
\end{semilogxaxis}
\end{tikzpicture}
\caption{Reported Kinetics-400 accuracy versus inference compute for the backbones in Table~\ref{tab:backbone_efficiency}. Compute is per-view GFLOPs times evaluation views; the plot is illustrative because training data and test protocols differ.}
\label{fig:backbone_tradeoff}
\end{figure}
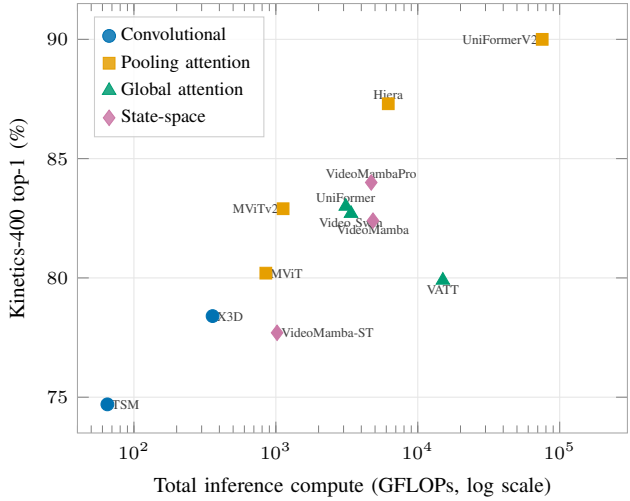

\subsubsection{Discussion and Synthesis}
Convolutional backbones occupy the low-compute regime, while pooling-attention transformers span the widest accuracy range and reach the highest absolute accuracies. State-space backbones are competitive (VideoMambaPro~\cite{luSnakesLadders2025} reaches 84.0 top-1 at 4.7~TFLOPs, above the global-attention baselines at comparable cost) but none matches Hiera~\cite{ryaliHieraHierarchicalVision2023} or UniFormerV2~\cite{liUniFormerV2Spatiotemporal2023} at any compute, as shown in Figure~\ref{fig:backbone_tradeoff}. The evaluation protocol also reshapes the apparent trade-off: MViTv2~\cite{liMViTv2ImprovedMultiscale2022} and Video Swin~\cite{liuVideoSwinTransformer2021} report nearly identical Kinetics-400 accuracy, yet 1.13 versus 3.38~TFLOPs because they evaluate with five versus twelve views. Table~\ref{tab:backbone_efficiency} therefore supports comparisons between reported operating points, not attribution of gaps to architecture alone; and encoder-only FLOPs do not establish end-to-end VideoLLM efficiency, which also depends on the output token count and the downstream connector and LLM.

Several recent encoder mechanisms, including EarlyTom, STC and CoPE, are evaluated in VideoLLMs and report time-to-first-token or end-to-end latency. Compact image encoders and standalone retrieval methods were evaluated on different tasks, so their gains cannot be transferred numerically to VideoLLM QA.

\subsection{Encoded Representations and Connector}
\label{subsec:connector}
This stage reduces encoded representations before the answering LLM consumes them. The reduction may act on whole frames, individual tokens, pooled grids, latent representations or a maintained memory bank. Its immediate benefit is a smaller LLM input; encoding costs have already been paid unless a separate upstream mechanism also reduces them.

\subsubsection{Selection and Merging of Encoded Representations}\label{subsec:encoded_selection} This family removes or fuses encoded tokens. Many methods operate on a frozen host, while others train the system around the reduction. VisionZip~\cite{yang2024visionzip} keeps only the most informative tokens (retaining $6.6\%$ of them at a $7.8\times$ prefilling speed-up), LLaVA-PruMerge~\cite{shang2024prumerge} adaptively prunes and merges for $14\times$ average visual-token compression, and Chat-UniVi~\cite{jinChatUniViUnifiedVisual2024} uses parameter-free clustering to merge tokens. PruneVid~\cite{huang-etal-2025-prunevid} and HoliTom~\cite{shao2025holitom} exploit spatiotemporal redundancy (HoliTom runs at roughly $10\%$ of the baseline FLOPs), and LongVU~\cite{shenLongVUSpatiotemporalAdaptive2024} combines frame selection before SigLIP~\cite{zhaiSigmoidLossLanguage2023} with query-conditioned pooling and token pruning after encoding. FlashVID~\cite{fan2026flashvid} combines attention- and diversity-based token selection with tree-based spatiotemporal merging, holding $99.1\%$ relative accuracy at $10\%$ retention with a $6.3\times$ prefilling speed-up, and EchoPrune~\cite{li2026echoprune} drops tokens that are reconstructible from previous frames, interpreting them as temporal echoes, which allows using up to $20\times$ more frames under a fixed token budget. DyToK~\cite{li2025dytok} introduces a budget-allocation policy: an assistant model supplies a query-conditioned per-frame prior, which is converted into per-frame retention ratios. Where the reduction happens depends on the compressor it drives: before the LLM with VisionZip~\cite{yang2024visionzip}, inside it with FastV~\cite{chen2024fastv}, and at both with DyCoke~\cite{tao2024dycoke}. The assistant model's forward pass adds cost that a complete comparison must count.
Recent papers differ mainly in how the token budget is allocated across time. FastVID~\cite{shen2025fastvid} partitions the video into temporally ordered segments and prunes by density within each, reducing FLOPs to $8.3\%$ for a $7.1\times$ prefilling speed-up at $98\%$ retained accuracy; LLaVA-Scissor~\cite{sun2025llavascissor} compresses through semantic connected components; and VidCom2~\cite{liu2025vidcom2} adapts per-frame compression intensity to frame uniqueness, reducing LLM-generation latency by $70.8\%$ at a quarter of the tokens. Segment-level budget allocation recurs in MMG-Vid~\cite{ma2025mmgvid} (marginal-gain maximization, $3.9\times$ prefilling speed-up at $25\%$ retention), OTT-Vid~\cite{kang2026ottvid} (optimal-transport cost between neighboring frames), InfoMerge~\cite{liu2026infomerge} (second-order temporal fingerprints with spectral-entropy budgets, $4.2\times$ prefilling speed-up at $15\%$ tokens), DynaTok~\cite{park2026dynatok} (an EMA novelty memory with positional-bias-aware spatial selection), and ForestPrune~\cite{ju2026forestprune} (globally optimized pruning over spatio-temporal token forests). MeToM~\cite{wu2026metom} allocates post-projector token budgets from groups of pictures packet sizes, then merges redundant tokens across time and within frames. This stage complements its input-patch merging and LLM-layer merging; the reported $2.65\times$ time-to-first-token speed-up measures their combined effect. Related methods condition on the query or train the reduction: KTV~\cite{song2026ktv} runs video through an image-only VLM without training by clustering frames into keyframes and then pruning each keyframe's tokens by importance and redundancy, LGTTP~\cite{kumar2025lgttp} prunes tokens outside a query-predicted temporal window through a trained auxiliary classifier, and DynTok~\cite{zhang2025dyntok} trains the grouping-and-merging step into the model itself to avoid a training--inference mismatch.

Frame-Voyager~\cite{yuFrameVoyagerLearningQuery2025} selects whole frames at this stage. All candidate frames first pass through the host visual encoder and projector; pooled features and the query then enter a scorer built from frozen bottom LLM layers and trained reward heads. The selected frames supply the answering context. This reduces the context relative to answering over all candidates, but does not spare their initial encoding. Its Appendix~C reports $27.6\%$ higher latency than uniform sampling in the tested setting, illustrating why selection quality and net speedup require separate comparisons.

FlexSelect~\cite{zhang2025flexselect} also selects tokens before a final answering pass. Its base variant scores encoded frame sets through partial host-model forwards, then aggregates the selected tokens; FlexSelect-Lite replaces that scorer with a trained lightweight selector.

VideoChat-Flash~\cite{li2025videochatflash} merges similar clip tokens before the LLM and progressively drops tokens inside its layers. TimeChat-Online~\cite{yao2025timechatonline} drops encoded tokens whose content is unchanged between successive frames ($82.8\%$ reduction at approximately $98\%$ retained streaming accuracy and $1.76\times$ faster responses); the dropping rule also transfers without training to Qwen2.5-VL. StreamingTOM~\cite{streamingTOM2025} combines causal temporal selection and merging here with quantized memory and retrieval in stage~4 (Section~\ref{subsubsec:llm_side}).

Audio can supply the selection signal. OmniZip~\cite{taoOmniZipAudioGuidedDynamic2025} and DASH~\cite{li2026dash} use audio to guide visual-token reduction; OmniZip also compresses the audio stream. Its Qwen2.5-Omni-7B configuration reports a $3.42\times$ speed-up and $1.4\times$ memory reduction at $35\%$ token retention (Table~\ref{tab:context_reduction}). The audio encoder that supplies the guidance is part of the cost and should appear in the reported savings.

\subsubsection{Grid Pooling and Downsampling} Spatial or temporal downsampling reduces the encoded sequence through a prescribed grid structure: pixel-shuffle in InternVL2.5~\cite{chenExpandingPerformanceBoundaries2025}, adaptive pooling in PLLaVA~\cite{xuPLLaVAParameterfreeLLaVA2024}, a SlowFast two-stream projector in SF-LLaVA~\cite{xu2024slowfastllava}, the spatial-temporal convolution connector of VideoLLaMA~2~\cite{chengVideoLLaMA2Advancing2024}, and a learned Mamba temporal connector in STORM~\cite{jiangSTORMTokenEfficientLong2025} that cuts computation by up to $8\times$ and decoding latency by $2.4$--$2.9\times$ at a fixed frame count.
Several efficiency-first VideoLLM architectures make this stage their central design. NVILA~\cite{liu2024nvila} scales spatial and temporal resolution first, then compresses tokens, for $1.6$--$2.2\times$ lower prefilling and $1.2$--$2.8\times$ lower decoding latency than comparable open VLMs; PVC~\cite{yang2024pvc} uses temporal attention to enrich frame features before compressing each frame to 64 tokens; TS-LLaVA~\cite{qu2024tsllava} builds a fixed 3{,}456-token budget from a detail thumbnail plus tokens sampled across 50 frames.

VideoScan~\cite{li2025videoscan} pools each frame into a semantic-carrier token before the LLM and learns a KV propagation policy inside it. Qwen2-Audio~\cite{chuQwen2AudioTechnicalReport2024} is an audio-only precedent: stride-2 pooling follows the audio encoder's transformer blocks. Baichuan-Omni~\cite{liBaichuanOmniTechnicalReport2024} uses convolutional downsampling, while HyperCLOVA~X~8B~\cite{hyperclovaxteamHyperCLOVA8BOmni2026} adopts MambaMia to reduce the audio rate from 25~Hz to 1~Hz after its adapter. The latter report does not describe how its visual tokens are reduced, so we classify only the audio downsampling; its reported visual-token budget and training-cost savings are not attributable to a described mechanism.

\subsubsection{Latent Resampling and Compact Representation Construction} Learned resamplers construct a compact representation from the encoder outputs, often through cross-attention with a small set of latent queries. The Perceiver Resampler of Flamingo~\cite{alayracFlamingoVisualLanguage2022} and the Q-Former of Video-LLaMA~\cite{zhangVideoLLaMAInstructiontunedAudioVisual2023} are early influential examples of learned resampling; LLaMA-VID~\cite{liLLaMAVIDImageWorth2023} combines a query-conditioned context token with pooled content, reaching two tokens per frame in its compressed setting, and LLaVA-Mini~\cite{zhangLLaVAMiniEfficientImage2025} reaches a single vision token via modality pre-fusion ($-77\%$ FLOPs). BLIP-3-Video~\cite{ryoo2024blip3video} abstracts an entire video into 16--32 learned tokens; Quicksviewer~\cite{qi2025quicksviewer} learns nonuniform temporal ``cubes'' through Gumbel-Softmax and resamples 64 tokens per cube for a $45\times$ overall compression; and VidCompress~\cite{lan2024vidcompress} pairs a memory-enhanced compressor emitting one token per frame with a text-perceived Q-Former branch. VQToken~\cite{zhang2025vqtoken} replaces the continuous bottleneck with a discrete one: adaptive vector quantization maps ViT embeddings onto a learned codebook, with a token hash preserving spatiotemporal position, shrinking the video stream to $0.07\%$ of its tokens at a $0.66$-point drop on NExT-QA.

Oryx~\cite{liu2024oryx} combines native-resolution encoding with an on-demand cross-attention compressor at $1\times$--$16\times$ reduction. Audiovisual resamplers follow the same principle: FAVOR~\cite{sunFinegrainedAudioVisualJoint2023} uses a windowed causal Q-Former to enforce a joint budget, and video-SALMONN~\cite{sunVideoSALMONNSpeechEnhancedAudioVisual2024} queries features at fine (approximately 0.5~s) and coarse (approximately 5~s) temporal resolutions. Table~\ref{tab:context_reduction} reports their 7B configurations.

\subsubsection{Representation-Memory Compression and Retrieval}
This family compresses a maintained history of encoded representations outside the answering LLM. MovieChat~\cite{songMovieChatDenseToken2024} merges features as its short-term buffer fills, MA-LMM~\cite{heMALMMMemoryAugmented2024} maintains compressed visual and query memory banks around the Q-Former, and $\infty$-Video~\cite{santos$infty$VideoTrainingFreeApproach2025} uses a continuous-time long-term representation with resampling. VidCompress~\cite{lan2024vidcompress} and the Token Turing Machine variant of BLIP-3-Video~\cite{ryoo2024blip3video} also maintain state while constructing compressed outputs.

Flash-VStream~\cite{zhang2025flashvstream} maintains a two-part memory combining compact context with selected high-resolution details, while VideoLLaMB~\cite{wang2025videollamb} propagates recurrent memory bridges across semantic segments. AdaCM$^2$~\cite{man2025adacm2} prunes the Q-Former video cache using cross-modal attention, processing videos beyond two hours with a reported $65\%$ reduction in GPU memory. This cache belongs to the connector; it is distinct from the answering LLM's KV cache.

\subsubsection{Discussion and Synthesis}
Table~\ref{tab:context_reduction} summarizes these methods. Because the quoted accuracies come from different language backbones and host VLMs, we use its rows only as indicative evidence. Where the literature provides a controlled comparison, we report it separately: Table~\ref{tab:context_holitom} compares the training-free methods that HoliTom~\cite{shao2025holitom} re-ran on a single frozen LLaVA-OneVision-7B host~\cite{li2024llavaonevision} at matched token budgets, a comparison of post-encoder and joint-stage methods alongside Tables~\ref{tab:frame_sampling_isobackbone} and~\ref{tab:llmside_hieravid}. At a $25\%$ budget the compared methods with post-encoder reduction stay within $1.5\%$ of the uncompressed baseline average while DyCoke~\cite{tao2024dycoke}, which combines pre-LLM temporal merging and decoder KV management (Section~\ref{subsubsec:llm_side}), loses over $7\%$. At $10\%$ the near-tie breaks down, and methods that model temporal redundancy explicitly (PruneVid~\cite{huang-etal-2025-prunevid}, HoliTom~\cite{shao2025holitom}) degrade far more gracefully than spatial-only selection (VisionZip~\cite{yang2024visionzip}). The lower block adds two methods whose own runs reproduce the same host and harness: FlashVID~\cite{fan2026flashvid} matches HoliTom's near-lossless behavior at both budgets, while EarlyTom~\cite{wang2026earlytom}, which prunes inside the vision encoder and therefore also cuts encoding compute, stays competitive at $25\%$ but sits between spatial-only and temporal-aware methods at $10\%$. Connector-stage reduction lowers LLM prefilling cost without reducing the cost of encoding the retained frames; the saved budget can instead widen temporal coverage, as in EchoPrune~\cite{li2026echoprune}, so fixed-input and fixed-downstream-budget evaluations measure different benefits.

\begin{table*}[!tp]
\centering
\caption{Reported performance of token reduction methods before and within the LLM. Groups follow the families in Figure~\ref{fig:efficiency_taxonomy}; joint methods may also act at other stages. $^h$ marks a plug-in host size. Retained budgets follow the source and may describe tokens, memory or audio rate. Hosts, inputs, protocols and baselines differ, so these rows are indicative and do not rank methods. NX-QA = NExT-QA, MSR = MSR-VTT-QA, MSVD = MSVD-QA, MVB = MVBench, VME = Video-MME w/o, ES = EgoSchema, ActNet = ActivityNet-QA. HyperCLOVA\textquotesingle s budget describes its documented audio compression; its QA score is a system-level result.}
\label{tab:context_reduction}
\scriptsize
\setlength{\tabcolsep}{2pt}
\begin{tabular}{lccccccccccc}
\toprule
\textbf{Method} & \textbf{Year} & \textbf{Params (B)} & \textbf{LLM/host} & \textbf{Retained budget} & \textbf{NX-QA} & \textbf{MSR} & \textbf{MSVD} & \textbf{MVB} & \textbf{VME} & \textbf{ES} & \textbf{ActNet} \\
\midrule
\multicolumn{12}{l}{\colorbox{color-context}{\textcolor{white}{\textit{3a. Selection and merging of encoded representations}}}} \\
VisionZip~\cite{yang2024visionzip} & 2024 & 7$^h$  & Video-LLaVA & 6.6\% & —    & 52.1 & 63.5 & —    & —    & —    & 43.0 \\
LLaVA-PruMerge~\cite{shang2024prumerge} & 2024 & 7$^h$  & Video-LLaVA & 256/img & —    & 59.3 & 71.1 & —    & —    & —    & 47.7 \\
Chat-UniVi~\cite{jinChatUniViUnifiedVisual2024} & 2024 & 7      & Vicuna-1.5 & 44\% & —    & 55.0 & 69.3 & —    & —    & —    & 46.1 \\
LongVU~\cite{shenLongVUSpatiotemporalAdaptive2024} & 2024 & 7      & Qwen2 & 45\% & —    & —    & —    & 66.9 & 60.6 & 67.6 & —    \\
PruneVid~\cite{huang-etal-2025-prunevid} & 2025 & 7$^h$  & LLaVA-OV & 15--17\% & —    & —    & —    & 57.5 & 58.6 & 59.5 & —    \\
HoliTom~\cite{shao2025holitom} & 2025 & 7      & LLaVA-OV & 10\% & —    & —    & —    & 57.3 & 56.8 & 61.2 & —    \\
FlashVID~\cite{fan2026flashvid} & 2026 & 7$^h$  & LLaVA-OV & 10\% & —    & —    & —    & 57.4 & 57.8 & 60.0 & —    \\
EchoPrune~\cite{li2026echoprune} & 2026 & 7$^h$  & LLaVA-OV & 10\%/320f & — & —  & —    & —    & 61.8 & 60.4 & —    \\
FastVID~\cite{shen2025fastvid} & 2025 & 7$^h$  & LLaVA-OV & 25\% & —    & —    & —    & 56.3 & 58.0 & —    & —    \\
LLaVA-Scissor~\cite{sun2025llavascissor} & 2025 & 7$^h$  & LLaVA-OV & 10\% & 80.0 & — & —  & 57.9 & 55.2 & 57.5 & 47.8 \\
VidCom2~\cite{liu2025vidcom2} & 2025 & 7$^h$  & LLaVA-OV & 25\% & —    & —    & —    & 57.2 & 58.6 & 59.7 & —    \\
MMG-Vid~\cite{ma2025mmgvid} & 2025 & 7$^h$  & LLaVA-OV & 25\% & —    & —    & —    & 56.7 & 58.6 & —    & —    \\ 
TS-LLaVA~\cite{qu2024tsllava} & 2024 & 7      & Vicuna-1.5 & 3456/50f & 66.5 & 65.1 & 79.0 & 45.5 & —    & 50.2 & 56.7 \\ 
VideoChat-Flash~\cite{li2025videochatflash} & 2025 & 7      & Qwen2 & 16/frame & —    & —    & —    & 74.0 & 65.3 & —    & —    \\
StreamingTOM~\cite{streamingTOM2025} & 2025 & 7      & LLaVA-OV & 25.5\% & —    & —    & —    & —    & 59.9 & 63.7 & —    \\ 
TimeChat-Online~\cite{yao2025timechatonline} & 2025 & 7      & Qwen2.5-VL & $\sim$17\% & —    & —    & —    & —    & 62.5 & —    & —    \\ 
OmniZip~\cite{taoOmniZipAudioGuidedDynamic2025} & 2025 & 7$^h$  & Qwen2.5-Omni & 35\% & —    & —    & —    & —    & 66.1 & —    & —    \\ 
DASH~\cite{li2026dash} & 2026 & 7      & Qwen2.5-Omni & 25\% & —    & —    & —    & —    & 66.0 & —    & —    \\ 
\midrule
\multicolumn{12}{l}{\colorbox{color-context}{\textcolor{white}{\textit{3b. Grid pooling and downsampling}}}} \\
VideoLLaMA 2~\cite{chengVideoLLaMA2Advancing2024} & 2024 & 7$^h$  & Mistral & 50\% & —    & —    & 70.9 & 54.6 & 47.9 & 51.7 & 50.2 \\ 
InternVL2.5~\cite{chenExpandingPerformanceBoundaries2025} & 2024 & 8.1    & InternLM2.5 & 25\% & —    & —    & —    & 72.0 & 64.2 & —    & —    \\ 
PLLaVA~\cite{xuPLLaVAParameterfreeLLaVA2024} & 2024 & 7      & LLaVA-NeXT & 25\% & —    & 62.0 & 76.6 & —    & —    & —    & 56.3 \\
SF-LLaVA~\cite{xu2024slowfastllava} & 2024 & 7      & LLaVA-NeXT & 3680 total & 64.2 & 65.8 & 79.1 & —    & —    & 47.2 & 55.5 \\
STORM~\cite{jiangSTORMTokenEfficientLong2025} & 2025 & 7      & Qwen2 & 25\% & —    & —    & —    & 71.3 & 63.4 & —    & —    \\ 
NVILA~\cite{liu2024nvila} & 2024 & 8      & Qwen2 & 1/8 & 82.2 & —    & —    & 68.1 & 64.2 & —    & 60.9 \\ 
PVC~\cite{yang2024pvc} & 2024 & 8      & InternLM2.5 & 64/frame & 82.0 & —    & —    & 73.8 & 64.1 & 59.6 & 57.1 \\ 
VideoScan~\cite{li2025videoscan} & 2025 & 7      & LLaVA-Video & 1/frame & —    & —    & —    & 48.9 & 53.7 & —    & —    \\ 
Baichuan-Omni~\cite{liBaichuanOmniTechnicalReport2024} & 2024 & 7      & own & 182--546/video & — & — & 72.2 & 60.9 & 58.2 & 58.8 & 58.6 \\ 
HyperCLOVA X 8B~\cite{hyperclovaxteamHyperCLOVA8BOmni2026} & 2026 & 8      & own & 1/s audio & — & — & — & — & 58.2 & — & — \\
\midrule
\multicolumn{12}{l}{\colorbox{color-context}{\textcolor{white}{\textit{3c. Latent resampling and compact representation construction}}}} \\
LLaMA-VID~\cite{liLLaMAVIDImageWorth2023} & 2023 & 7      & Vicuna & 2/frame & —    & 57.7 & 69.7 & —    & —    & —    & 47.4 \\ 
LLaVA-Mini~\cite{zhangLLaVAMiniEfficientImage2025} & 2025 & 7      & Vicuna-1.5 & 1/frame & —    & 59.5 & 70.9 & 44.5 & —    & 51.2 & 53.5 \\
BLIP-3-Video~\cite{ryoo2024blip3video} & 2024 & 4      & Phi-3-Mini & 32/video & 76.4 & 60.0 & 77.7 & 54.9 & —    & —    & 55.7 \\
Quicksviewer~\cite{qi2025quicksviewer} & 2025 & 8      & Qwen2.5-7B & 64/cube & 77.5 & —    & —    & 55.6 & 56.9 & —    & 47.6 \\ 
VidCompress~\cite{lan2024vidcompress} & 2024 & 7      & Vicuna & 1/frame+QF & —    & 57.7 & 68.9 & 46.9 & 43.0 & —    & 48.3 \\
Oryx~\cite{liu2024oryx} & 2024 & 7      & Qwen2 & 1/4--1/16 & 81.9 & —    & —    & 63.9 & 58.3 & —    & —    \\ 
FAVOR~\cite{sunFinegrainedAudioVisualJoint2023} & 2023 & 7$^h$ & Vicuna & 160/25~s & 42.5 & —    & —    & —    & —    & —    & —    \\
video-SALMONN~\cite{sunVideoSALMONNSpeechEnhancedAudioVisual2024} & 2024 & 7$^h$  & Vicuna-1.5 & 160/25~s & 42.5 & —    & —    & —    & —    & —    & —    \\
\midrule
\multicolumn{12}{l}{\colorbox{color-context}{\textcolor{white}{\textit{3d. Representation-memory compression and retrieval}}}} \\
MovieChat~\cite{songMovieChatDenseToken2024} & 2024 & 7$^h$  & Vicuna & 576 mem & —    & 52.7 & 75.2 & —    & —    & —    & 45.7 \\
MA-LMM~\cite{heMALMMMemoryAugmented2024} & 2024 & 7      & Vicuna & 32 mem & —    & 48.5 & 60.6 & —    & —    & —    & 49.8 \\ 
$\infty$-Video~\cite{santos$infty$VideoTrainingFreeApproach2025} & 2025 & 7      & V-LLaMA/VC2 & — & 41.1 & —    & —    & —    & 42.4 & 46.8 & —    \\ 
Flash-VStream~\cite{zhang2025flashvstream} & 2025 & 7      & Qwen2 & 11.5K/stream & —    & —    & —    & 65.4 & 61.2 & 68.2 & —    \\ 
VideoLLaMB~\cite{wang2025videollamb} & 2025 & 7      & Vicuna-1.5 & 32 mem/seg & 71.1 & —    & —    & 52.5 & 41.4 & 53.8 & —    \\ 
\midrule
\multicolumn{12}{l}{\colorbox{color-llm}{\textcolor{black}{\textit{4a. Decoder token pruning and merging}}}} \\
STTM~\cite{hyun2025sttm} & 2025 & 7$^h$  & LLaVA-OV & 50\% & 80.4 & —    & —    & —    & 60.7 & 61.7 & —    \\ 
\midrule
\multicolumn{12}{l}{\colorbox{color-llm}{\textcolor{black}{\textit{4c. LLM-computed summary tokens}}}} \\
VoCo-LLaMA~\cite{yeVoCoLLaMAVisionCompression2024} & 2024 & 7      & Vicuna & 2/frame & —    & 61.1 & 72.3 & —    & —    & —    & 47.9 \\ 
Video-XL~\cite{shu2024videoxl} & 2024 & 7      & Qwen2 & KV 1/16 & —    & —    & —    & 55.3 & 55.5 & —    & —    \\ 
\bottomrule
\end{tabular}
\end{table*}

\begin{table*}[!tp]
\centering
\caption{Training-free reduction on a shared \textbf{LLaVA-OneVision-7B} host (32 frames, LMMs-Eval). HoliTom re-runs the upper blocks under one harness \cite{shao2025holitom}; the lower block collects own-paper runs on the same host and frame count. FLOPs are relative LLM-prefill costs, except $^e$, which also includes vision encoding. Avg.\ is relative to the 58.4 baseline mean; $^p$ marks a paper-reported relative average against that paper's own baseline. $^m$ marks methods whose own MVBench baseline reproduction differs from the shared 58.3 (VidCom2 and FastVID report 56.9, MMG-Vid 57.6).}
\label{tab:context_holitom}
\scriptsize
\setlength{\tabcolsep}{4pt}
\begin{tabular}{lccccccc}
\toprule
\textbf{Method} & \textbf{Tokens kept} & \textbf{FLOPs} & \textbf{MVBench} & \textbf{EgoSch.} & \textbf{LongVideoBench} & \textbf{V-MME w/o} & \textbf{Avg.\ \%} \\
\midrule
LLaVA-OV-7B (base) & 100\% & 100\% & 58.3 & 60.4 & 56.4 & 58.6 & 100 \\
\midrule
DyCoke~\cite{tao2024dycoke}       & 25\% & 21.3\% & 53.1 & 59.5 & 49.5 & 54.3 & 92.6 \\
VisionZip~\cite{yang2024visionzip} & 25\% & 21.3\% & 57.9 & 60.3 & 56.5 & 58.2 & 99.7 \\
PruneVid~\cite{huang-etal-2025-prunevid} & 25\% & 21.3\% & 57.4 & 59.9 & 55.7 & 57.4 & 98.6 \\
FastVID$^m$~\cite{shen2025fastvid} & 25\% & 21.3\% & 56.5 & --- & 56.3 & 58.0 & --- \\
HoliTom~\cite{shao2025holitom}    & 25\% & 17.4\% & 58.4 & 61.2 & 56.7 & 58.9 & 100.7 \\
\midrule
VisionZip~\cite{yang2024visionzip} & 10\% & 8.3\%  & 53.5 & 58.0 & 49.3 & 53.4 & 91.6 \\
PruneVid~\cite{huang-etal-2025-prunevid} & 10\% & 8.3\%  & 56.2 & 59.8 & 54.5 & 56.0 & 96.9 \\
FastVID$^m$~\cite{shen2025fastvid} & 10\% & 8.3\% & 55.9 & --- & 56.3 & 57.3 & --- \\
HoliTom~\cite{shao2025holitom}    & 10\% & 6.9\%  & 57.3 & 61.2 & 56.3 & 56.8 & 99.1 \\
\midrule
FlashVID~\cite{fan2026flashvid}   & 25\% & ---    & 58.0 & 60.4 & 56.8 & 59.2 & 100.3 \\
EarlyTom~\cite{wang2026earlytom}  & 25\% & 44.2\%$^e$ & 57.4 & 60.5 & 56.3 & 58.5 & 99.7 \\
FlashVID~\cite{fan2026flashvid}   & 10\% & ---    & 57.4 & 60.0 & 56.5 & 57.8 & 99.1 \\
EarlyTom~\cite{wang2026earlytom}  & 10\% & 39.0\%$^e$ & 56.5 & 60.1 & 52.4 & 55.8 & 96.2 \\
VidCom2$^m$~\cite{liu2025vidcom2} & 25\% & ---    & 57.2 & 59.7 & 54.9 & 58.6 & 99.6$^p$ \\
MMG-Vid$^m$~\cite{ma2025mmgvid}   & 25\% & ---    & 56.7 & ---  & 56.6 & 58.6 & 99.5$^p$ \\
\bottomrule
\end{tabular}
\end{table*}

\subsection{LLM Execution and State}
\label{subsubsec:llm_side}
The final pipeline stage targets already-projected visual tokens inside the language model, where they dominate the context length $L$ that drives prefilling cost and KV-cache memory (Section~\ref{subsec:bottlenecks}). We distinguish token pruning and merging, sparse attention, learned summary tokens, KV compaction, and KV offloading or retrieval.

\subsubsection{Decoder Token Pruning and Merging}
The first family reduces how many visual tokens propagate through the decoder layers. FastV~\cite{chen2024fastv} shows that visual tokens receive little attention in deeper decoder layers and exploits this finding by pruning the lowest-attention half after an early layer, roughly halving prefilling FLOPs. HieraVid's controlled video re-run at $39.3\%$ FLOPs costs 3--5 points across MVBench, NExT-QA, EgoSchema and Video-MME (Table~\ref{tab:llmside_hieravid}). SparseVLM~\cite{zhang2024sparsevlm} performs progressive, text-guided pruning across decoder layers, retaining fewer than $10\%$ of the visual tokens, while a recycling step compresses selected pruned tokens into a smaller set of representative tokens. FrameFusion~\cite{fu2025framefusion} and HieraVid~\cite{hieravid2025} specialize the idea for video by first merging temporally redundant tokens across frames and only then pruning by importance: FrameFusion as a two-phase merge-then-prune cascade reporting $1.6$--$3.6\times$ end-to-end speed-ups, and HieraVid as a three-level segment/frame/layer hierarchy that cuts prefilling FLOPs to roughly a quarter of baseline at $30\%$ token retention while retaining about $98\%$ of average accuracy. As decoder backbones themselves diversify, Jiang \textit{et al.}~\cite{jiang2026stateful} extend the family to Mamba--Transformer hybrids~\cite{lieberJambaHybridTransformerMamba2024}: they show that recurrent state layers compress the information carried by removed tokens into their hidden state, and their progressive, query-conditioned schedule yields a $3.8$--$4.2\times$ prefilling speed-up at a $25\%$ token budget at near-baseline accuracy, improving with light finetuning. In the same hybrid direction, TimeViper~\cite{xu2025timeviper} folds visual-token information into the instruction tokens at two decoder depths and drops the visual tokens, reaching over 10{,}000 frames with a $15.7$\% shorter prefill at 4{,}096 frames for a 1--2 point accuracy cost.
STTM~\cite{hyun2025sttm} merges quadtree-derived spatial tokens across time at an early LLM layer. AdaTP~\cite{sun-etal-2025-adatp} corrects attention-sink and positional biases in pruning scores, retaining baseline accuracy at $27\%$ of FLOPs. PruneVid~\cite{huang-etal-2025-prunevid}, HoliTom~\cite{shao2025holitom}, FlashVID~\cite{fan2026flashvid}, VideoChat-Flash~\cite{li2025videochatflash}, HieraVid~\cite{hieravid2025}, and MeToM~\cite{wu2026metom} combine reduction before the LLM with reduction inside its layers.

\subsubsection{Sparse Decoder Attention}
MMInference~\cite{li2025mminference} accelerates prefilling by skipping attention pairs while retaining the token sequence. A modality-aware permutation gathers the grid-structured sparse attention induced by video into GPU-friendly blocks, yielding up to $8.3\times$ prefilling speed-up at million-token contexts with at most 0.4-point accuracy differences across the reported 7B hosts. ReKV's sliding-window attention~\cite{di2025rekv} also restricts the attended context during stream encoding, combined with cache offloading and retrieval below.

\subsubsection{LLM-Computed Summary Tokens}
VoCo-LLaMA~\cite{yeVoCoLLaMAVisionCompression2024} learns compression tokens whose representations are computed by the LLM's own layers under an attention constraint. Subsequent processing uses these compact summaries in place of the full visual context. This differs from a Q-Former or Perceiver resampler operating before the language model. Video-XL~\cite{shu2024videoxl} condenses each interval's visual KV pairs into summarization tokens inside the LLM, reaching 2{,}048 frames on one A100, with successors pushing past 10{,}000 frames through reconstructive compression and task-aware KV sparsification~\cite{liu2025videoxlpro,qin2025videoxl2}.

\subsubsection{KV-Cache Compaction}

KV compaction reduces stored state through eviction, merging or quantization. Eviction removes entries; quantization reduces the precision of those retained. DyCoke~\cite{tao2024dycoke} dynamically evicts the least-attended visual tokens from the KV cache at each decode step, on top of a prefilling temporal-merging stage, for a $1.5\times$ inference speed-up and $1.4\times$ memory reduction against its baseline VideoLLM. VidKV~\cite{tao2025vidkv} quantizes the visual KV cache to mixed precision ($ \approx 1.5$-bit keys and $1.58$-bit values) and finds that, unlike text LLMs, the value cache of video models is better quantized per channel than per token, with almost no performance drop against FP16 on six benchmarks with LLaVA-OneVision-7B~\cite{li2024llavaonevision} and Qwen2.5-VL-7B~\cite{bai2025qwen25vltechnicalreport}. ReTaKe~\cite{wang2024retake} couples keyframe-level pruning (DPSelect) with pivot-guided KV eviction (PivotKV) for $8\times$ context compression, fitting 2{,}048 frames into a 16K context on Qwen2-VL-7B with a $20\%$ lower time-per-output-token, and AdaReTaKe~\cite{wang2025adaretake} adapts the compression ratio across time and layers. MEDA~\cite{wan2025meda} allocates per-layer KV budgets from cross-modal attention entropy, reaching $72\%$ KV-memory reduction and $2.82\times$ faster decoding on multimodal long-context suites. InfiniPot-V~\cite{kim2025infinipotv} evicts entries by temporal redundancy and value norms whenever its budget fills, reporting up to $94\%$ lower peak GPU memory. StreamMem~\cite{yang2025streammem} uses attention from generic proxy queries to compress the cache without the eventual user question, while VideoScan~\cite{li2025videoscan} learns which KV state to propagate alongside its pooled carrier inputs. Image-only MLLMs have parallel lines of work on visual-token withdrawal~\cite{linBoostingMultimodalVisualTokensWithdrawal2025}, layer-wise dropping~\cite{xingPyramidDropAcceleratingLarge2025} and KV eviction~\cite{wanLOOKMLookOnceOptimization2024}, which several of the video methods above adapt.

\subsubsection{KV-Cache Offloading and Retrieval}
Offloading preserves historical state outside GPU memory, then retrieves only the relevant portion for answering. ReKV~\cite{di2025rekv} encodes the stream with sliding-window attention, offloads KV blocks to CPU RAM or disk, and retrieves query-relevant blocks at question time. StreamKV~\cite{chen2025streamkv} combines per-segment cache compression with question-conditioned retrieval. StreamingTOM~\cite{streamingTOM2025} stores quantized groups and selectively dequantizes relevant groups at generation time. These methods can bound active GPU state while allowing total stored history to grow. Table~\ref{tab:streaming_kv} compares the reported streaming and offline long-video protocols.

\begin{table*}[!tp]
\centering
\caption{Streaming memory systems, including representation memory and LLM KV state, under the two protocols the literature shares. \textbf{Top:} offline long-video QA on a shared \textbf{Qwen2-VL-7B} backbone against its full-KV baseline. Baseline reproductions drift with frame count (Video-MME w/o 63.3--63.9, MLVU 63.9--65.8, LongVideoBench 55.6--58.8); each method is judged against its own reproduction, so cross-row gaps within a point are not meaningful. \textbf{Bottom:} streaming QA on a shared \textbf{LLaVA-OneVision-7B} backbone (RVS-Ego / RVS-Movie~\cite{zhang2025flashvstream}), reproduced under one protocol by StreamMem~\cite{yang2025streammem}; peak memory is for a 1-hour 0.5-FPS stream where reported.}
\label{tab:streaming_kv}
\scriptsize
\setlength{\tabcolsep}{5pt}
\begin{tabular}{lccccc}
\toprule
\multicolumn{6}{l}{\textit{Offline long video, Qwen2-VL-7B}} \\
\textbf{Method} & \textbf{KV budget} & \textbf{V-MME w/o} & \textbf{MLVU} & \textbf{LongVideoBench} & \textbf{EgoSch.} \\
\midrule
Full KV cache (range of reproductions) & 100\% & 63.3--63.9 & 63.9--65.8 & 55.6--58.8 & 65.2 \\
ReTaKe~\cite{wang2024retake} & $8\times$ compr. & 63.9 & 69.8 & 57.7 & --- \\
InfiniPot-V~\cite{kim2025infinipotv} & 6K tokens & 62.8 & 65.8 & 58.4 & 65.6 \\
StreamMem~\cite{yang2025streammem} & 6K tokens & 62.1 & 65.9 & --- & 67.2 \\
\midrule
\multicolumn{6}{l}{\textit{Streaming QA, LLaVA-OneVision-7B (RVS-Ego / RVS-Movie), StreamMem reproduction}} \\
\textbf{Method} & \textbf{Mechanism} & \textbf{RVS-Ego} & \textbf{RVS-Movie} & \multicolumn{2}{c}{\textbf{Peak memory}} \\
\midrule
Full KV / backbone baseline & --- & 56.2--60.1 & 43.0--53.4 & \multicolumn{2}{c}{37.5~GB} \\
ReKV~\cite{di2025rekv} & KV offload + retrieval & 63.7 & 54.4 & \multicolumn{2}{c}{38~GB$^{o}$} \\
Flash-VStream~\cite{zhang2025flashvstream} & learned fixed memory & 57.0 & 53.1 & \multicolumn{2}{c}{---} \\
InfiniPot-V~\cite{kim2025infinipotv} & capped KV eviction & 57.9 & 51.4 & \multicolumn{2}{c}{27.8~GB} \\
StreamMem~\cite{yang2025streammem} & query-agnostic KV memory & 57.6 & 52.7 & \multicolumn{2}{c}{$<$28~GB$^{c}$} \\
\bottomrule
\end{tabular}

\vspace{2pt}
\raggedright\footnotesize $^{o}$GPU-resident peak with internal retrieval; ReKV additionally offloads 18.8~GB per stream-hour to CPU RAM or disk. $^{c}$Reported as the experiment's memory constraint rather than a measured peak.
\end{table*}

\subsubsection{Discussion and Synthesis}
Cross-method comparison at this stage is intrinsically limited: each method reports retained accuracy against its own backbone and baseline at a different token budget. Table~\ref{tab:llmside_hieravid} gives the one controlled same-backbone comparison available, covering methods with decoder-layer reduction on LLaVA-Video-7B~\cite{zhangLLaVAVideoVideoInstruction2025}: HieraVid~\cite{hieravid2025} uses 24.5\% of the baseline prefilling FLOPs while remaining within 0.2--2.1 points across the five reported settings, outperforming FastV~\cite{chen2024fastv} at a larger budget and FrameFusion~\cite{fu2025framefusion} at a similar one, supporting temporal merging and pruning under this particular protocol. HieraVid also reduces tokens before the LLM, so this comparison does not isolate its decoder operation. Decoder-layer reduction primarily cuts prefilling computation, whereas visual KV-cache compression targets memory and latency during decoding, so the two mechanisms compose; KV eviction, merging and quantization have different effects on available evidence, and their published results do not share the protocol of Table~\ref{tab:llmside_hieravid}. The streaming systems of Table~\ref{tab:streaming_kv} expose a retrieval--eviction trade-off: ReKV's~\cite{di2025rekv} retrieval preserves streaming accuracy best but keeps peak GPU memory near the full-cache level, whereas hard-capped eviction (InfiniPot-V~\cite{kim2025infinipotv}, StreamMem~\cite{yang2025streammem}) trades roughly six RVS-Ego~\cite{zhang2025flashvstream} points for a constant memory ceiling about 10~GB lower. Sparse-attention prefilling (MMInference~\cite{li2025mminference}) reduces a different part of the workload. Combining it with token or cache reduction requires checking compatibility and measuring the joint system; separate speedups cannot be multiplied.

\begin{table*}[t]
\centering
\caption{\colorbox{color-llm}{Decoder-layer visual-token pruning} on a shared \textbf{LLaVA-Video-7B} backbone, as
re-run by HieraVid \cite{hieravid2025} at matched ${\sim}30\%$ token budgets (FastV runs at a
larger $39.3\%$ FLOPs budget). ``FLOPs'' is prefilling FLOPs relative to the unpruned model;
accuracy is \%.}
\label{tab:llmside_hieravid}
\scriptsize
\setlength{\tabcolsep}{4pt}
\begin{tabular}{lcccccc}
\toprule
\textbf{Method} & \textbf{FLOPs} & \textbf{MVBench} & \textbf{NExT-QA} & \textbf{EgoSch.} & \textbf{VME w/o} & \textbf{VME w/} \\
\midrule
LLaVA-Video (base) & 100\% & 60.4 & 80.2 & 59.4 & 64.1 & 71.4 \\
FastV \cite{chen2024fastv}        & 39.3\% & 56.6 & 77.2 & 55.1 & 59.3 & 66.7 \\
FrameFusion \cite{fu2025framefusion} & 23.8\% & 56.7 & 78.8 & 56.8 & 61.9 & 70.1 \\
HieraVid \cite{hieravid2025}      & 24.5\% & 58.3 & 79.9 & 59.2 & 62.3 & 70.8 \\
\bottomrule
\end{tabular}
\end{table*}

\section{Discussion and Future Directions}
\label{sec:discussion}

\noindent\textbf{Convergent trends across mechanisms.}
Several mechanism families report near-baseline accuracy at \emph{25\% visual-token retention}: pixel-shuffle~\cite{chenExpandingPerformanceBoundaries2025}, pooling~\cite{xuPLLaVAParameterfreeLLaVA2024}, temporal compression~\cite{jiangSTORMTokenEfficientLong2025}, audio-guided pruning~\cite{li2026dash}, and decoder-layer hierarchies~\cite{hieravid2025}. These results come from different hosts and protocols, so they do not imply that every VideoLLM can discard 75\% of its visual tokens without loss; controlled results in Table~\ref{tab:context_holitom} show that method choice is more important at 10\% retention.

Two patterns emerge from placing methods by where they remove computation. First, selection is not always upstream: query-conditioned selectors such as Frame-Voyager~\cite{yuFrameVoyagerLearningQuery2025} and FlexSelect~\cite{zhang2025flexselect} encode every candidate before choosing, so they shorten the LLM context but spare no encoder work, and Frame-Voyager reports higher latency than uniform sampling. Second, the strongest 2025--2026 results combine stages, reducing tokens before the LLM and again inside it~\cite{huang-etal-2025-prunevid,shao2025holitom,fan2026flashvid,hieravid2025,wu2026metom}. Their end-to-end gains cannot be attributed to either stage, and only HoliTom~\cite{shao2025holitom} reports the ablation that separates them. A nominal reduction ratio therefore says little about where the savings occur or which component produced them.

Cross-stage bottleneck shifts are a shared observation in prior efficiency surveys: Zhang \textit{et al.}~\cite{zhang2026efficientinference} analyze interactions among encoding, prefilling and decoding, while Wu \textit{et al.}~\cite{wu2026compressionlifecycle} discuss global resource allocation across compression stages. The video-specific evidence reviewed here shows how this interaction affects the choice between encoding fewer frames, representing each frame more cheaply, and compressing the resulting context. Once downstream compression reduces the LLM token load, recent methods move upstream again by pruning inside the encoder \cite{wang2026earlytom}, caching features across similar frames \cite{wang2025stc}, or consuming codec primitives \cite{sarkar2026cope}.

Some methods reinvest the saved compute: under a fixed LLM budget, token reduction can admit $10$--$20\times$ more frames \cite{li2026echoprune,fan2026flashvid}, allocate tokens continuously across frames \cite{li2025dytok}, or support longer training contexts \cite{jiang2026stateful}, so compression may improve accuracy by increasing temporal coverage. Evidence for audiovisual efficiency remains comparatively sparse: reported costs often omit the audio encoder and modality ablations are uncommon, making the benefit and cost of audio-guided selection hard to isolate.

The four stages are unevenly represented: reduction of encoded representations attracts the most papers, while the LLM-side families are small. 
The first connector-side and decoder-side reductions applied to video were image-only VLM methods evaluated frame by frame, such as FastV~\cite{chen2024fastv}, SparseVLM~\cite{zhang2024sparsevlm}, VisionZip~\cite{yang2024visionzip}, and LLaVA-PruMerge~\cite{shang2024prumerge}. In contrast, the 2025–2026 methods in the same families exploit temporal redundancy directly, merging tokens across frames~\cite{fu2025framefusion,shao2025holitom,liu2025vidcom2} or evicting cache entries by inter-frame similarity~\cite{tao2024dycoke,wang2024retake}. The LLM-side families have not completed this move: KV eviction, quantization and sparse attention were developed for text-only LLMs and transfer to video with little modification, so fewer video-specific papers are needed to cover the same ground. Yet once pre-LLM compression has removed redundant tokens, the cost that remains is decoding memory and cache growth under multi-turn and streaming use, which only LLM-side mechanisms address, so we expect these families to grow fastest. Decoder backbones are also diversifying beyond dense transformers~\cite{jiang2026stateful,xu2025timeviper,lieberJambaHybridTransformerMamba2024}, making evidence preservation across attention and recurrent state a further evaluation question.

\noindent\textbf{From action recognition to question answering.}
Read chronologically, the comparison tables of Section~\ref{sec:efficiency-mechanisms} document a shift both in \emph{how} efficiency is achieved and in how it is \emph{evidenced}. Vision-encoder methods (Table~\ref{tab:backbone_efficiency}), proposed mostly between 2019 and 2023, evaluate on action recognition (Kinetics~\cite{kayKineticsHumanAction2017}, Moments in Time~\cite{monfortMomentsTimeDataset2019}, UCF101~\cite{soomroUCF101Dataset101}) and report inference GFLOPs per view under an explicit input protocol, whereas the later pipeline stages evaluate almost exclusively on video question answering (Tables~\ref{tab:frame_sampling_isobackbone}, \ref{tab:context_reduction} and~\ref{tab:llmside_hieravid} contain no classification benchmark), moving from GPT-assisted open-ended scoring in 2023--2024 to cheaper, less judge-dependent multiple choice in 2025--2026. The shift is in fact stage-dependent and not purely chronological: state-space encoder papers from 2025 still evaluate on Kinetics \cite{luSnakesLadders2025}. Because each stage evaluates on different tasks and metrics, efficiency progress cannot be compared consistently across years, and an efficient Kinetics backbone does not by itself establish end-to-end VideoLLM efficiency.

\noindent\textbf{Prioritized research agenda.}
\emph{1) Establish a common analytical protocol.} The most urgent need is a reproducible accuracy--compute protocol: candidate methods processing the same videos, prompts and modality inputs under fixed resolution and decoding settings, plug-in methods additionally sharing a frozen host, candidate-frame pool and frame or token budget. The protocol should report encoder, connector and LLM prefilling FLOPs under fixed accounting boundaries, including the cost of selection or allocation itself, together with retained-token counts and task performance both on the same input and under the same compute budget. FLOPs provide a hardware-independent common denominator without standing in for deployment speed; latency, memory and energy remain useful deployment measurements, but meaningful comparison requires a fixed hardware--software stack and measurement boundary~\cite{reddiMLPerfInferenceBenchmark2020, tschandMLPerfPowerBenchmarking2024}, and aggregating values from different stacks would create false precision. The emerging LLaVA-OneVision-7B, 32-frame, LMMs-Eval setup~\cite{zhang-etal-2025-lmms,shao2025holitom,wang2026earlytom,li2026echoprune,li2025dytok} is a practical starting point.

\emph{2) Report across video domains and task families.} Efficiency results are reported almost exclusively as one aggregate accuracy on multiple-choice QA. Video-MME~\cite{fuVideoMMEFirstEverComprehensive2025} annotates content domain as well as duration, yet most surveyed methods report only the aggregate, so a token budget validated on static lecture footage is indistinguishable from one validated on fast-cut sports. No comparison table in this survey contains a captioning, retrieval or grounding metric, and the one selector evaluated on captioning~\cite{steunou2026peek} had to be excluded from Table~\ref{tab:frame_sampling_isobackbone} for that reason. Multiple-choice questions supply the candidate answers and can often be settled by coarse object and scene cues; captions and temporal boundaries must be produced from finer detail. A retention ratio that is lossless on MCQ therefore need not be lossless on generation. Reporting per domain and re-testing one budget on a generation or localization task would check both assumptions cheaply.

\emph{3) Learn when audio should influence compression.} Omni-modal models process synchronized audio and video~\cite{chenSurveyOmnimodalLanguage2025,xuQwen25OmniTechnicalReport2025,yeOmniVinciEnhancingArchitecture2025}, but the efficiency literature remains predominantly visual. OmniZip~\cite{taoOmniZipAudioGuidedDynamic2025} and DASH~\cite{li2026dash} show that audio can guide visual-token reduction, yet their audio anchor may help in one segment and mislead in another: speech may refer to an off-screen event, while a visible event may have no informative sound. A stronger direction is a learned, query- and context-dependent allocation across modalities, evaluated on visual-only, audio-only, jointly answerable and deliberately conflicting examples, with the audio encoder and allocation module included in the cost.

\section{Conclusion}
\label{sec:conclusion}

This survey organized efficiency mechanisms for VideoLLMs by the stage of the encoder--connector--LLM pipeline at which they act: input construction and selection, encoder computation, encoded representations and connector, and LLM execution and state. Efficiency emerges from system-level trade-offs among semantic performance, input coverage, compute, latency and memory. Across the heterogeneous evidence reviewed here, retaining roughly one quarter of the visual-token budget often preserves near-baseline accuracy, though the achievable reduction depends on the host model, task and evaluation protocol. Consistent with prior pipeline analyses~\cite{zhang2026efficientinference,wu2026compressionlifecycle}, reducing LLM prefilling and cache costs can make vision encoding the limiting stage. In video systems, saved compute can also be reinvested in processing more frames, so gains must be interpreted together with temporal coverage and the cost of encoding those frames. Progress now requires a reproducible accuracy--compute protocol with common backbones, inputs and FLOP-accounting boundaries, with methods compared both on the same inputs and under the same compute budget and complemented by system measurements on a shared reference stack. Without it, reported gains remain difficult to compare across papers and to reproduce on a deployment target. Among mechanism directions, learned audiovisual allocation is especially promising because audio is widely available in video but still weakly represented in the efficiency literature.

\bibliographystyle{IEEEtran}
\bibliography{main}

@IEEEtranBSTCTL{IEEEbstctl,
  CTLuse_forced_etal       = "yes",
  CTLmax_names_forced_etal = "6",
  CTLnames_show_etal       = "1"
}

@inproceedings{tongVideoMAEMaskedAutoencoders2022,
  author =        {Tong, Zhan and Song, Yibing and Wang, Jue and
                   Wang, Limin},
  title =         {{VideoMAE}: {Masked} {Autoencoders} are
                   {Data}-{Efficient} {Learners} for {Self}-{Supervised}
                   {Video} {Pre}-{Training}},
  year =          {2022},
  doi =           {10.48550/arXiv.2203.12602},
  booktitle = {Proc. NeurIPS}
}

@article{tangVideoUnderstandingLarge2023,
  author =        {Tang, Yolo Yunlong and Bi, Jing and Xu, Siting and
                   Song, Luchuan and Liang, Susan and Wang, Teng and
                   Zhang, Daoan and An, Jie and Lin, Jingyang and
                   Zhu, Rongyi and Vosoughi, Ali and Huang, Chao and
                   Zhang, Zeliang and Liu, Pinxin and Feng, Mingqian and
                   Zheng, Feng and Zhang, Jianguo and Luo, Ping and
                   Luo, Jiebo and Xu, Chenliang},
  title =         {Video {{Understanding}} with {{Large Language
                   Models}}: {{A Survey}}},
  year =          {2026},
  journal = {IEEE Transactions on Circuits and Systems for Video Technology},
  volume = {36},
  number = {2},
  pages = {1355--1376},
  doi = {10.1109/TCSVT.2025.3566695}
}

@misc{madanFoundationModelsVideo2024,
  author =        {Madan, Neelu and Moegelmose, Andreas and Modi, Rajat and
                   Rawat, Yogesh S. and Moeslund, Thomas B.},
  title =         {Foundation {{Models}} for {{Video Understanding}}:
                   {{A Survey}}},
  year =          {2024},
  howpublished = {arXiv preprint arXiv:2405.03770}
}

@inproceedings{nguyenVideoLanguageUnderstandingSurvey2024,
  author =        {Nguyen, Thong and Bin, Yi and Xiao, Junbin and
                   Qu, Leigang and Li, Yicong and Wu, Jay Zhangjie and
                   Nguyen, Cong-Duy and Ng, See-Kiong and Tuan, Luu Anh},
  journal =       {arXiv.org},
  title =         {Video-{{Language Understanding}}: {{A Survey}} from
                   {{Model Architecture}}, {{Model Training}}, and
                   {{Data Perspectives}}},
  year =          {2024},
  booktitle = {Findings of the Association for Computational Linguistics ACL 2024}
}

@article{faragVideoCaptioningUsing2026,
  author =        {Farag, Mohamed Ali and Khafagy, Mohamed H. and
                   Hussien, Shereen A.},
  journal =       {Franklin Open},
  pages =         {100497},
  title =         {Video {Captioning} using {Deep} {Learning} with
                   {Greedy} {Search} ({VCDLGS})},
  volume =        {14},
  year =          {2026},
  doi =           {10.1016/j.fraope.2026.100497},
  issn =          {2773-1863},
}

@article{yinSurveyMultimodalLarge2024,
  author =        {Yin, Shukang and Fu, Chaoyou and Zhao, Sirui and
                   Li, Ke and Sun, Xing and Xu, Tong and Chen, Enhong},
  journal =       {National Science Review},
  number =        {12},
  pages =         {nwae403},
  title =         {A {{Survey}} on {{Multimodal Large Language Models}}},
  volume =        {11},
  year =          {2024},
  doi =           {10.1093/nsr/nwae403},
  issn =          {2095-5138, 2053-714X},
}

@inproceedings{wengLongVLMEfficientLong2024,
  author =        {Weng, Yuetian and Han, Mingfei and He, Haoyu and
                   Chang, Xiaojun and Zhuang, Bohan},
  title =         {{{LongVLM}}: {{Efficient Long Video Understanding}}
                   via {{Large Language Models}}},
  year =          {2024},
  doi =           {10.48550/arXiv.2404.03384},
  booktitle = {Proc. ECCV}
}

@inproceedings{chatterjeeMemoryefficientStreamingVideoLLMs2025,
  author =        {Chatterjee, Dibyadip and Remelli, Edoardo and
                   Song, Yale and Tekin, Bugra and Mittal, Abhay and
                   Bhatnagar, Bharat and Camg{\"o}z, Necati Cihan and
                   Hampali, Shreyas and Sauser, Eric and Ma, Shugao and
                   Yao, Angela and Sener, Fadime},
  title =         {Memory-Efficient {{Streaming VideoLLMs}} for
                   {{Real-time Procedural Video Understanding}}},
  year =          {2025},
  doi =           {10.48550/arXiv.2504.13915},
  booktitle = {Proc. ICCV}
}

@misc{ningLiveVLM2025,
  author =        {Ning, Zhenyu and Liu, Guangda and Jin, Qihao and
                   Li, Chengwei and Ding, Wenchao and Guo, Minyi and
                   Zhao, Jieru},
  title =         {LiveVLM: Efficient Online Video Understanding via
                   Streaming-Oriented KV Cache and Retrieval},
  year =          {2025},
  doi =           {10.48550/ARXIV.2505.15269},
  howpublished = {arXiv preprint arXiv:2505.15269}
}

@inproceedings{bhardwajEfficientVideoClassification2019,
  address =       {Long Beach, CA, USA},
  author =        {Bhardwaj, Shweta and Srinivasan, Mukundhan and
                   Khapra, Mitesh M.},
  booktitle =     {2019 {{IEEE}}/{{CVF Conference}} on {{Computer
                   Vision}} and {{Pattern Recognition}} ({{CVPR}})},
  pages =         {354--363},
  publisher =     {IEEE},
  title =         {Efficient {{Video Classification Using Fewer
                   Frames}}},
  year =          {2019},
  doi =           {10.1109/CVPR.2019.00044},
  isbn =          {978-1-7281-3293-8},
}

@inproceedings{huang-etal-2025-prunevid,
  author =        {Huang, Xiaohu and Zhou, Hao and Han, Kai},
  booktitle =     {Findings of ACL},
  pages =         {19959--19973},
  title =         {{PruneVid}: Visual Token Pruning for Efficient Video
                   Large Language Models},
  year =          {2025},
  doi =           {10.18653/v1/2025.findings-acl.1024},
}

@article{li2026echoprune,
  author =        {Jiameng Li and Minye Wu and Jiezhang Cao and
                   Aleksei Tiulpin and Matthew B. Blaschko},
  title =         {{EchoPrune}: Interpreting Redundancy as Temporal
                   Echoes for Efficient {VideoLLMs}},
  year =          {2026},
  journal =       {arXiv preprint arXiv:2605.10050},
}

@article{tao2025vidkv,
  author =        {Tao, Keda and You, Haoxuan and Sui, Yang and Qin, Can and
                   Wang, Huan},
  journal =       {arXiv preprint arXiv:2503.16257},
  title =         {Plug-and-Play 1.x-Bit {KV} Cache Quantization for
                   Video Large Language Models},
  year =          {2025},
}

@misc{taoOmniZipAudioGuidedDynamic2025,
  author =        {Tao, Keda and Shao, Kele and Yu, Bohan and
                   Wang, Weiqiang and {liu}, Jian and Wang, Huan},
  title =         {{{OmniZip}}: {{Audio-Guided Dynamic Token
                   Compression}} for {{Fast Omnimodal Large Language
                   Models}}},
  year =          {2025},
  doi =           {10.48550/arXiv.2511.14582},
  howpublished = {arXiv preprint arXiv:2511.14582}
}

@misc{taoOmniAgentAudioGuidedActive2025,
  author =        {Tao, Keda and Du, Wenjie and Yu, Bohan and
                   Wang, Weiqiang and Liu, Jian and Wang, Huan},
  title =         {{{OmniAgent}}: {{Audio-Guided Active Perception
                   Agent}} for {{Omnimodal Audio-Video Understanding}}},
  year =          {2025},
  doi =           {10.48550/arXiv.2512.23646},
  howpublished = {arXiv preprint arXiv:2512.23646}
}

@misc{zouSecondsHoursReviewing2024,
  author =        {Zou, Heqing and Luo, Tianze and Xie, Guiyang and
                   Victor and Zhang and Lv, Fengmao and Wang, Guangcong and
                   Chen, Junyang and Wang, Zhuochen and Zhang, Hansheng and
                   Zhang, Huaijian},
  title =         {From {{Seconds}} to {{Hours}}: {{Reviewing MultiModal
                   Large Language Models}} on {{Comprehensive Long Video
                   Understanding}}},
  year =          {2024},
  howpublished = {arXiv preprint arXiv:2409.18938}
}

@article{wuSurveyVideoTemporal2025,
  author =        {Wu, Jianlong and Liu, Wei and Liu, Ye and Liu, Meng and
                   Nie, Liqiang and Lin, Zhouchen and Chen, Chang Wen},
  title =         {A {{Survey}} on {{Video Temporal Grounding}} with
                   {{Multimodal Large Language Model}}},
  year =          {2026},
  doi =           {10.48550/arXiv.2508.10922},
  journal = {IEEE Transactions on Pattern Analysis and Machine Intelligence}
}

@misc{kumarVideoLLMBenchmarksEvaluation2025,
  author =        {Kumar, Yogesh},
  title =         {{{VideoLLM Benchmarks}} and {{Evaluation}}: {{A
                   Survey}}},
  year =          {2025},
  doi =           {10.48550/arXiv.2505.03829},
  howpublished = {arXiv preprint arXiv:2505.03829}
}

@article{chenSurveyOmnimodalLanguage2025,
  author =        {Chen, Lu and Mu, Jiajie and Wang, Jiarui and
                   Kang, Xiao and Xi, Xiaoming and Qin, Zheyun},
  journal =       {AI+},
  title =         {A Survey on Omni-Modal Language Models},
  year =          {2025},
  doi =           {10.55092/aiplus20260001},
  issn =          {3007-7443, 3007-7451},
}

@misc{baiSurveyMultimodalLarge2024,
  author =        {Bai, Tianyi and Liang, Hao and Wan, Binwang and
                   Xu, Yanran and Li, Xi and Li, Shiyu and Yang, Ling and
                   Li, Bozhou and Wang, Yifan and Cui, Bin and
                   Huang, Ping and Shan, Jiulong and He, Conghui and
                   Yuan, Binhang and Zhang, Wentao},
  title =         {A {{Survey}} of {{Multimodal Large Language Model}}
                   from {{A Data-centric Perspective}}},
  year =          {2024},
  doi =           {10.48550/arXiv.2405.16640},
  howpublished = {arXiv preprint arXiv:2405.16640}
}

@inproceedings{caffagniRevolutionMultimodalLarge2024,
  author =        {Caffagni, Davide and Cocchi, Federico and
                   Barsellotti, Luca and Moratelli, Nicholas and
                   Sarto, Sara and Baraldi, Lorenzo and Baraldi, Lorenzo and
                   Cornia, Marcella and Cucchiara, Rita},
  title =         {The {{Revolution}} of {{Multimodal Large Language
                   Models}}: {{A Survey}}},
  year =          {2024},
  booktitle = {Findings of ACL}
}

@misc{jinEfficientMultimodalLarge2024,
  author =        {Jin, Yizhang and Li, Jian and Liu, Yexin and
                   Gu, Tianjun and Wu, Kai and Jiang, Zhengkai and
                   He, Muyang and Zhao, Bo and Tan, Xin and Gan, Zhenye and
                   Wang, Yabiao and Wang, Chengjie and Ma, Lizhuang},
  title =         {Efficient {{Multimodal Large Language Models}}: {{A
                   Survey}}},
  year =          {2024},
  doi =           {10.48550/arXiv.2405.10739},
  howpublished = {arXiv preprint arXiv:2405.10739}
}

@misc{shaoWhenTokensTalk2025,
  author =        {Shao, Kele and Tao, Keda and Zhang, Kejia and
                   Feng, Sicheng and Cai, Mu and Shang, Yuzhang and
                   You, Haoxuan and Qin, Can and Sui, Yang and
                   Wang, Huan},
  title =         {When {{Tokens Talk Too Much}}: {{A Survey}} of
                   {{Multimodal Long-Context Token Compression}} across
                   {{Images}}, {{Videos}}, and {{Audios}}},
  year =          {2025},
  doi =           {10.48550/arXiv.2507.20198},
  howpublished = {arXiv preprint arXiv:2507.20198}
}

@misc{kayKineticsHumanAction2017,
  author =        {Kay, Will and Carreira, Joao and Simonyan, Karen and
                   Zhang, Brian and Hillier, Chloe and
                   Vijayanarasimhan, Sudheendra and Viola, Fabio and
                   Green, Tim and Back, Trevor and Natsev, Paul and
                   Suleyman, Mustafa and Zisserman, Andrew},
  title =         {The {{Kinetics Human Action Video Dataset}}},
  year =          {2017},
  doi =           {10.48550/arXiv.1705.06950},
  howpublished = {arXiv preprint arXiv:1705.06950}
}

@inproceedings{goyalSomethingSomethingVideo2017,
  author =        {Goyal, Raghav and Kahou, Samira Ebrahimi and
                   Michalski, Vincent and Materzy{\'n}ska, Joanna and
                   Westphal, Susanne and Kim, Heuna and Haenel, Valentin and
                   Fruend, Ingo and Yianilos, Peter and
                   {Mueller-Freitag}, Moritz and Hoppe, Florian and
                   Thurau, Christian and Bax, Ingo and
                   Memisevic, Roland},
  title =         {The "Something Something" Video Database for Learning
                   and Evaluating Visual Common Sense},
  year =          {2017},
  doi =           {10.48550/arXiv.1706.04261},
  booktitle = {Proc. ICCV}
}

@inproceedings{graumanEgo4DWorld30002022,
  author =        {Grauman, Kristen and Westbury, Andrew and
                   Byrne, Eugene and Chavis, Zachary and
                   Furnari, Antonino and Girdhar, Rohit and
                   Hamburger, Jackson and Jiang, Hao and Liu, Miao and
                   Liu, Xingyu and Martin, Miguel and Nagarajan, Tushar and
                   Radosavovic, Ilija and Ramakrishnan, Santhosh Kumar and
                   Ryan, Fiona and Sharma, Jayant and Wray, Michael and
                   Xu, Mengmeng and Xu, Eric Zhongcong and Zhao, Chen and
                   Bansal, Siddhant and Batra, Dhruv and
                   Cartillier, Vincent and Crane, Sean and Do, Tien and
                   Doulaty, Morrie and Erapalli, Akshay and
                   Feichtenhofer, Christoph and Fragomeni, Adriano and
                   Fu, Qichen and Gebreselasie, Abrham and
                   Gonzalez, Cristina and Hillis, James and Huang, Xuhua and
                   Huang, Yifei and Jia, Wenqi and Khoo, Weslie and
                   Kolar, Jachym and Kottur, Satwik and Kumar, Anurag and
                   Landini, Federico and Li, Chao and Li, Yanghao and
                   Li, Zhenqiang and Mangalam, Karttikeya and
                   Modhugu, Raghava and Munro, Jonathan and
                   Murrell, Tullie and Nishiyasu, Takumi and Price, Will and
                   Puentes, Paola Ruiz and Ramazanova, Merey and
                   Sari, Leda and Somasundaram, Kiran and
                   Southerland, Audrey and Sugano, Yusuke and
                   Tao, Ruijie and Vo, Minh and Wang, Yuchen and
                   Wu, Xindi and Yagi, Takuma and Zhao, Ziwei and
                   Zhu, Yunyi and Arbelaez, Pablo and Crandall, David and
                   Damen, Dima and Farinella, Giovanni Maria and
                   Fuegen, Christian and Ghanem, Bernard and
                   Ithapu, Vamsi Krishna and Jawahar, C. V. and
                   Joo, Hanbyul and Kitani, Kris and Li, Haizhou and
                   Newcombe, Richard and Oliva, Aude and Park, Hyun Soo and
                   Rehg, James M. and Sato, Yoichi and Shi, Jianbo and
                   Shou, Mike Zheng and Torralba, Antonio and
                   Torresani, Lorenzo and Yan, Mingfei and
                   Malik, Jitendra},
  title =         {{{Ego4D}}: {{Around}} the {{World}} in 3,000
                   {{Hours}} of {{Egocentric Video}}},
  year =          {2022},
  booktitle = {Proc. CVPR}
}

@inproceedings{xuMSRVTTLargeVideo2016,
  address =       {Las Vegas, NV, USA},
  author =        {Xu, Jun and Mei, Tao and Yao, Ting and Rui, Yong},
  booktitle =     {2016 {{IEEE Conference}} on {{Computer Vision}} and
                   {{Pattern Recognition}} ({{CVPR}})},
  pages =         {5288--5296},
  publisher =     {IEEE},
  title =         {{{MSR-VTT}}: {{A Large Video Description Dataset}}
                   for {{Bridging Video}} and {{Language}}},
  year =          {2016},
  doi =           {10.1109/CVPR.2016.571},
  isbn =          {978-1-4673-8851-1},
}

@inproceedings{krishnaDenseCaptioningEventsVideos2017,
  author =        {Krishna, Ranjay and Hata, Kenji and Ren, Frederic and
                   {Fei-Fei}, Li and Niebles, Juan Carlos},
  booktitle =     {Proceedings of the IEEE International Conference on
                   Computer Vision (ICCV)},
  pages =         {706--715},
  title =         {Dense-{{Captioning Events}} in {{Videos}}},
  year =          {2017},
  doi =           {10.1109/ICCV.2017.83},
}

@article{yuActivityNetQADatasetUnderstanding2019,
  author =        {Yu, Zhou and Xu, Dejing and Yu, Jun and Yu, Ting and
                   Zhao, Zhou and Zhuang, Yueting and Tao, Dacheng},
  title =         {{{ActivityNet-QA}}: {{A Dataset}} for {{Understanding
                   Complex Web Videos}} via {{Question Answering}}},
  year =          {2019},
  doi =           {10.48550/arXiv.1906.02467},
  journal = {Proc. AAAI}
}

@inproceedings{xiaoNExTQANextPhase2021,
  author =        {Xiao, Junbin and Shang, Xindi and Yao, Angela and
                   Chua, Tat-Seng},
  title =         {{{NExT-QA}}:{{Next Phase}} of {{Question-Answering}}
                   to {{Explaining Temporal Actions}}},
  year =          {2021},
  doi =           {10.48550/arXiv.2105.08276},
  booktitle = {Proc. CVPR}
}

@inproceedings{mangalamEgoSchemaDiagnosticBenchmark2023,
  author =        {Mangalam, Karttikeya and Akshulakov, Raiymbek and
                   Malik, Jitendra},
  title =         {{{EgoSchema}}: {{A Diagnostic Benchmark}} for {{Very
                   Long-form Video Language Understanding}}},
  year =          {2023},
  doi =           {10.48550/arXiv.2308.09126},
  booktitle = {Proc. NeurIPS}
}

@inproceedings{miechHowTo100MLearningTextVideo2019,
  author =        {Miech, Antoine and Zhukov, Dimitri and
                   Alayrac, Jean-Baptiste and Tapaswi, Makarand and
                   Laptev, Ivan and Sivic, Josef},
  title =         {{{HowTo100M}}: {{Learning}} a {{Text-Video
                   Embedding}} by {{Watching Hundred Million Narrated
                   Video Clips}}},
  year =          {2019},
  doi =           {10.48550/arXiv.1906.03327},
  booktitle = {Proc. ICCV}
}

@inproceedings{gaoTALLTemporalActivity2017,
  author =        {Gao, Jiyang and Sun, Chen and Yang, Zhenheng and
                   Nevatia, Ram},
  title =         {{{TALL}}: {{Temporal Activity Localization}} via
                   {{Language Query}}},
  year =          {2017},
  doi =           {10.48550/arXiv.1705.02101},
  booktitle = {Proc. ICCV}
}

@inproceedings{liMVBenchComprehensiveMultimodal2024,
  author =        {Li, Kunchang and Wang, Yali and He, Yinan and
                   Li, Yizhuo and Wang, Yi and Liu, Yi and Wang, Zun and
                   Xu, Jilan and Chen, Guo and Luo, Ping and Wang, Limin and
                   Qiao, Yu},
  title =         {{{MVBench}}: {{A Comprehensive Multi-modal Video
                   Understanding Benchmark}}},
  year =          {2024},
  doi =           {10.48550/arXiv.2311.17005},
  booktitle = {Proc. CVPR}
}

@inproceedings{fuVideoMMEFirstEverComprehensive2025,
  author =        {Fu, Chaoyou and Dai, Yuhan and Luo, Yongdong and
                   Li, Lei and Ren, Shuhuai and Zhang, Renrui and
                   Wang, Zihan and Zhou, Chenyu and Shen, Yunhang and
                   Zhang, Mengdan and Chen, Peixian and Li, Yanwei and
                   Lin, Shaohui and Zhao, Sirui and Li, Ke and Xu, Tong and
                   Zheng, Xiawu and Chen, Enhong and Shan, Caifeng and
                   He, Ran and Sun, Xing},
  title =         {Video-{{MME}}: {{The First-Ever Comprehensive
                   Evaluation Benchmark}} of {{Multi-modal LLMs}} in
                   {{Video Analysis}}},
  year =          {2025},
  doi =           {10.48550/arXiv.2405.21075},
  booktitle = {Proc. CVPR}
}

@inproceedings{wuLongVideoBenchBenchmarkLongcontext2024,
  author =        {Wu, Haoning and Li, Dongxu and Chen, Bei and
                   Li, Junnan},
  journal =       {arXiv.org},
  title =         {{{LongVideoBench}}: {{A Benchmark}} for
                   {{Long-context Interleaved Video-Language
                   Understanding}}},
  year =          {2024},
  booktitle = {Proc. NeurIPS}
}

@inproceedings{zhouMLVUBenchmarkingMultitask2025,
  author =        {Zhou, Junjie and Shu, Yan and Zhao, Bo and Wu, Boya and
                   Liang, Zhengyang and Xiao, Shitao and Qin, Minghao and
                   Yang, Xi and Xiong, Yongping and Zhang, Bo and
                   Huang, Tiejun and Liu, Zheng},
  title =         {{{MLVU}}: {{Benchmarking Multi-task Long Video
                   Understanding}}},
  year =          {2025},
  doi =           {10.48550/arXiv.2406.04264},
  booktitle = {Proc. CVPR}
}

@inproceedings{zhangVideoLLaMAInstructiontunedAudioVisual2023,
  author =        {Zhang, Hang and Li, Xin and Bing, Lidong},
  title =         {Video-{{LLaMA}}: {{An Instruction-tuned Audio-Visual
                   Language Model}} for {{Video Understanding}}},
  year =          {2023},
  doi =           {10.48550/arXiv.2306.02858},
  booktitle = {Proceedings of the 2023 Conference on Empirical Methods in Natural Language Processing: System Demonstrations}
}

@inproceedings{radfordLearningTransferableVisual2021,
  author =        {Radford, Alec and Kim, Jong Wook and Hallacy, Chris and
                   Ramesh, Aditya and Goh, Gabriel and Agarwal, Sandhini and
                   Sastry, Girish and Askell, Amanda and Mishkin, Pamela and
                   Clark, Jack and Krueger, Gretchen and
                   Sutskever, Ilya},
  title =         {Learning {{Transferable Visual Models From Natural
                   Language Supervision}}},
  year =          {2021},
  booktitle = {Proc. ICML}
}

@inproceedings{dosovitskiyImageWorth16x162021,
  author =        {Dosovitskiy, Alexey and Beyer, Lucas and
                   Kolesnikov, Alexander and Weissenborn, Dirk and
                   Zhai, Xiaohua and Unterthiner, Thomas and
                   Dehghani, Mostafa and Minderer, Matthias and
                   Heigold, Georg and Gelly, Sylvain and
                   Uszkoreit, Jakob and Houlsby, Neil},
  title =         {An {{Image}} Is {{Worth}} 16x16 {{Words}}:
                   {{Transformers}} for {{Image Recognition}} at
                   {{Scale}}},
  year =          {2021},
  booktitle = {Proc. ICLR}
}

@inproceedings{girdharImageBindOneEmbedding2023,
  author =        {Girdhar, Rohit and {El-Nouby}, Alaaeldin and
                   Liu, Zhuang and Singh, Mannat and
                   Alwala, Kalyan Vasudev and Joulin, Armand and
                   Misra, Ishan},
  title =         {{{ImageBind}}: {{One Embedding Space To Bind Them
                   All}}},
  year =          {2023},
  doi =           {10.48550/arXiv.2305.05665},
  booktitle = {Proc. CVPR}
}

@inproceedings{liBLIP2BootstrappingLanguageImage2023,
  author =        {Li, Junnan and Li, Dongxu and Savarese, Silvio and
                   Hoi, Steven},
  title =         {{{BLIP-2}}: {{Bootstrapping Language-Image
                   Pre-training}} with {{Frozen Image Encoders}} and
                   {{Large Language Models}}},
  year =          {2023},
  booktitle = {Proc. ICML}
}

@inproceedings{VicunaOpenSourceChatbot,
  author = {Zheng, Lianmin and Chiang, Wei-Lin and Sheng, Ying and Zhuang, Siyuan and Wu, Zhanghao and Zhuang, Yonghao and Lin, Zi and Li, Zhuohan and Li, Dacheng and Xing, Eric P. and Zhang, Hao and Gonzalez, Joseph E. and Stoica, Ion},
  title = {Judging {LLM}-as-a-Judge with {MT-Bench} and {Chatbot Arena}},
  booktitle = {Proc. NeurIPS},
  volume = {36},
  pages = {46595--46623},
  year = {2023}
}

@article{liVideoChatChatCentricVideo2024,
  author =        {Li, KunChang and He, Yinan and Wang, Yi and
                   Li, Yizhuo and Wang, Wenhai and Luo, Ping and
                   Wang, Yali and Wang, Limin and Qiao, Yu},
  title =         {{{VideoChat}}: {{Chat-Centric Video Understanding}}},
  year =          {2025},
  doi =           {10.48550/arXiv.2305.06355},
  journal = {Science China Information Sciences}
}

@misc{luoValleyVideoAssistant2025,
  author =        {Luo, Ruipu and Zhao, Ziwang and Yang, Min and
                   Yang, Zheming and Qiu, Minghui and Wang, Tao and
                   Wei, Zhongyu and Wang, Yanhao and Chen, Cen},
  title =         {Valley: {{Video Assistant}} with {{Large Language}}
                   Model {{Enhanced abilitY}}},
  year =          {2025},
  doi =           {10.48550/arXiv.2306.07207},
  howpublished = {arXiv preprint arXiv:2306.07207}
}

@inproceedings{maazVideoChatGPTDetailedVideo2024,
  author =        {Maaz, Muhammad and Rasheed, Hanoona and Khan, Salman and
                   Khan, Fahad Shahbaz},
  title =         {Video-{{ChatGPT}}: {{Towards Detailed Video
                   Understanding}} via {{Large Vision}} and {{Language
                   Models}}},
  year =          {2024},
  doi =           {10.48550/arXiv.2306.05424},
  booktitle = {Proc. ACL}
}

@misc{xuYoukumPLUG10Million2023,
  author =        {Xu, Haiyang and Ye, Qinghao and Wu, Xuan and
                   Yan, Ming and Miao, Yuan and Ye, Jiabo and Xu, Guohai and
                   Hu, Anwen and Shi, Yaya and Xu, Guangwei and
                   Li, Chenliang and Qian, Qi and Que, Maofei and
                   Zhang, Ji and Zeng, Xiao and Huang, Fei},
  title =         {Youku-{{mPLUG}}: {{A}} 10 {{Million Large-scale
                   Chinese Video-Language Dataset}} for {{Pre-training}}
                   and {{Benchmarks}}},
  year =          {2023},
  howpublished = {arXiv preprint arXiv:2306.04362}
}

@inproceedings{xuMPLUG2ModularizedMultimodal2023,
  author =        {Xu, Haiyang and Ye, Qinghao and Yan, Ming and
                   Shi, Yaya and Ye, Jiabo and Xu, Yuanhong and
                   Li, Chenliang and Bi, Bin and Qian, Qi and Wang, Wei and
                   Xu, Guohai and Zhang, Ji and Huang, Songfang and
                   Huang, Fei and Zhou, Jingren},
  title =         {{{mPLUG-2}}: {{A Modularized Multi-modal Foundation
                   Model Across Text}}, {{Image}} and {{Video}}},
  year =          {2023},
  booktitle = {Proc. ICML}
}

@misc{zhangLLaVAVideoVideoInstruction2025,
  author =        {Zhang, Yuanhan and Wu, Jinming and Li, Wei and Li, Bo and
                   Ma, Zejun and Liu, Ziwei and Li, Chunyuan},
  title =         {{{LLaVA-Video}}: {{Video Instruction Tuning With
                   Synthetic Data}}},
  year =          {2025},
  doi =           {10.48550/arXiv.2410.02713},
  howpublished = {arXiv preprint arXiv:2410.02713}
}

@inproceedings{linVideoLLaVALearningUnited2023,
  author =        {Lin, Bin and Ye, Yang and Zhu, Bin and Cui, Jiaxi and
                   Ning, Munan and Jin, Peng and Yuan, Li},
  journal =       {arXiv.org},
  title =         {Video-{{LLaVA}}: {{Learning United Visual
                   Representation}} by {{Alignment Before Projection}}},
  year =          {2024},
  booktitle = {Proceedings of the 2024 Conference on Empirical Methods in Natural Language Processing}
}

@inproceedings{liLLaMAVIDImageWorth2023,
  author =        {Li, Yanwei and Wang, Chengyao and Jia, Jiaya},
  title =         {{{LLaMA-VID}}: {{An Image}} Is {{Worth}} 2 {{Tokens}}
                   in {{Large Language Models}}},
  year =          {2024},
  doi =           {10.48550/arXiv.2311.17043},
  booktitle = {Proc. ECCV}
}

@misc{ataallahMiniGPT4VideoAdvancingMultimodal2024,
  author =        {Ataallah, Kirolos and Shen, Xiaoqian and
                   Abdelrahman, Eslam and Sleiman, Essam and Zhu, Deyao and
                   Ding, Jian and Elhoseiny, Mohamed},
  title =         {{{MiniGPT4-Video}}: {{Advancing Multimodal LLMs}} for
                   {{Video Understanding}} with {{Interleaved
                   Visual-Textual Tokens}}},
  year =          {2024},
  howpublished = {arXiv preprint arXiv:2404.03413}
}

@inproceedings{ataallahGoldfishVisionLanguageUnderstanding2024,
  author =        {Ataallah, Kirolos and Shen, Xiaoqian and
                   Abdelrahman, Eslam and Sleiman, Essam and
                   Zhuge, Mingchen and Ding, Jian and Zhu, Deyao and
                   Schmidhuber, J{\"u}rgen and Elhoseiny, Mohamed},
  title =         {Goldfish: {{Vision-Language Understanding}} of
                   {{Arbitrarily Long Videos}}},
  year =          {2024},
  doi =           {10.48550/arXiv.2407.12679},
  booktitle = {Proc. ECCV}
}

@misc{wangQwen2VLEnhancingVisionLanguage2024,
  author =        {Wang, Peng and Bai, Shuai and Tan, Sinan and
                   Wang, Shijie and Fan, Zhihao and Bai, Jinze and
                   Chen, Keqin and Liu, Xuejing and Wang, Jialin and
                   Ge, Wenbin and Fan, Yang and Dang, Kai and
                   Du, Mengfei and Ren, Xuancheng and Men, Rui and
                   Liu, Dayiheng and Zhou, Chang and Zhou, Jingren and
                   Lin, Junyang},
  title =         {Qwen2-{{VL}}: {{Enhancing Vision-Language Model}}'s
                   {{Perception}} of the {{World}} at {{Any
                   Resolution}}},
  year =          {2024},
  doi =           {10.48550/arXiv.2409.12191},
  howpublished = {arXiv preprint arXiv:2409.12191}
}

@misc{chenExpandingPerformanceBoundaries2025,
  author =        {Chen, Zhe and Wang, Weiyun and Cao, Yue and
                   Liu, Yangzhou and Gao, Zhangwei and Cui, Erfei and
                   Zhu, Jinguo and Ye, Shenglong and Tian, Hao and
                   Liu, Zhaoyang and Gu, Lixin and Wang, Xuehui and
                   Li, Qingyun and Ren, Yiming and Chen, Zixuan and
                   Luo, Jiapeng and Wang, Jiahao and Jiang, Tan and
                   Wang, Bo and He, Conghui and Shi, Botian and
                   Zhang, Xingcheng and Lv, Han and Wang, Yi and
                   Shao, Wenqi and Chu, Pei and Tu, Zhongying and
                   He, Tong and Wu, Zhiyong and Deng, Huipeng and
                   Ge, Jiaye and Chen, Kai and Zhang, Kaipeng and
                   Wang, Limin and Dou, Min and Lu, Lewei and
                   Zhu, Xizhou and Lu, Tong and Lin, Dahua and Qiao, Yu and
                   Dai, Jifeng and Wang, Wenhai},
  title =         {Expanding {{Performance Boundaries}} of {{Open-Source
                   Multimodal Models}} with {{Model}}, {{Data}}, and
                   {{Test-Time Scaling}}},
  year =          {2025},
  doi =           {10.48550/arXiv.2412.05271},
  howpublished = {arXiv preprint arXiv:2412.05271}
}

@inproceedings{wangInternVideo2ScalingFoundation2024,
  author =        {Wang, Yi and Li, Kunchang and Li, Xinhao and
                   Yu, Jiashuo and He, Yinan and Wang, Chenting and
                   Chen, Guo and Pei, Baoqi and Yan, Ziang and
                   Zheng, Rongkun and Xu, Jilan and Wang, Zun and
                   Shi, Yansong and Jiang, Tianxiang and Li, Songze and
                   Zhang, Hongjie and Huang, Yifei and Qiao, Yu and
                   Wang, Yali and Wang, Limin},
  title =         {{{InternVideo2}}: {{Scaling Foundation Models}} for
                   {{Multimodal Video Understanding}}},
  year =          {2024},
  doi =           {10.48550/arXiv.2403.15377},
  booktitle = {Proc. ECCV}
}

@misc{wangInternVideo25EmpoweringVideo2025,
  author =        {Wang, Yi and Li, Xinhao and Yan, Ziang and He, Yinan and
                   Yu, Jiashuo and Zeng, Xiangyu and Wang, Chenting and
                   Ma, Changlian and Huang, Haian and Gao, Jianfei and
                   Dou, Min and Chen, Kai and Wang, Wenhai and Qiao, Yu and
                   Wang, Yali and Wang, Limin},
  title =         {{{InternVideo2}}.5: {{Empowering Video MLLMs}} with
                   {{Long}} and {{Rich Context Modeling}}},
  year =          {2025},
  doi =           {10.48550/arXiv.2501.12386},
  howpublished = {arXiv preprint arXiv:2501.12386}
}

@inproceedings{songMovieChatDenseToken2024,
  author =        {Song, Enxin and Chai, Wenhao and Wang, Guanhong and
                   Zhang, Yucheng and Zhou, Haoyang and Wu, Feiyang and
                   Chi, Haozhe and Guo, Xun and Ye, Tian and
                   Zhang, Yanting and Lu, Yan and Hwang, Jenq-Neng and
                   Wang, Gaoang},
  title =         {{{MovieChat}}: {{From Dense Token}} to {{Sparse
                   Memory}} for {{Long Video Understanding}}},
  year =          {2024},
  doi =           {10.48550/arXiv.2307.16449},
  booktitle = {Proc. CVPR}
}

@inproceedings{heMALMMMemoryAugmented2024,
  author =        {He, Bo and Li, Hengduo and Jang, Young Kyun and
                   Jia, Menglin and Cao, Xuefei and Shah, Ashish and
                   Shrivastava, Abhinav and Lim, Ser-Nam},
  booktitle =     {Proc. CVPR},
  title =         {{MA-LMM}: Memory-Augmented Large Multimodal Model for
                   Long-Term Video Understanding},
  year =          {2024},
}

@inproceedings{qianStreamingLongVideo2024,
  author =        {Qian, Rui and Dong, Xiaoyi and Zhang, Pan and
                   Zang, Yuhang and Ding, Shuangrui and Lin, Dahua and
                   Wang, Jiaqi},
  journal =       {arXiv.org},
  title =         {Streaming {{Long Video Understanding}} with {{Large
                   Language Models}}},
  year =          {2024},
  booktitle = {Proc. NeurIPS}
}

@inproceedings{chenVideoLLMonlineOnlineVideo2024,
  author =        {Chen, Joya and Lv, Zhaoyang and Wu, Shiwei and
                   Lin, Kevin Qinghong and Song, Chenan and Gao, Difei and
                   Liu, Jia-Wei and Gao, Ziteng and Mao, Dongxing and
                   Shou, Mike Zheng},
  title =         {{{VideoLLM-online}}: {{Online Video Large Language
                   Model}} for {{Streaming Video}}},
  year =          {2024},
  doi =           {10.48550/arXiv.2406.11816},
  booktitle = {Proc. CVPR}
}

@inproceedings{wuVideoLLMMoDEfficientVideoLanguage2024,
  author =        {Wu, Shiwei and Chen, Joya and Lin, Kevin Qinghong and
                   Wang, Qimeng and Gao, Yan and Xu, Qianli and Xu, Tong and
                   Hu, Yao and Chen, Enhong and Shou, Mike Zheng},
  title =         {{{VideoLLM-MoD}}: {{Efficient Video-Language
                   Streaming}} with {{Mixture-of-Depths Vision
                   Computation}}},
  year =          {2024},
  doi =           {10.48550/arXiv.2408.16730},
  booktitle = {Proc. NeurIPS}
}

@inproceedings{santos$infty$VideoTrainingFreeApproach2025,
  author =        {Santos, Saul and Farinhas, Ant{\'o}nio and
                   McNamee, Daniel C. and Martins, Andr{\'e} F. T.},
  journal =       {arXiv.org},
  title =         {\$\textbackslash infty\$-{{Video}}: {{A Training-Free
                   Approach}} to {{Long Video Understanding}} via
                   {{Continuous-Time Memory Consolidation}}},
  year =          {2025},
  booktitle = {Proc. ICML}
}

@inproceedings{zengTimeSuiteImprovingMLLMs2024,
  author =        {Zeng, Xiangyu and Li, Kunchang and Wang, Chenting and
                   Li, Xinhao and Jiang, Tianxiang and Yan, Ziang and
                   Li, Songze and Shi, Yansong and Yue, Zhengrong and
                   Wang, Yi and Wang, Yali and Qiao, Yu and Wang, Limin},
  title =         {{{TimeSuite}}: {{Improving MLLMs}} for {{Long Video
                   Understanding}} via {{Grounded Tuning}}},
  year =          {2025},
  booktitle = {Proc. ICLR},
  pages = {38057--38081}
}

@misc{liVideoChatR1EnhancingSpatioTemporal2025,
  author =        {Li, Xinhao and Yan, Ziang and Meng, Desen and
                   Dong, Lu and Zeng, Xiangyu and He, Yinan and
                   Wang, Yali and Qiao, Yu and Wang, Yi and Wang, Limin},
  title =         {{{VideoChat-R1}}: {{Enhancing Spatio-Temporal
                   Perception}} via {{Reinforcement Fine-Tuning}}},
  year =          {2025},
  howpublished = {arXiv preprint arXiv:2504.06958}
}

@inproceedings{chenVideoChatM1CollaborativePolicy2025,
  author =        {Chen, Boyu and Wang, Zikang and Yue, Zhengrong and
                   Yan, Kainan and Yu, Chenyun and Huang, Yi and
                   Liu, Zijun and Wen, Yafei and Chen, Xiaoxin and
                   Liu, Yang and Li, Peng and Wang, Yali},
  title =         {{{VideoChat-M1}}: {{Collaborative Policy Planning}}
                   for {{Video Understanding}} via {{Multi-Agent
                   Reinforcement Learning}}},
  year =          {2026},
  booktitle = {Proc. CVPR}
}

@misc{chengVideoLLaMA2Advancing2024,
  author =        {Cheng, Zesen and Leng, Sicong and Zhang, Hang and
                   Xin, Yifei and Li, Xin and Chen, Guanzheng and
                   Zhu, Yongxin and Zhang, Wenqi and Luo, Ziyang and
                   Zhao, Deli and Bing, Lidong},
  title =         {{{VideoLLaMA}} 2: {{Advancing Spatial-Temporal
                   Modeling}} and {{Audio Understanding}} in
                   {{Video-LLMs}}},
  year =          {2024},
  howpublished = {arXiv preprint arXiv:2406.07476}
}

@misc{xuQwen25OmniTechnicalReport2025,
  author =        {Xu, Jin and Guo, Zhifang and He, Jinzheng and
                   Hu, Hangrui and He, Ting and Bai, Shuai and
                   Chen, Keqin and Wang, Jialin and Fan, Yang and
                   Dang, Kai and Zhang, Bin and Wang, Xiong and
                   Chu, Yunfei and Lin, Junyang},
  title =         {Qwen2.5-{{Omni Technical Report}}},
  year =          {2025},
  doi =           {10.48550/arXiv.2503.20215},
  howpublished = {arXiv preprint arXiv:2503.20215}
}

@misc{yeOmniVinciEnhancingArchitecture2025,
  author =        {Ye, Hanrong and Yang, Chao-Han Huck and Goel, Arushi and
                   Huang, Wei and Zhu, Ligeng and Su, Yuanhang and
                   Lin, Sean and Cheng, An-Chieh and Wan, Zhen and
                   Tian, Jinchuan and Lou, Yuming and Yang, Dong and
                   Liu, Zhijian and Chen, Yukang and Dantrey, Ambrish and
                   Jahangiri, Ehsan and Ghosh, Sreyan and Xu, Daguang and
                   {Hosseini-Asl}, Ehsan and Taheri, Danial Mohseni and
                   Murali, Vidya and Liu, Sifei and Lu, Yao and
                   Olabiyi, Oluwatobi and Wang, Yu-Chiang Frank and
                   Valle, Rafael and Catanzaro, Bryan and Tao, Andrew and
                   Han, Song and Kautz, Jan and Yin, Hongxu and
                   Molchanov, Pavlo},
  title =         {{{OmniVinci}}: {{Enhancing Architecture}} and
                   {{Data}} for {{Omni-Modal Understanding LLM}}},
  year =          {2025},
  doi =           {10.48550/arXiv.2510.15870},
  howpublished = {arXiv preprint arXiv:2510.15870}
}

@inproceedings{hanLongInsightBenchComprehensiveBenchmark2025,
  author =        {Han, ZhaoYang and Lin, Qihan and Liang, Hao and
                   Chen, Bowen and Liu, Zhou and Zhang, Wentao},
  title =         {{{LongInsightBench}}: {{A Comprehensive Benchmark}}
                   for {{Evaluating Omni-Modal Models}} on
                   {{Human-Centric Long-Video Understanding}}},
  year =          {2026},
  booktitle = {Findings of ACL},
  pages = {19332--19358}
}

@inproceedings{daoFlashAttentionFastMemoryEfficient2022,
  author =        {Dao, Tri and Fu, Dan and Ermon, Stefano and
                   Rudra, Atri and Ré, Christopher},
  booktitle = {Proc. NeurIPS},
  pages =         {16344–16359},
  publisher =     {Neural Information Processing Systems Foundation,
                   Inc. (NeurIPS)},
  series =        {NeurIPS 2022},
  title =         {FlashAttention: Fast and Memory-Efficient Exact
                   Attention with IO-Awareness},
  year =          {2022},
  doi =           {10.52202/068431-1189},
}

@inproceedings{wangTemporalSegmentNetworks2016,
  author =        {Wang, Limin and Xiong, Yuanjun and Wang, Zhe and
                   Qiao, Yu and Lin, Dahua and Tang, Xiaoou and
                   Gool, Luc Van},
  title =         {Temporal {{Segment Networks}}: {{Towards Good
                   Practices}} for {{Deep Action Recognition}}},
  year =          {2016},
  doi =           {10.48550/arXiv.1608.00859},
  booktitle = {Proc. ECCV}
}

@inproceedings{potapovCategorySpecificVideoSummarization2014,
  address =       {Cham},
  author =        {Potapov, Danila and Douze, Matthijs and
                   Harchaoui, Zaid and Schmid, Cordelia},
  booktitle =     {Proc. ECCV},
  editor =        {Fleet, David and Pajdla, Tomas and Schiele, Bernt and
                   Tuytelaars, Tinne},
  pages =         {540--555},
  publisher =     {Springer International Publishing},
  title =         {Category-{{Specific Video Summarization}}},
  year =          {2014},
  doi =           {10.1007/978-3-319-10599-4_35},
  isbn =          {978-3-319-10599-4},
}

@inproceedings{afhamRevisitingKernelTemporal2023,
  author =        {Afham, Mohamed and Shukla, Satya Narayan and
                   Poursaeed, Omid and Zhang, Pengchuan and Shah, Ashish and
                   Lim, Sernam},
  journal =       {arXiv.org},
  title =         {Revisiting {{Kernel Temporal Segmentation}} as an
                   {{Adaptive Tokenizer}} for {{Long-form Video
                   Understanding}}},
  year =          {2023},
  booktitle = {Proc. ICCV}
}

@misc{sunFramesClipsTrainingfree2025,
  author =        {Sun, Guangyu and Singhal, Archit and Uzkent, Burak and
                   Shah, Mubarak and Chen, Chen and Kessler, Garin},
  title =         {From {{Frames}} to {{Clips}}: {{Training-free
                   Adaptive Key Clip Selection}} for {{Long-Form Video
                   Understanding}}},
  year =          {2025},
  doi =           {10.48550/arXiv.2510.02262},
  howpublished = {arXiv preprint arXiv:2510.02262}
}

@inproceedings{liMaxInfoTrainingFreeKeyFrame2025,
  author =        {Li, Pengyi and Abdullaeva, Irina and
                   Gambashidze, Alexander and Kuznetsov, Andrey and
                   Oseledets, Ivan},
  title =         {{{MaxInfo}}: {{A Training-Free Key-Frame Selection
                   Method Using Maximum Volume}} for {{Enhanced Video
                   Understanding}}},
  year =          {2026},
  doi =           {10.48550/arXiv.2502.03183},
  booktitle = {Proc. WACV}
}

@inproceedings{wuAdaFrameAdaptiveFrame2018,
  author =        {Wu, Zuxuan and Xiong, Caiming and Ma, Chih-Yao and
                   Socher, Richard and Davis, Larry S.},
  title =         {{{AdaFrame}}: {{Adaptive Frame Selection}} for {{Fast
                   Video Recognition}}},
  year =          {2019},
  booktitle = {Proc. CVPR}
}

@inproceedings{zhiMGSamplerExplainableSampling2021,
  author =        {Zhi, Yuan and Tong, Zhan and Wang, Limin and
                   Wu, Gangshan},
  title =         {{{MGSampler}}: {{An Explainable Sampling Strategy}}
                   for {{Video Action Recognition}}},
  year =          {2021},
  doi =           {10.48550/arXiv.2104.09952},
  booktitle = {Proc. ICCV}
}

@inproceedings{steunou2026peek,
  author =        {Steunou, Killian and Filali Razzouki, Anas and
                   Guetari, Khalil and El-Yacoubi, Moun{\^i}m A. and
                   Tevissen, Yannis},
  booktitle={British Machine Vision Conference (BMVC)},
  title =         {PEEK: Picking Essential frames via Efficient
                   Knowledge distillation},
  year =          {2026},
}

@inproceedings{tangAdaptiveKeyframeSampling2025,
  author =        {Tang, Xi and Qiu, Jihao and Xie, Lingxi and
                   Tian, Yunjie and Jiao, Jianbin and Ye, Qixiang},
  title =         {Adaptive {{Keyframe Sampling}} for {{Long Video
                   Understanding}}},
  year =          {2025},
  doi =           {10.48550/arXiv.2502.21271},
  booktitle = {Proc. CVPR}
}

@inproceedings{zhangQFrameQueryawareFrame2025,
  author =        {Zhang, Shaojie and Yang, Jiahui and Yin, Jianqin and
                   Luo, Zhenbo and Luan, Jian},
  title =         {Q-{{Frame}}: {{Query-aware Frame Selection}} and
                   {{Multi-Resolution Adaptation}} for {{Video-LLMs}}},
  year =          {2025},
  doi =           {10.48550/arXiv.2506.22139},
  booktitle = {Proc. ICCV}
}

@misc{zhuFOCUSEfficientKeyframe2025,
  author =        {Zhu, Zirui and Xu, Hailun and Luo, Yang and Liu, Yong and
                   Sarkar, Kanchan and Yang, Zhenheng and You, Yang},
  title =         {{{FOCUS}}: {{Efficient Keyframe Selection}} for
                   {{Long Video Understanding}}},
  year =          {2025},
  doi =           {10.48550/arXiv.2510.27280},
  howpublished = {arXiv preprint arXiv:2510.27280}
}

@misc{zhangAdaRDkeyAdaptiveRelevanceDiversity2025,
  author =        {Zhang, Xian and Wu, Zexi and Li, Zinuo and
                   Xu, Hongming and Gong, Luqi and Boussaid, Farid and
                   Werghi, Naoufel and Bennamoun, Mohammed},
  title =         {{{AdaRD-key}}: {{Adaptive Relevance-Diversity
                   Keyframe Sampling}} for {{Long-form Video}}
                   Understanding},
  year =          {2025},
  doi =           {10.48550/arXiv.2510.02778},
  howpublished = {arXiv preprint arXiv:2510.02778}
}

@article{li2025dytok,
  author =        {Yulin Li and Haokun Gui and Ziyang Fan and
                   Junjie Wang and Bin Kang and Bin Chen and
                   Zhuotao Tian},
  title =         {Less Is More, but Where? Dynamic Token Compression
                   via {LLM}-Guided Keyframe Prior},
  year =          {2025},
  journal =     {Proc. NeurIPS},
}

@article{chen2026lddr,
  author =        {Jingfeng Chen and Jiawen Qian and Wendi Deng and
                   Yinuo Guo and Jiaqi Yu and Sicong Leng and
                   Raghuveer Thirukovalluru and Bhuwan Dhingra},
  title =         {{LDDR}: Linear-{DPP}-Based Dynamic-Resolution Frame
                   Sampling for Video {MLLMs}},
  year =          {2026},
  journal =      {arXiv preprint arXiv:2605.11477},
}

@inproceedings{liu2025bolt,
  author =        {Liu, Shuming and Zhao, Chen and Xu, Tianqi and
                   Ghanem, Bernard},
  booktitle =     {Proc. CVPR},
  title =         {{BOLT}: Boost Large Vision-Language Model Without
                   Training for Long-form Video Understanding},
  year =          {2025},
}

@inproceedings{ye2025tstar,
  author =        {Ye, Jinhui and Wang, Zihan and Sun, Haosen and
                   Chandrasegaran, Keshigeyan and Durante, Zane and
                   Eyzaguirre, Cristobal and Bisk, Yonatan and
                   Niebles, Juan Carlos and Adeli, Ehsan and Fei-Fei, Li and
                   Wu, Jiajun and Li, Manling},
  booktitle =     {Proc. CVPR},
  title =         {{T*}: Re-thinking Temporal Search for Long-Form Video
                   Understanding},
  year =          {2025},
}

@inproceedings{peng2026qca,
  author =        {Peng, Jun and Song, Baiyang and Li, Jie and Li, Hui and
                   Zhou, Yiyi and Ji, Rongrong and Tian, Yonghong},
  booktitle = {Proc. ECCV},
  title =         {{QCA}: Query- and Content-Aware Keyframe Selection
                   for Long Video Understanding},
  year =          {2026},
}

@article{chen2026efs,
  author =        {Chen, Wang and Luo, Yongdong and Zeng, Yuhui and
                   Lin, Luojun and Xie, Tianyu and Chao, Fei and
                   Ji, Rongrong and Zheng, Xiawu},
  journal =       {arXiv preprint arXiv:2603.00983},
  title =         {Event-Anchored Frame Selection for Effective
                   Long-Video Understanding},
  year =          {2026},
}

@inproceedings{ma2026gift,
  author =        {Ma, Junpeng and Zhou, Sashuai and Li, Guanghao and
                   Gao, Xin and Cao, Yue and Zeng, Hengyu and
                   Yan, Yuxiang and Wang, Zhibin and Song, Jun and
                   Zheng, Bo and Zhang, Shanghang and Pu, Jian},
  booktitle =     {Proc. CVPR},
  title =         {{GIFT}: Global Irreplaceability Frame Targeting for
                   Efficient Video Understanding},
  year =          {2026},
}

@misc{yuFrameVoyagerLearningQuery2025,
  author =        {Yu, Sicheng and Jin, Chengkai and Wang, Huanyu and
                   Chen, Zhenghao and Jin, Sheng and Zuo, Zhongrong and
                   Xu, Xiaolei and Sun, Zhenbang and Zhang, Bingni and
                   Wu, Jiawei and Zhang, Hao and Sun, Qianru},
  title =         {Frame-{{Voyager}}: {{Learning}} to {{Query Frames}}
                   for {{Video Large Language Models}}},
  year =          {2025},
  doi =           {10.48550/arXiv.2410.03226},
  howpublished = {arXiv preprint arXiv:2410.03226}
}

@inproceedings{yaoGenerativeFrameSampler2025,
  author =        {Yao, Linli and Wu, Haoning and Ouyang, Kun and
                   Zhang, Yuanxing and Xiong, Caiming and Chen, Bei and
                   Sun, Xu and Li, Junnan},
  title =         {Generative {{Frame Sampler}} for {{Long Video
                   Understanding}}},
  year =          {2025},
  doi =           {10.48550/arXiv.2503.09146},
  booktitle = {Findings of ACL}
}

@misc{yangHFSHolisticQueryAware2025,
  author =        {Yang, Yiqing and Lam, Kin-Man},
  title =         {{{HFS}}: {{Holistic Query-Aware Frame Selection}} for
                   {{Efficient Video Reasoning}}},
  year =          {2025},
  doi =           {10.48550/arXiv.2512.11534},
  howpublished = {arXiv preprint arXiv:2512.11534}
}

@inproceedings{qin2026efficient,
  author =        {Yaxuan Qin and Hefei Li and Wenqi Mu and Yancheng He},
  booktitle =     {Proc. CVPR},
  pages =         {16944--16953},
  title =         {Efficient Frame Selection for Long Video
                   Understanding via Reinforcement Learning},
  year =          {2026},
}

@inproceedings{tang2025tspo,
  author =        {Tang, Canhui and Han, Zifan and Sun, Hongbo and
                   Zhou, Sanping and Zhang, Xuchong and Wei, Xin and
                   Yuan, Ye and Zhang, Huayu and Xu, Jinglin and
                   Sun, Hao},
  booktitle =     {Proc. AAAI},
  title =         {{TSPO}: Temporal Sampling Policy Optimization for
                   Long-form Video Language Understanding},
  year =          {2026},
}

@inproceedings{lee2025refocus,
  author =        {Lee, Hosu and Kim, Junho and Kim, Hyunjun and
                   Ro, Yong Man},
  booktitle =     {Proc. CVPR},
  title =         {{ReFoCUS}: Reinforcement-guided Frame Optimization
                   for Contextual Understanding},
  year =          {2026},
}

@article{xu2025viarl,
  author =        {Xu, Ziqiang and Dai, Qi and Xie, Tian and Yang, Yifan and
                   Qiu, Kai and Chen, DongDong and Wu, Zuxuan and
                   Luo, Chong},
  journal =       {arXiv preprint arXiv:2505.15447},
  title =         {{ViaRL}: Adaptive Temporal Grounding via Visual
                   Iterated Amplification Reinforcement Learning},
  year =          {2025},
}

@article{wang2025videoitg,
  author =        {Wang, Shihao and Chen, Guo and Huang, De-an and
                   Li, Zhiqi and Li, Minghan and Liu, Guilin and
                   Alvarez, Jose M. and Zhang, Lei and Yu, Zhiding},
  journal =       {arXiv preprint arXiv:2507.13353},
  title =         {{VideoITG}: Multimodal Video Understanding with
                   Instructed Temporal Grounding},
  year =          {2025},
}

@inproceedings{alayracSelfSupervisedMultiModalVersatile2020,
  author =        {Alayrac, Jean-Baptiste and Recasens, Adri{\`a} and
                   Schneider, Rosalia and Arandjelovi{\'c}, Relja and
                   Ramapuram, Jason and Fauw, Jeffrey De and
                   Smaira, Lucas and Dieleman, Sander and
                   Zisserman, Andrew},
  title =         {Self-{{Supervised MultiModal Versatile Networks}}},
  year =          {2020},
  booktitle = {Proc. NeurIPS}
}

@inproceedings{akbariVATTTransformersMultimodal2021,
  author =        {Akbari, Hassan and Yuan, Liangzhe and Qian, Rui and
                   Chuang, Wei-Hong and Chang, Shih-Fu and Cui, Yin and
                   Gong, Boqing},
  title =         {{{VATT}}: {{Transformers}} for {{Multimodal
                   Self-Supervised Learning}} from {{Raw Video}},
                   {{Audio}} and {{Text}}},
  year =          {2021},
  booktitle = {Proc. NeurIPS}
}

@inproceedings{linTSMTemporalShift2019,
  author =        {Lin, Ji and Gan, Chuang and Han, Song},
  title =         {{{TSM}}: {{Temporal Shift Module}} for {{Efficient
                   Video Understanding}}},
  year =          {2019},
  doi =           {10.48550/arXiv.1811.08383},
  booktitle = {Proc. ICCV}
}

@inproceedings{feichtenhoferX3DExpandingArchitectures2020,
  author =        {Feichtenhofer, Christoph},
  booktitle =     {Proc. CVPR},
  pages =         {203--213},
  title =         {{X3D}: Expanding Architectures for Efficient Video
                   Recognition},
  year =          {2020},
  doi =           {10.1109/CVPR42600.2020.00028},
}

@inproceedings{kondratyukMoViNetsMobileVideo2021,
  author =        {Kondratyuk, Dan and Yuan, Liangzhe and Li, Yandong and
                   Zhang, Li and Tan, Mingxing and Brown, Matthew and
                   Gong, Boqing},
  booktitle =     {Proc. CVPR},
  pages =         {16020--16030},
  title =         {{MoViNets}: Mobile Video Networks for Efficient Video
                   Recognition},
  year =          {2021},
  doi =           {10.1109/CVPR46437.2021.01576},
}

@inproceedings{fanMultiscaleVisionTransformers2021,
  author =        {Fan, Haoqi and Xiong, Bo and Mangalam, Karttikeya and
                   Li, Yanghao and Yan, Zhicheng and Malik, Jitendra and
                   Feichtenhofer, Christoph},
  title =         {Multiscale {{Vision Transformers}}},
  year =          {2021},
  doi =           {10.48550/arXiv.2104.11227},
  booktitle = {Proc. ICCV}
}

@inproceedings{liMViTv2ImprovedMultiscale2022,
  author =        {Li, Yanghao and Wu, Chao-Yuan and Fan, Haoqi and
                   Mangalam, Karttikeya and Xiong, Bo and
                   Malik, Jitendra and Feichtenhofer, Christoph},
  title =         {{{MViTv2}}: {{Improved Multiscale Vision
                   Transformers}} for {{Classification}} and
                   {{Detection}}},
  year =          {2022},
  doi =           {10.48550/arXiv.2112.01526},
  booktitle = {Proc. CVPR}
}

@inproceedings{ryaliHieraHierarchicalVision2023,
  author =        {Ryali, Chaitanya and Hu, Yuan-Ting and Bolya, Daniel and
                   Wei, Chen and Fan, Haoqi and Huang, Po-Yao and
                   Aggarwal, Vaibhav and Chowdhury, Arkabandhu and
                   Poursaeed, Omid and Hoffman, Judy and Malik, Jitendra and
                   Li, Yanghao and Feichtenhofer, Christoph},
  booktitle =     {Proc. ICML},
  title =         {Hiera: {A} Hierarchical Vision Transformer without
                   the Bells-and-Whistles},
  year =          {2023},
}

@inproceedings{liuVideoSwinTransformer2021,
  author =        {Liu, Ze and Ning, Jia and Cao, Yue and Wei, Yixuan and
                   Zhang, Zheng and Lin, Stephen and Hu, Han},
  title =         {Video {{Swin Transformer}}},
  year =          {2022},
  booktitle = {Proc. CVPR}
}

@inproceedings{liUniFormerUnifiedTransformer2022,
  author =        {Li, Kunchang and Wang, Yali and Gao, Peng and
                   Song, Guanglu and Liu, Yu and Li, Hongsheng and
                   Qiao, Yu},
  title =         {{{UniFormer}}: {{Unified Transformer}} for
                   {{Efficient Spatiotemporal Representation Learning}}},
  year =          {2022},
  booktitle = {Proc. ICLR}
}

@inproceedings{liUniFormerV2Spatiotemporal2023,
  author =        {Li, Kunchang and Wang, Yali and He, Yinan and
                   Li, Yizhuo and Wang, Yi and Wang, Limin and Qiao, Yu},
  booktitle =     {Proc. ICCV},
  title =         {{UniFormerV2}: Spatiotemporal Learning by Arming
                   Image {ViTs} with Video {UniFormer}},
  year =          {2023},
}

@inproceedings{liVideoMambaState2024,
  author =        {Li, Kunchang and Li, Xinhao and Wang, Yi and
                   He, Yinan and Wang, Yali and Wang, Limin and
                   Qiao, Yu},
  booktitle = {Proc. ECCV},
  pages =         {237--255},
  publisher =     {Springer},
  title =         {{VideoMamba}: State Space Model for Efficient Video
                   Understanding},
  year =          {2024},
  doi =           {10.1007/978-3-031-73347-5_14},
}

@inproceedings{parkVideoMambaSpatio2024,
  author =        {Park, Jinyoung and Kim, Hee-Seon and Ko, Kangwook and
                   Kim, Minbeom and Kim, Changick},
  booktitle = {Proc. ECCV},
  title =         {{VideoMamba}: Spatio-Temporal Selective State Space
                   Model},
  year =          {2024},
  doi =           {10.1007/978-3-031-72698-9_1},
}

@inproceedings{luSnakesLadders2025,
  author =        {Lu, Hui and Salah, Albert Ali and Poppe, Ronald},
  booktitle =     {Proc. ICCV},
  title =         {Snakes and Ladders: Two Steps Up for {VideoMamba}},
  year =          {2025},
}

@inproceedings{bolyaTokenMergingYour2023,
  author =        {Bolya, Daniel and Fu, Cheng-Yang and Dai, Xiaoliang and
                   Zhang, Peizhao and Feichtenhofer, Christoph and
                   Hoffman, Judy},
  booktitle =     {Proc. ICLR},
  title =         {Token Merging: Your {ViT} but Faster},
  year =          {2023},
}

@inproceedings{soldanResidualViT2025,
  author =        {Soldan, Mattia and Caba Heilbron, Fabian and Ghanem, Bernard and Sivic, Josef and Russell, Bryan},
  booktitle =     {Proc. ICCV},
  title =         {{ResidualViT} for Efficient Temporally Dense Video Encoding},
  year =          {2025},
}

@article{wang2026earlytom,
  author =        {Hesong Wang and Xin Jin and Lu Lu and Chenhaowen Li and
                   Jian Chen and Qiang Liu and Huan Wang},
  title =         {{EarlyTom}: Early Token Compression Completes Fast
                   Video Understanding},
  year =          {2026},
  journal =       {Proc. CVPR},
}

@article{wang2025stc,
  author =        {Yiyu Wang and Xuyang Liu and Xiyan Gui and
                   Xinying Lin and Boxue Yang and Chenfei Liao and
                   Tailai Chen and Linfeng Zhang},
  title =         {Accelerating Streaming Video Large Language Models
                   via Hierarchical Token Compression},
  year =          {2026},
  journal =       {Proc. CVPR},
}

@inproceedings{shi2026autogaze,
  author =        {Shi, Baifeng and Fu, Stephanie and Lian, Long and Ye, Hanrong and Eigen, David and Reite, Aaron and Li, Boyi and Kautz, Jan and Han, Song and Chan, David M. and Molchanov, Pavlo and Darrell, Trevor and Yin, Hongxu},
  booktitle =     {Proc. CVPR},
  title =         {Attend Before Attention: Efficient and Scalable Video
                   Understanding via Autoregressive Gazing},
  year =          {2026},
}

@article{sarkar2026cope,
  author =        {Sayan Deb Sarkar and R{\'e}mi Pautrat and
                   Ondrej Miksik and Marc Pollefeys and Iro Armeni and
                   Mahdi Rad and Mihai Dusmanu},
  title =         {{CoPE-VideoLM}: Leveraging Codec Primitives For
                   Efficient Video Language Modeling},
  year =          {2026},
  journal =       {arXiv preprint arXiv:2602.13191},
}

@inproceedings{wuTinyCLIPCLIP2023,
  author =        {Wu, Kan and Peng, Houwen and Zhou, Zhenghong and
                   Xiao, Bin and Liu, Mengchen and Yuan, Lu and
                   Xuan, Hong and Valenzuela, Michael and Chen, Xi and
                   Wang, Xinggang and Chao, Hongyang and Hu, Han},
  booktitle =     {Proc. ICCV},
  title =         {{TinyCLIP}: {CLIP} Distillation via Affinity
                   Mimicking and Weight Inheritance},
  year =          {2023},
}

@inproceedings{vasuMobileCLIPFast2023,
  author =        {Vasu, Pavan Kumar Anasosalu and Pouransari, Hadi and
                   Faghri, Fartash and Vemulapalli, Raviteja and
                   Tuzel, Oncel},
  booktitle =     {Proc. CVPR},
  title =         {{MobileCLIP}: Fast Image-Text Models through
                   Multi-Modal Reinforced Training},
  year =          {2024},
}

@article{faghri2025mobileclip2,
  author =        {Faghri, Fartash and Vasu, Pavan Kumar Anasosalu and
                   Koc, Cem and Shankar, Vaishaal and Toshev, Alexander and
                   Tuzel, Oncel and Pouransari, Hadi},
  journal =       {Transactions on Machine Learning Research},
  title =         {{MobileCLIP2}: Improving Multi-Modal Reinforced
                   Training},
  year =          {2025},
}

@inproceedings{yangMobileViCLIPEfficientVideoText2025,
  author =        {Yang, Min and Jia, Zihan and Dai, Zhilin and
                   Guo, Sheng and Wang, Limin},
  title =         {{{MobileViCLIP}}: {{An Efficient Video-Text Model}}
                   for {{Mobile Devices}}},
  year =          {2025},
  doi =           {10.48550/arXiv.2508.07312},
  booktitle = {Proc. ICCV}
}

@inproceedings{vasuFastVLMEfficient2024,
  author =        {Vasu, Pavan Kumar Anasosalu and Faghri, Fartash and
                   Li, Chun-Liang and Koc, Cem and True, Nate and
                   Antony, Albert and Santhanam, Gokul and
                   Gabriel, James and Grasch, Peter and Tuzel, Oncel and
                   Pouransari, Hadi},
  booktitle =     {Proc. CVPR},
  title =         {{FastVLM}: Efficient Vision Encoding for Vision
                   Language Models},
  year =          {2025},
}

@misc{kimLiteFrameEfficient2026,
  author =        {Kim, Jihwan and Parthasarathy, Nikhil and
                   Qin, Danfeng and Hur, Junhwa and Sun, Deqing and
                   Han, Bohyung and Yang, Ming-Hsuan and Gong, Boqing},
  title =         {{LiteFrame}: Efficient Vision Encoders Unlock Frame
                   Scaling in Video {LLMs}},
  year =          {2026},
  howpublished = {arXiv preprint arXiv:2605.17260}
}

@inproceedings{zhang2026moevie,
  author =        {Zhang, Bonan and Dong, Shiyu and Tran, Quan Hung and
                   Gschwind, Katharina and Yang, Shuqi and Chen, Sijia and
                   Ahmadyan, Adel and Moon, Seungwhan and Zhang, Lu and
                   Kirmani, Ahmed and Damavandi, Babak and Kumar, Anuj},
  booktitle = {Proc. ECCV},
  title =         {{MoE-ViE}: Mixture of Experts Vision Encoder for
                   Efficient Image and Video Understanding},
  year =          {2026},
}

@inproceedings{yang2024visionzip,
  author =        {Senqiao Yang and Yukang Chen and Zhuotao Tian and
                   Chengyao Wang and Jingyao Li and Bei Yu and
                   Jiaya Jia},
  booktitle =     {Proc. CVPR},
  pages =         {19792--19802},
  title =         {{VisionZip}: Longer is Better but Not Necessary in
                   Vision Language Models},
  year =          {2025},
  doi =           {10.1109/CVPR52734.2025.01843},
}

@inproceedings{shang2024prumerge,
  author =        {Yuzhang Shang and Mu Cai and Bingxin Xu and
                   Yong Jae Lee and Yan Yan},
  booktitle =     {Proc. ICCV},
  pages =         {22857--22867},
  title =         {{LLaVA-PruMerge}: Adaptive Token Reduction for
                   Efficient Large Multimodal Models},
  year =          {2025},
}

@inproceedings{jinChatUniViUnifiedVisual2024,
  author =        {Jin, Peng and Takanobu, Ryuichi and Zhang, Wancai and
                   Cao, Xiaochun and Yuan, Li},
  title =         {Chat-{{UniVi}}: {{Unified Visual Representation
                   Empowers Large Language Models}} with {{Image}} and
                   {{Video Understanding}}},
  year =          {2024},
  doi =           {10.48550/arXiv.2311.08046},
  booktitle = {Proc. CVPR}
}

@inproceedings{shao2025holitom,
  author =        {Shao, Kele and Tao, Keda and Qin, Can and
                   You, Haoxuan and Sui, Yang and Wang, Huan},
  booktitle =     {Proc. NeurIPS},
  title =         {{HoliTom}: Holistic Token Merging for Fast Video
                   Large Language Models},
  year =          {2025},
}

@inproceedings{shenLongVUSpatiotemporalAdaptive2024,
  author =        {Shen, Xiaoqian and Xiong, Yunyang and
                   Zhao, Changsheng and Wu, Lemeng and Chen, Jun and
                   Zhu, Chenchen and Liu, Zechun and Xiao, Fanyi and
                   Varadarajan, Balakrishnan and Bordes, Florian and
                   Liu, Zhuang and Xu, Hu and Kim, Hyunwoo J. and
                   Soran, Bilge and Krishnamoorthi, Raghuraman and
                   Elhoseiny, Mohamed and Chandra, Vikas},
  title =         {{LongVU}: Spatiotemporal Adaptive Compression for
                   Long Video-Language Understanding},
  year =          {2025},
  booktitle =     {Proc. ICML},
}

@article{fan2026flashvid,
  author =        {Ziyang Fan and Keyu Chen and Ruilong Xing and
                   Yulin Li and Li Jiang and Zhuotao Tian},
  title =         {{FlashVID}: Efficient Video Large Language Models via
                   Training-free Tree-based Spatiotemporal Token
                   Merging},
  year =          {2026},
  journal = {Proc. ICLR},
}

@inproceedings{shen2025fastvid,
  author =        {Shen, Leqi and Gong, Guoqiang and He, Tao and
                   Zhang, Yifeng and Liu, Pengzhang and Zhao, Sicheng and
                   Ding, Guiguang},
  booktitle =     {Proc. NeurIPS},
  title =         {{FastVID}: Dynamic Density Pruning for Fast Video
                   Large Language Models},
  year =          {2025},
}

@inproceedings{hyun2025sttm,
  author =        {Hyun, Jeongseok and Hwang, Sukjun and Han, Su Ho and
                   Kim, Taeoh and Lee, Inwoong and Wee, Dongyoon and
                   Lee, Joon-Young and Kim, Seon Joo and Shim, Minho},
  booktitle =     {Proc. ICCV},
  title =         {Multi-Granular Spatio-Temporal Token Merging for
                   Training-Free Acceleration of Video {LLMs}},
  year =          {2025},
}

@misc{sun2025llavascissor,
  author =        {Sun, Boyuan and Zhao, Jiaxing and Wei, Xihan and
                   Hou, Qibin},
  title =         {{LLaVA-Scissor}: Token Compression with Semantic
                   Connected Components for Video {LLMs}},
  year =          {2025},
  howpublished = {arXiv preprint arXiv:2506.21862}
}

@inproceedings{liu2025vidcom2,
  author =        {Liu, Xuyang and Wang, Yiyu and Ma, Junpeng and
                   Zhang, Linfeng},
  booktitle =     {Proc. EMNLP},
  title =         {Video Compression Commander: Plug-and-Play Inference
                   Acceleration for Video Large Language Models},
  year =          {2025},
}

@inproceedings{ma2025mmgvid,
  author =        {Ma, Junpeng and Zhang, Qizhe and Lu, Ming and
                   Wang, Zhibin and Zhou, Qiang and Song, Jun and
                   Zhang, Shanghang},
  booktitle =     {Proc. AAAI},
  title =         {{MMG-Vid}: Maximizing Marginal Gains at Segment-level
                   and Token-level for Efficient Video {LLMs}},
  year =          {2026},
}

@misc{kang2026ottvid,
  author =        {Kang, Minseok and Lee, Minhyeok and Lee, Jungho and
                   Kim, Minjung and Kim, Donghyeong and Lee, Dayeon and
                   Choi, Heeseung and Kim, Ig-jae and Lee, Sangyoun},
  title =         {{OTT-Vid}: Optimal Transport Temporal Token
                   Compression for Video Large Language Models},
  year =          {2026},
  howpublished = {arXiv preprint arXiv:2605.11803}
}

@misc{park2026dynatok,
  author =        {Park, Minyoung and Kong, Taehun and Ahn, Sangjun},
  title =         {{DynaTok}: Temporally Adaptive and Positional
                   Bias-Aware Token Compression for {Video-LLMs}},
  year =          {2026},
  howpublished = {arXiv preprint arXiv:2605.19322}
}

@misc{liu2026infomerge,
  author =        {Liu, Xinxin and Gan, Shiwei and Liu, Xiao and
                   Yin, Yafeng and Xie, Lei and Lu, Sanglu},
  title =         {{InfoMerge}: Information-aware Token Compression for
                   Efficient Video Large Language Models},
  year =          {2026},
  howpublished = {arXiv preprint arXiv:2606.02161}
}

@inproceedings{ju2026forestprune,
  author =        {Ju, Shaobo and Song, Baiyang and Chen, Tao and
                   Zhang, Jiapeng and Wu, Qiong and Chang, Chao and
                   Wang, HuaiXi and Zhou, Yiyi and Ji, Rongrong},
  booktitle =     {Proc. CVPR},
  title =         {{ForestPrune}: High-ratio Visual Token Compression
                   for Video Multimodal Large Language Models via
                   Spatial-Temporal Forest Modeling},
  year =          {2026},
}

@inproceedings{wu2026metom,
  author =        {Wu, Zhuojie and Wang, Shijie and Yu, Xin},
  booktitle =     {Proc. CVPR},
  pages =         {10441--10448},
  title =         {{MeToM}: Metadata-Guided Token Merging for Efficient
                   Video {LLMs}},
  year =          {2026},
}

@inproceedings{song2026ktv,
  author =        {Song, Baiyang and Peng, Jun and Zhang, Yuxin and
                   Chen, Guangyao and Yang, Feidiao and Guo, Jianyuan},
  booktitle =     {Proc. AAAI},
  title =         {{KTV}: Keyframes and Key Tokens Selection for
                   Efficient Training-Free Video {LLMs}},
  year =          {2026},
}

@inproceedings{alayracFlamingoVisualLanguage2022,
  author =        {Alayrac, Jean-Baptiste and Donahue, Jeff and
                   Luc, Pauline and Miech, Antoine and Barr, Iain and
                   Hasson, Yana and Lenc, Karel and Mensch, Arthur and
                   Millican, Katie and Reynolds, Malcolm and Ring, Roman and
                   Rutherford, Eliza and Cabi, Serkan and Han, Tengda and
                   Gong, Zhitao and Samangooei, Sina and
                   Monteiro, Marianne and Menick, Jacob and
                   Borgeaud, Sebastian and Brock, Andrew and
                   Nematzadeh, Aida and Sharifzadeh, Sahand and
                   Binkowski, Mikolaj and Barreira, Ricardo and
                   Vinyals, Oriol and Zisserman, Andrew and
                   Simonyan, Karen},
  title =         {Flamingo: A {{Visual Language Model}} for {{Few-Shot
                   Learning}}},
  year =          {2022},
  booktitle = {Proc. NeurIPS}
}

@inproceedings{yeVoCoLLaMAVisionCompression2024,
  author =        {Ye, Xubing and Gan, Yukang and Huang, Xiaoke and
                   Ge, Yixiao and Shan, Ying and Tang, Yansong},
  booktitle =     {Proc. CVPR},
  pages =         {29836--29846},
  title =         {{VoCo-LLaMA}: Towards Vision Compression with Large
                   Language Models},
  year =          {2025},
}

@inproceedings{zhangLLaVAMiniEfficientImage2025,
  author =        {Zhang, Shaolei and Fang, Qingkai and Yang, Zhe and
                   Feng, Yang},
  booktitle =     {Proc. ICLR},
  title =         {{LLaVA-Mini}: Efficient Image and Video Large
                   Multimodal Models with One Vision Token},
  year =          {2025},
}

@misc{ryoo2024blip3video,
  author =        {Ryoo, Michael S. and Zhou, Honglu and
                   Kendre, Shrikant and Qin, Can and Xue, Le and
                   Shu, Manli and Park, Jongwoo and Ranasinghe, Kanchana and
                   Savarese, Silvio and Xu, Ran and Xiong, Caiming and
                   Niebles, Juan Carlos},
  title =         {{xGen-MM-Vid} ({BLIP-3-Video}): You Only Need 32
                   Tokens to Represent a Video Even in {VLMs}},
  year =          {2024},
  howpublished = {arXiv preprint arXiv:2410.16267}
}

@misc{qi2025quicksviewer,
  author =        {Qi, Ji and Yao, Yuan and Bai, Yushi and Xu, Bin and
                   Li, Juanzi and Liu, Zhiyuan and Chua, Tat-Seng},
  title =         {An {LMM} for Efficient Video Understanding via
                   Reinforced Compression of Video Cubes},
  year =          {2025},
  howpublished = {arXiv preprint arXiv:2504.15270}
}

@misc{lan2024vidcompress,
  author =        {Lan, Xiaohan and Yuan, Yitian and Jie, Zequn and
                   Ma, Lin},
  title =         {{VidCompress}: Memory-Enhanced Temporal Compression
                   for Video Understanding in Large Language Models},
  year =          {2024},
  howpublished = {arXiv preprint arXiv:2410.11417}
}

@misc{xuPLLaVAParameterfreeLLaVA2024,
  author =        {Xu, Lin and Zhao, Yilin and Zhou, Daquan and
                   Lin, Zhijie and Ng, See Kiong and Feng, Jiashi},
  title =         {{{PLLaVA}} : {{Parameter-free LLaVA Extension}} from
                   {{Images}} to {{Videos}} for {{Video Dense
                   Captioning}}},
  year =          {2024},
  howpublished = {arXiv preprint arXiv:2404.16994}
}

@misc{xu2024slowfastllava,
  author =        {Mingze Xu and Mingfei Gao and Zhe Gan and
                   Hong-You Chen and Zhengfeng Lai and Haiming Gang and
                   Kai Kang and Afshin Dehghan},
  title =         {{SlowFast-LLaVA}: A Strong Training-Free Baseline for
                   Video Large Language Models},
  year =          {2024},
  howpublished = {arXiv preprint arXiv:2407.15841}
}

@inproceedings{jiangSTORMTokenEfficientLong2025,
  title =         {{STORM}: Token-Efficient Long Video Understanding for
                   Multimodal {LLMs}},
  DOI = {10.1109/iccvw69036.2025.00614},
  booktitle = {Proc. ICCV},
  publisher = {IEEE},
  author = {Jiang,  Jindong and Li,  Xiuyu and Liu,  Zhijian and Li,  Muyang and Chen,  Guo and Li,  Zhiqi and Huang,  De-An and Liu,  Guilin and Yu,  Zhiding and Keutzer,  Kurt and Ahn,  Sungjin and Kautz,  Jan and Yin,  Hongxu and Lu,  Yao and Han,  Song and Byeon,  Wonmin},
  year = {2025},
  pages = {5889–5900}
}

@inproceedings{liu2024nvila,
  author =        {Liu, Zhijian and Zhu, Ligeng and Shi, Baifeng and
                   Zhang, Zhuoyang and Lou, Yuming and Yang, Shang and
                   Xi, Haocheng and Cao, Shiyi and Gu, Yuxian and
                   Li, Dacheng and Li, Xiuyu and Fang, Yunhao and
                   Chen, Yukang and Hsieh, Cheng-Yu and Huang, De-An and
                   Cheng, An-Chieh and Nath, Vishwesh and Hu, Jinyi and
                   Liu, Sifei and Krishna, Ranjay and Xu, Daguang and
                   Wang, Xiaolong and Molchanov, Pavlo and Kautz, Jan and
                   Yin, Hongxu and Han, Song and Lu, Yao},
  booktitle =     {Proc. CVPR},
  title =         {{NVILA}: Efficient Frontier Visual Language Models},
  year =          {2025},
}

@inproceedings{yang2024pvc,
  author =        {Yang, Chenyu and Dong, Xuan and Zhu, Xizhou and
                   Su, Weijie and Wang, Jiahao and Tian, Hao and
                   Chen, Zhe and Wang, Wenhai and Lu, Lewei and
                   Dai, Jifeng},
  booktitle =     {Proc. CVPR},
  title =         {{PVC}: Progressive Visual Token Compression for
                   Unified Image and Video Processing in Large
                   Vision-Language Models},
  year =          {2025},
}

@inproceedings{liu2024oryx,
  author =        {Liu, Zuyan and Dong, Yuhao and Liu, Ziwei and
                   Hu, Winston and Lu, Jiwen and Rao, Yongming},
  booktitle =     {Proc. ICLR},
  title =         {{Oryx} {MLLM}: On-Demand Spatial-Temporal
                   Understanding at Arbitrary Resolution},
  year =          {2025},
}

@misc{qu2024tsllava,
  author =        {Qu, Tingyu and Li, Mingxiao and Tuytelaars, Tinne and
                   Moens, Marie-Francine},
  title =         {{TS-LLaVA}: Constructing Visual Tokens through
                   Thumbnail-and-Sampling for Training-Free Video Large
                   Language Models},
  year =          {2024},
  howpublished = {arXiv preprint arXiv:2411.11066}
}

@inproceedings{li2025videochatflash,
  author =        {Li, Xinhao and Wang, Yi and Yu, Jiashuo and
                   Zeng, Xiangyu and Zhu, Yuhan and Huang, Haian and
                   Gao, Jianfei and Li, Kunchang and He, Yinan and
                   Wang, Chenting and Qiao, Yu and Wang, Yali and
                   Wang, Limin},
  booktitle =     {Proc. ICLR},
  title =         {{VideoChat-Flash}: Hierarchical Compression for
                   Long-Context Video Modeling},
  year =          {2026},
}

@inproceedings{streamingTOM2025,
  author =        {Xueyi Chen and Keda Tao and Kele Shao and Huan Wang},
  booktitle =     {Proc. CVPR},
  title =         {{StreamingTOM}: Streaming Token Compression for
                   Efficient Video Understanding},
  year =          {2026},
}

@inproceedings{zhang2025flashvstream,
  author =        {Zhang, Haoji and Wang, Yiqin and Tang, Yansong and
                   Liu, Yong and Feng, Jiashi and Jin, Xiaojie},
  booktitle =     {Proc. ICCV},
  title =         {{Flash-VStream}: Efficient Real-Time Understanding
                   for Long Video Streams},
  year =          {2025},
}

@inproceedings{wang2025videollamb,
  author =        {Wang, Yuxuan and Song, Yiqi and Xie, Cihang and
                   Liu, Yang and Zheng, Zilong},
  booktitle =     {Proc. ICCV},
  title =         {{VideoLLaMB}: Long Streaming Video Understanding with
                   Recurrent Memory Bridges},
  year =          {2025},
}

@misc{li2025videoscan,
  author =        {Li, Ruanjun and Tan, Yuedong and Shi, Yuanming and
                   Shao, Jiawei},
  title =         {{VideoScan}: Enabling Efficient Streaming Video
                   Understanding via Frame-level Semantic Carriers},
  year =          {2025},
  howpublished = {arXiv preprint arXiv:2503.09387}
}

@inproceedings{yao2025timechatonline,
  author =        {Yao, Linli and Li, Yicheng and Wei, Yuancheng and
                   Li, Lei and Ren, Shuhuai and Liu, Yuanxin and
                   Ouyang, Kun and Wang, Lean and Li, Shicheng and
                   Li, Sida and Kong, Lingpeng and Liu, Qi and
                   Zhang, Yuanxing and Sun, Xu},
  booktitle =     {Proceedings of the 33rd ACM International Conference
                   on Multimedia (ACM MM)},
  title =         {{TimeChat-Online}: 80\% Visual Tokens are Naturally
                   Redundant in Streaming Videos},
  year =          {2025},
}

@inproceedings{shu2024videoxl,
  author =        {Shu, Yan and Liu, Zheng and Zhang, Peitian and
                   Qin, Minghao and Zhou, Junjie and Liang, Zhengyang and
                   Huang, Tiejun and Zhao, Bo},
  booktitle =     {Proc. CVPR},
  title =         {{Video-XL}: Extra-Long Vision Language Model for
                   Hour-Scale Video Understanding},
  year =          {2025},
}

@misc{chuQwen2AudioTechnicalReport2024,
  author =        {Chu, Yunfei and Xu, Jin and Yang, Qian and
                   Wei, Haojie and Wei, Xipin and Guo, Zhifang and
                   Leng, Yichong and Lv, Yuanjun and He, Jinzheng and
                   Lin, Junyang and Zhou, Chang and Zhou, Jingren},
  title =         {{Qwen2-Audio} Technical Report},
  year =          {2024},
  howpublished = {arXiv preprint arXiv:2407.10759}
}

@article{li2026dash,
  author =        {Bingzhou Li and Tao Huang},
  title =         {{DASH}: Dynamic Audio-Driven Semantic Chunking for
                   Efficient Omnimodal Token Compression},
  year =          {2026},
  journal = {Proc. ECCV},
}

@misc{sunFinegrainedAudioVisualJoint2023,
  author =        {Sun, Guangzhi and Yu, Wenyi and Tang, Changli and
                   Chen, Xianzhao and Tan, Tian and Li, Wei and Lu, Lu and
                   Ma, Zejun and Zhang, Chao},
  title =         {Fine-Grained {{Audio-Visual Joint Representations}}
                   for {{Multimodal Large Language Models}}},
  year =          {2023},
  howpublished = {arXiv preprint arXiv:2310.05863}
}

@inproceedings{sunVideoSALMONNSpeechEnhancedAudioVisual2024,
  author =        {Sun, Guangzhi and Yu, Wenyi and Tang, Changli and
                   Chen, Xianzhao and Tan, Tian and Li, Wei and Lu, Lu and
                   Ma, Zejun and Wang, Yuxuan and Zhang, Chao},
  title =         {Video-{{SALMONN}}: {{Speech-Enhanced Audio-Visual
                   Large Language Models}}},
  year =          {2024},
  booktitle = {Proc. ICML},
  pages = {47198--47217}
}

@misc{hyperclovaxteamHyperCLOVA8BOmni2026,
  author =        {HyperCLOVA X Team and Naver Cloud},
  title =         {{{HyperCLOVA X 8B Omni}}},
  year =          {2026},
  doi =           {10.48550/arXiv.2601.01792},
  howpublished = {arXiv preprint arXiv:2601.01792}
}

@misc{liBaichuanOmniTechnicalReport2024,
  author =        {Li, Yadong and Sun, Haoze and Lin, Mingan and
                   Li, Tianpeng and Dong, Guosheng and Zhang, Tao and
                   Ding, Bowen and Song, Wei and Cheng, Zhenglin and
                   Huo, Yuqi and Chen, Song and Li, Xu and Pan, Da and
                   Zhang, Shusen and Wu, Xin and Liang, Zheng and
                   Liu, Jun and Zhang, Tao and Lu, Keer and Zhao, Yaqi and
                   Shen, Yanjun and Yang, Fan and Yu, Kaicheng and
                   Lin, Tao and Xu, Jianhua and Zhou, Zenan and
                   Chen, Weipeng},
  title =         {Baichuan-{{Omni Technical Report}}},
  year =          {2024},
  doi =           {10.48550/arXiv.2410.08565},
  howpublished = {arXiv preprint arXiv:2410.08565}
}

@inproceedings{chen2024fastv,
  author =        {Chen, Liang and Zhao, Haozhe and Liu, Tianyu and
                   Bai, Shuai and Lin, Junyang and Zhou, Chang and
                   Chang, Baobao},
  booktitle = {Proc. ECCV},
  title =         {An Image is Worth 1/2 Tokens After Layer 2:
                   Plug-and-Play Inference Acceleration for Large
                   Vision-Language Models},
  year =          {2024},
}

@inproceedings{zhang2024sparsevlm,
  author =        {Zhang, Yuan and Fan, Chun-Kai and Ma, Junpeng and
                   Zheng, Wenzhao and Huang, Tao and Cheng, Kuan and
                   Gudovskiy, Denis and Okuno, Tomoyuki and
                   Nakata, Yohei and Keutzer, Kurt and Zhang, Shanghang},
  booktitle =     {Proc. ICML},
  title =         {{SparseVLM}: Visual Token Sparsification for
                   Efficient Vision-Language Model Inference},
  year =          {2025},
}

@inproceedings{fu2025framefusion,
  author =        {Fu, Tianyu and Liu, Tengxuan and Han, Qinghao and
                   Dai, Guohao and Yan, Shengen and Yang, Huazhong and
                   Ning, Xuefei and Wang, Yu},
  booktitle =     {Proc. ICCV},
  title =         {{FrameFusion}: Combining Similarity and Importance
                   for Video Token Reduction on Large Vision Language
                   Models},
  year =          {2025},
}

@article{hieravid2025,
  author =        {Guo, Yansong and Zhu, Chaoyang and Ji, Jiayi and
                   Lin, Jianghang and Cao, Liujuan},
  journal =       {arXiv preprint arXiv:2604.01881},
  title =         {{HieraVid}: Hierarchical Token Pruning for Fast Video
                   Large Language Models},
  year =          {2026},
}

@article{jiang2026stateful,
  author =        {Jindong Jiang and Amala Sanjay Deshmukh and
                   Kateryna Chumachenko and Karan Sapra and Zhiding Yu and
                   Guilin Liu and Andrew Tao and Pavlo Molchanov and
                   Jan Kautz and Wonmin Byeon},
  title =         {Stateful Token Reduction for Long-Video Hybrid
                   {VLMs}},
  year =          {2026},
  journal =       {arXiv preprint arXiv:2603.00198},
}

@inproceedings{zhang2025flexselect,
  author =        {Zhang, Yunzhu and Lu, Yu and Wang, Tianyi and
                   Rao, Fengyun and Yang, Yi and Zhu, Linchao},
  booktitle =     {Proc. NeurIPS},
  title =         {{FlexSelect}: Flexible Token Selection for Efficient
                   Long Video Understanding},
  year =          {2025},
}

@inproceedings{sun-etal-2025-adatp,
  address =       {Suzhou, China},
  author =        {Sun, Fengyuan and Shen, Leqi and Chen, Hui and
                   Zhao, Sicheng and Han, Jungong and Ding, Guiguang},
  booktitle =     {Findings of EMNLP},
  editor =        {Christodoulopoulos, Christos and Chakraborty, Tanmoy and
                   Rose, Carolyn and Peng, Violet},
  pages =         {3273--3286},
  publisher =     {Association for Computational Linguistics},
  title =         {{A}da{TP}: Attention-Debiased Token Pruning for Video
                   Large Language Models},
  year =          {2025},
  doi =           {10.18653/v1/2025.findings-emnlp.174},
  isbn =          {979-8-89176-335-7},
}

@inproceedings{zhang2025vqtoken,
  author =        {Zhang, Haichao and Fu, Yun},
  booktitle =     {Proc. NeurIPS},
  title =         {{VQToken}: Neural Discrete Token Representation
                   Learning for Extreme Token Reduction in Video Large
                   Language Models},
  year =          {2025},
}

@inproceedings{li2025mminference,
  author =        {Li, Yucheng and Jiang, Huiqiang and
                   Zhang, Chengruidong and Wu, Qianhui and Luo, Xufang and
                   Ahn, Surin and Abdi, Amir H. and Li, Dongsheng and
                   Gao, Jianfeng and Yang, Yuqing and Qiu, Lili},
  booktitle =     {Proc. ICML},
  title =         {{MMInference}: Accelerating Pre-filling for
                   Long-Context {VLMs} via Modality-Aware Permutation
                   Sparse Attention},
  year =          {2025},
}

@inproceedings{tao2024dycoke,
  author =        {Tao, Keda and Qin, Can and You, Haoxuan and Sui, Yang and
                   Wang, Huan},
  booktitle =     {Proc. CVPR},
  pages =         {18992--19001},
  title =         {{DyCoke}: Dynamic Compression of Tokens for Fast
                   Video Large Language Models},
  year =          {2025},
}

@misc{wang2024retake,
  author =        {Wang, Xiao and Si, Qingyi and Wu, Jianlong and
                   Zhu, Shiyu and Cao, Li and Nie, Liqiang},
  title =         {{ReTaKe}: Reducing Temporal and Knowledge Redundancy
                   for Long Video Understanding},
  year =          {2024},
  howpublished = {arXiv preprint arXiv:2412.20504}
}

@inproceedings{wan2025meda,
  author =        {Wan, Zhongwei and Shen, Hui and Wang, Xin and
                   Liu, Che and Mai, Zheda and Zhang, Mi},
  booktitle =     {Proceedings of the 2025 Conference of the North
                   American Chapter of the Association for Computational
                   Linguistics (NAACL)},
  title =         {{MEDA}: Dynamic {KV} Cache Allocation for Efficient
                   Multimodal Long-Context Inference},
  year =          {2025},
}

@inproceedings{di2025rekv,
  author =        {Di, Shangzhe and Yu, Zhelun and Zhang, Guanghao and
                   Li, Haoyuan and Zhong, Tao and Cheng, Hao and
                   Li, Bolin and He, Wanggui and Shu, Fangxun and
                   Jiang, Hao},
  booktitle =     {Proc. ICLR},
  title =         {Streaming Video Question-Answering with In-context
                   Video {KV}-Cache Retrieval},
  year =          {2025},
}

@inproceedings{kim2025infinipotv,
  author =        {Kim, Minsoo and Shim, Kyuhong and Choi, Jungwook and
                   Chang, Simyung},
  booktitle =     {Proc. NeurIPS},
  title =         {{InfiniPot-V}: Memory-Constrained {KV} Cache
                   Compression for Streaming Video Understanding},
  year =          {2025},
}

@article{yang2025streammem,
  author =        {Yang, Yanlai and Zhao, Zhuokai and
                   Shukla, Satya Narayan and Singh, Aashu and
                   Mishra, Shlok Kumar and Zhang, Lizhu and Ren, Mengye},
  journal =       {arXiv preprint arXiv:2508.15717},
  title =         {{StreamMem}: Query-Agnostic {KV} Cache Memory for
                   Streaming Video Understanding},
  year =          {2025},
}

@inproceedings{chen2025streamkv,
  author =        {Chen, Yilong and Bai, Xiang and Wang, Zhibin and
                   Bai, Chengyu and Dai, Yuhan and Lu, Ming and
                   Zhang, Shanghang},
  booktitle =     {Proc. AAAI},
  title =         {{StreamKV}: Streaming Video Question-Answering with
                   Segment-based {KV} Cache Retrieval and Compression},
  year =          {2026},
}

@inproceedings{man2025adacm2,
  author =        {Man, Yuanbin and Huang, Ying and Zhang, Chengming and
                   Li, Bingzhe and Niu, Wei and Yin, Miao},
  booktitle =     {Proc. CVPR},
  title =         {{AdaCM$^2$}: On Understanding Extremely Long-Term
                   Video with Adaptive Cross-Modality Memory Reduction},
  year =          {2025},
}

@inproceedings{dehghani2022efficiencymisnomer,
  author =        {Mostafa Dehghani and Anurag Arnab and Lucas Beyer and
                   Ashish Vaswani and Yi Tay},
  title =         {The Efficiency Misnomer},
  year =          {2022},
  booktitle = {Proc. ICLR}
}

@inproceedings{reddiMLPerfInferenceBenchmark2020,
  author =        {Reddi, Vijay Janapa and Cheng, Christine and
                   Kanter, David and Mattson, Peter and
                   Schmuelling, Guenther and Wu, Carole-Jean and
                   Anderson, Brian and Breughe, Maximilien and
                   Charlebois, Mark and Chou, William and Chukka, Ramesh and
                   Coleman, Cody and Davis, Sam and Deng, Pan and
                   Diamos, Greg and Duke, Jared and Fick, Dave and
                   Gardner, J. Scott and Hubara, Itay and
                   Idgunji, Sachin and Jablin, Thomas B. and Jiao, Jeff and
                   John, Tom St and Kanwar, Pankaj and Lee, David and
                   Liao, Jeffery and Lokhmotov, Anton and
                   Massa, Francisco and Meng, Peng and
                   Micikevicius, Paulius and Osborne, Colin and
                   Pekhimenko, Gennady and Rajan, Arun Tejusve Raghunath and
                   Sequeira, Dilip and Sirasao, Ashish and Sun, Fei and
                   Tang, Hanlin and Thomson, Michael and Wei, Frank and
                   Wu, Ephrem and Xu, Lingjie and Yamada, Koichi and
                   Yu, Bing and Yuan, George and Zhong, Aaron and
                   Zhang, Peizhao and Zhou, Yuchen},
  title =         {{{MLPerf Inference Benchmark}}},
  year =          {2020},
  booktitle = {Proc. ISCA}
}

@inproceedings{tschandMLPerfPowerBenchmarking2024,
  author =        {Tschand, Arya and Rajan, Arun Tejusve Raghunath and
                   Idgunji, Sachin and Ghosh, Anirban and
                   Holleman, Jeremy and Kiraly, Csaba and
                   Ambalkar, Pawan and Borkar, Ritika and Chukka, Ramesh and
                   Cockrell, Trevor and Curtis, Oliver and
                   Fursin, Grigori and Hodak, Miro and Kassa, Hiwot and
                   Lokhmotov, Anton and Miskovic, Dejan and Pan, Yuechao and
                   Manmathan, Manu Prasad and Raymond, Liz and
                   John, Tom St and Suresh, Arjun and Taubitz, Rowan and
                   Zhan, Sean and Wasson, Scott and Kanter, David and
                   Reddi, Vijay Janapa},
  title =         {{{MLPerf Power}}: {{Benchmarking}} the {{Energy
                   Efficiency}} of {{Machine Learning Systems}} from
                   {{Microwatts}} to {{Megawatts}} for {{Sustainable
                   AI}}},
  year =          {2025},
  doi =           {10.48550/arXiv.2410.12032},
  booktitle = {2025 IEEE International Symposium on High Performance Computer Architecture (HPCA)}
}

@article{liuTrainingfreeLongVideo2025,
  author =        {Liu, Jingren and Wang, Yun and Zhang, Long and
                   Wang, Yiheng and Xu, Shuning and Wang, Ling and
                   Yan, Jiaqi and Zhang, Dell and Chen, Xiangyu},
  journal =       {Vicinagearth},
  number =        {1},
  pages =         {6},
  title =         {Towards Training-Free Long Video Understanding:
                   Methods, Benchmarks, and Open Challenges},
  volume =        {2},
  year =          {2025},
  doi =           {10.1007/s44336-025-00017-w},
  issn =          {3005-060X},
}

@misc{brkic2025framesamplingstrategiesmatter,
  author =        {Brkic, Marija and Razzouki, Anas Filali and
                   Tevissen, Yannis and Guetari, Khalil and
                   Yacoubi, Mounim A. El},
  title =         {Frame {{Sampling Strategies Matter}}: {{A Benchmark}}
                   for Small Vision Language Models},
  year =          {2025},
  doi =           {10.48550/arXiv.2509.14769},
  howpublished = {arXiv preprint arXiv:2509.14769}
}

@inproceedings{leiLessMoreClipBERT2021,
  author =        {Lei, Jie and Li, Linjie and Zhou, Luowei and Gan, Zhe and
                   Berg, Tamara L. and Bansal, Mohit and Liu, Jingjing},
  title =         {Less Is {{More}}: {{ClipBERT}} for
                   {{Video-and-Language Learning}} via {{Sparse
                   Sampling}}},
  year =          {2021},
  doi =           {10.48550/arXiv.2102.06183},
  booktitle = {Proc. CVPR}
}

@inproceedings{yeungEndtoendLearningAction2017,
  author =        {Yeung, Serena and Russakovsky, Olga and Mori, Greg and
                   {Fei-Fei}, Li},
  title =         {End-to-End {{Learning}} of {{Action Detection}} from
                   {{Frame Glimpses}} in {{Videos}}},
  year =          {2016},
  doi =           {10.48550/arXiv.1511.06984},
  booktitle = {Proc. CVPR}
}

@inproceedings{fanWatchingSmallPortion2018,
  author =        {Fan, Hehe and Xu, Zhongwen and Zhu, Linchao and
                   Yan, Chenggang and Ge, Jianjun and Yang, Yi},
  title =         {Watching a {{Small Portion}} Could Be as {{Good}} as
                   {{Watching All}}: {{Towards Efficient Video
                   Classification}}},
  year =          {2018},
  booktitle = {Proc. IJCAI}
}

@inproceedings{wuMultiAgentReinforcementLearning2019,
  author =        {Wu, Wenhao and He, Dongliang and Tan, Xiao and
                   Chen, Shifeng and Wen, Shilei},
  title =         {Multi-{{Agent Reinforcement Learning Based Frame
                   Sampling}} for {{Effective Untrimmed Video
                   Recognition}}},
  year =          {2019},
  booktitle = {Proc. ICCV}
}

@inproceedings{wuLiteEvalCoarsetoFineFramework2019,
  author =        {Wu, Zuxuan and Xiong, Caiming and Jiang, Yu-Gang and
                   Davis, Larry S.},
  title =         {{{LiteEval}}: {{A Coarse-to-Fine Framework}} for
                   {{Resource Efficient Video Recognition}}},
  year =          {2019},
  booktitle = {Proc. NeurIPS}
}

@inproceedings{mengARNetAdaptiveFrame2020,
  author =        {Meng, Yue and Lin, Chung-Ching and Panda, Rameswar and
                   Sattigeri, Prasanna and Karlinsky, Leonid and
                   Oliva, Aude and Saenko, Kate and Feris, Rogerio},
  title =         {{{AR-Net}}: {{Adaptive Frame Resolution}} for
                   {{Efficient Action Recognition}}},
  year =          {2020},
  booktitle = {Proc. ECCV}
}

@misc{liangKeyVideoLLMLargescaleVideo2024,
  author =        {Liang, Hao and Li, Jiapeng and Bai, Tianyi and
                   Chen, Chong and He, Conghui and Cui, Bin and
                   Zhang, Wentao},
  title =         {{{KeyVideoLLM}}: {{Towards Large-scale Video Keyframe
                   Selection}}},
  year =          {2024},
  howpublished = {arXiv preprint arXiv:2407.03104}
}

@inproceedings{sunMDP3TrainingfreeApproach2025,
  author =        {Sun, Hui and Lu, Shiyin and Wang, Huanyu and
                   Chen, Qing-Guo and Xu, Zhao and Luo, Weihua and
                   Zhang, Kaifu and Li, Ming},
  title =         {{{MDP3}}: {{A Training-free Approach}} for
                   {{List-wise Frame Selection}} in {{Video-LLMs}}},
  year =          {2025},
  doi =           {10.48550/arXiv.2501.02885},
  booktitle = {Proc. ICCV}
}

@inproceedings{guoLogicinFramesDynamicKeyframe2025,
  author =        {Guo, Weiyu and Chen, Ziyang and Wang, Shaoguang and
                   He, Jianxiang and Xu, Yijie and Ye, Jinhui and
                   Sun, Ying and Xiong, Hui},
  title =         {Logic-in-{{Frames}}: {{Dynamic Keyframe Search}} via
                   {{Visual Semantic-Logical Verification}} for {{Long
                   Video Understanding}}},
  year =          {2025},
  doi =           {10.48550/arXiv.2503.13139},
  booktitle = {Proc. NeurIPS}
}

@misc{chasmaiMomentSamplingVideo2025,
  author =        {Chasmai, Mustafa and Jagatap, Gauri and KV, Gouthaman and
                   Horn, Grant Van and Maji, Subhransu and
                   Fanelli, Andrea},
  title =         {Moment {{Sampling}} in {{Video LLMs}} for {{Long-Form
                   Video QA}}},
  year =          {2025},
  doi =           {10.48550/arXiv.2507.00033},
  howpublished = {arXiv preprint arXiv:2507.00033}
}

@inproceedings{janggumbelsoftmax2016,
  author =        {Jang, Eric and Gu, Shixiang and Poole, Ben},
  title =         {Categorical Reparameterization with Gumbel-Softmax},
  year =          {2017},
  booktitle = {Proc. ICLR}
}

@inproceedings{huMLLMBasedVideo2025,
  author =        {Hu, Kai and Gao, Feng and Nie, Xiaohan and Zhou, Peng and
                   Tran, Son and Neiman, Tal and Wang, Lingyun and
                   Shah, Mubarak and Hamid, Raffay and Yin, Bing and
                   Chilimbi, Trishul},
  title =         {M-{{LLM Based Video Frame Selection}} for {{Efficient
                   Video Understanding}}},
  year =          {2025},
  doi =           {10.48550/arXiv.2502.19680},
  booktitle = {Proc. CVPR}
}

@misc{yaoKframesSceneDrivenAnyk2025,
  author =        {Yao, Yifeng and Yun, Yike and Wang, Jing and
                   Zhang, Huishuai and Zhao, Dongyan and Tian, Ke and
                   Wang, Zhihao and Qiu, Minghui and Wang, Tao},
  title =         {K-Frames: {{Scene-Driven Any-k Keyframe Selection}}
                   for Long Video Understanding},
  year =          {2025},
  doi =           {10.48550/arXiv.2510.13891},
  howpublished = {arXiv preprint arXiv:2510.13891}
}

@misc{liFrameOracleLearningWhat2025,
  author =        {Li, Chaoyu and Li, Tianzhi and Tao, Fei and
                   Zhao, Zhenyu and Wu, Ziqian and Zhao, Maozheng and
                   Song, Juntong and Niu, Cheng and Fazli, Pooyan},
  title =         {{{FrameOracle}}: {{Learning What}} to {{See}} and
                   {{How Much}} to {{See}} in {{Videos}}},
  year =          {2025},
  doi =           {10.48550/arXiv.2510.03584},
  howpublished = {arXiv preprint arXiv:2510.03584}
}

@inproceedings{liKFSBenchComprehensiveEvaluation2025,
  author =        {Li, Zongyao and Ishida, Kengo and Yamazaki, Satoshi and
                   Ji, Xiaotong and Liu, Jianquan},
  title =         {{{KFS-Bench}}: {{Comprehensive Evaluation}} of {{Key
                   Frame Sampling}} in {{Long Video Understanding}}},
  year =          {2026},
  doi =           {10.48550/arXiv.2512.14017},
  booktitle = {Proc. WACV}
}

@inproceedings{guMambaLinearTimeSequence2023,
  author =        {Gu, Albert and Dao, Tri},
  title =         {Mamba: Linear-Time Sequence Modeling with Selective
                   State Spaces},
  year =          {2024},
  booktitle = {Proc. COLM}
}

@article{li2024llavaonevision,
  author =        {Li, Bo and Zhang, Yuanhan and Guo, Dong and
                   Zhang, Renrui and Li, Feng and Zhang, Hao and
                   Zhang, Kaichen and Li, Yanwei and Liu, Ziwei and
                   Li, Chunyuan},
  title =         {LLaVA-OneVision: Easy Visual Task Transfer},
  year =          {2025},
  journal = {Transactions on Machine Learning Research}
}

@inproceedings{kumar2025lgttp,
  author =        {Kumar, Yogesh},
  booktitle =     {Proc. EMNLP},
  title =         {Language-Guided Temporal Token Pruning for Efficient
                   {VideoLLM} Processing},
  year =          {2025},
}

@misc{zhang2025dyntok,
  author =        {Zhang, Hongzhi and Zhang, Jingyuan and Ji, Xingguang and
                   Wang, Qi and Zhang, Fuzheng},
  title =         {{DynTok}: Dynamic Compression of Visual Tokens for
                   Efficient and Effective Video Understanding},
  year =          {2025},
  howpublished = {arXiv preprint arXiv:2506.03990}
}

@misc{liu2025videoxlpro,
  author =        {Liu, Xiangrui and Shu, Yan and Liu, Zheng and Li, Ao and
                   Tian, Yang and Zhao, Bo},
  title =         {{Video-XL-Pro}: Reconstructive Token Compression for
                   Extremely Long Video Understanding},
  year =          {2025},
  howpublished = {arXiv preprint arXiv:2503.18478}
}

@misc{qin2025videoxl2,
  author =        {Qin, Minghao and Liu, Xiangrui and Liang, Zhengyang and
                   Shu, Yan and Yuan, Huaying and Zhou, Juenjie and
                   Xiao, Shitao and Zhao, Bo and Liu, Zheng},
  title =         {{Video-XL-2}: Towards Very Long-Video Understanding
                   Through Task-Aware {KV} Sparsification},
  year =          {2025},
  howpublished = {arXiv preprint arXiv:2506.19225}
}

@inproceedings{lieberJambaHybridTransformerMamba2024,
  author =        {Lenz, Barak and Lieber, Opher and Bata, Hofit and Cohen, Gal and Osin, Jhonathan and Dalmedigos, Itay and Safahi, Erez and Meirom, Shaked and Belinkov, Yonatan and Shalev-Shwartz, Shai and Abend, Omri and Alon, Raz and Asida, Tomer and Bergman, Amir and Glozman, Roman and Gokhman, Michael and Manevich, Avashalom and Ratner, Nir and Rozen, Noam and Shwartz, Erez and Zusman, Mor and Shoham, Yoav},
  title =         {Jamba: A Hybrid Transformer-Mamba Language Model},
  year =          {2025},
  booktitle = {Proc. ICLR}
}

@article{xu2025timeviper,
  author =        {Boshen Xu and Zihan Xiao and Jiaze Li and
                   Jianzhong Ju and Zhenbo Luo and Jian Luan and
                   Qin Jin},
  title =         {{TimeViper}: A Hybrid Mamba-Transformer
                   Vision-Language Model for Efficient Long Video
                   Understanding},
  year =          {2026},
  journal = {Proc. CVPR},
}

@misc{bai2025qwen25vltechnicalreport,
  author =        {Shuai Bai and Keqin Chen and Xuejing Liu and
                   Jialin Wang and Wenbin Ge and Sibo Song and Kai Dang and
                   Peng Wang and Shijie Wang and Jun Tang and
                   Humen Zhong and Yuanzhi Zhu and Mingkun Yang and
                   Zhaohai Li and Jianqiang Wan and Pengfei Wang and
                   Wei Ding and Zheren Fu and Yiheng Xu and Jiabo Ye and
                   Xi Zhang and Tianbao Xie and Zesen Cheng and
                   Hang Zhang and Zhibo Yang and Haiyang Xu and
                   Junyang Lin},
  title =         {Qwen2.5-VL Technical Report},
  year =          {2025},
  howpublished = {arXiv preprint arXiv:2502.13923}
}

@article{wang2025adaretake,
  author =        {Wang, Xiao and Si, Qingyi and Wu, Jianlong and
                   Zhu, Shiyu and Cao, Li and Nie, Liqiang},
  journal =       {arXiv preprint arXiv:2503.12559},
  title =         {{AdaReTaKe}: Adaptive Redundancy Reduction to
                   Perceive Longer for Video-language Understanding},
  year =          {2025},
}

@inproceedings{linBoostingMultimodalVisualTokensWithdrawal2025,
  author =        {Lin, Zhihang and Lin, Mingbao and Lin, Luxi and
                   Ji, Rongrong},
  booktitle =     {Proc. AAAI},
  title =         {Boosting Multimodal Large Language Models with Visual
                   Tokens Withdrawal for Rapid Inference},
  year =          {2025},
}

@inproceedings{xingPyramidDropAcceleratingLarge2025,
  author =        {Xing, Long and Huang, Qidong and Dong, Xiaoyi and
                   Lu, Jiajie and Zhang, Pan and Zang, Yuhang and
                   Cao, Yuhang and He, Conghui and Wang, Jiaqi and
                   Wu, Feng and Lin, Dahua},
  booktitle =     {Proc. CVPR},
  title =         {{PyramidDrop}: Accelerating Your Large
                   Vision-Language Models via Pyramid Visual Redundancy
                   Reduction},
  year =          {2025},
}

@inproceedings{wanLOOKMLookOnceOptimization2024,
  author =        {Wan, Zhongwei and Wu, Ziang and Liu, Che and
                   Huang, Jinfa and Zhu, Zhihong and Jin, Peng and
                   Wang, Longyue and Yuan, Li},
  booktitle =     {Findings of EMNLP},
  pages =         {4065--4078},
  title =         {{LOOK-M}: Look-Once Optimization in {KV} Cache for
                   Efficient Multimodal Long-Context Inference},
  year =          {2024},
  doi =           {10.18653/v1/2024.findings-emnlp.235},
}

@article{monfortMomentsTimeDataset2019,
  author =        {Monfort, Mathew and Andonian, Alex and Zhou, Bolei and
                   Ramakrishnan, Kandan and Bargal, Sarah Adel and
                   Yan, Tom and Brown, Lisa and Fan, Quanfu and
                   Gutfruend, Dan and Vondrick, Carl and Oliva, Aude},
  title =         {Moments in {{Time Dataset}}: One Million Videos for
                   Event Understanding},
  year =          {2020},
  doi =           {10.48550/arXiv.1801.03150},
  journal = {IEEE Transactions on Pattern Analysis and Machine Intelligence}
}

@article{soomroUCF101Dataset101,
  author =        {Soomro, Khurram and Zamir, Amir Roshan and
                   Shah, Mubarak},
  title =         {{{UCF101}}: {{A Dataset}} of 101 {{Human Actions
                   Classes From Videos}} in {{The Wild}}},
  year =          {2012},
  journal =       {arXiv preprint arXiv:1212.0402},
}

@inproceedings{zhang-etal-2025-lmms,
  address =       {Albuquerque, New Mexico},
  author =        {Zhang, Kaichen and Li, Bo and Zhang, Peiyuan and
                   Pu, Fanyi and Cahyono, Joshua Adrian and Hu, Kairui and
                   Liu, Shuai and Zhang, Yuanhan and Yang, Jingkang and
                   Li, Chunyuan and Liu, Ziwei},
  booktitle =     {Findings of NAACL},
  editor =        {Chiruzzo, Luis and Ritter, Alan and Wang, Lu},
  pages =         {881--916},
  publisher =     {Association for Computational Linguistics},
  title =         {{LMM}s-Eval: Reality Check on the Evaluation of Large
                   Multimodal Models},
  year =          {2025},
  doi =           {10.18653/v1/2025.findings-naacl.51},
  isbn =          {979-8-89176-195-7},
}

@inproceedings{kuehne2011hmdb,
  author =        {Kuehne, Hildegard and Jhuang, Hueihan and
                   Garrote, Est{\'i}baliz and Poggio, Tomaso and
                   Serre, Thomas},
  booktitle =     {Proc. ICCV},
  pages =         {2556--2563},
  publisher =     {IEEE},
  title =         {{HMDB}: A Large Video Database for Human Motion
                   Recognition},
  year =          {2011},
}

@inproceedings{gemmeke2017audioset,
  author =        {Gemmeke, Jort F. and Ellis, Daniel P. W. and
                   Freedman, Dylan and Jansen, Aren and
                   Lawrence, Wade and Moore, R. Channing and
                   Plakal, Manoj and Ritter, Marvin},
  booktitle =     {2017 IEEE International Conference on Acoustics,
                   Speech and Signal Processing (ICASSP)},
  pages =         {776--780},
  publisher =     {IEEE},
  title =         {Audio {Set}: An Ontology and Human-Labeled Dataset
                   for Audio Events},
  year =          {2017},
}

@inproceedings{zhang2026efficientinference,
  author = {Zhang, Jun and Ji, Yicheng and Ren, Feiyang and Li, Yihang and Zeng, Bowen and Chen, Zonghao and Chen, Ke and Shou, Lidan and Chen, Gang and Li, Huan},
  title = {Efficient Inference for Large Vision-Language Models: Bottlenecks, Techniques, and Prospects},
  booktitle = {Findings of ACL},
  year = {2026},
  pages = {21036--21066},
  publisher = {Association for Computational Linguistics},
  doi = {10.18653/v1/2026.findings-acl.1057},
  url = {https://aclanthology.org/2026.findings-acl.1057/}
}

@misc{wu2026compressionlifecycle,
  author = {Wu, Hao and Tong, Junlong and Wang, Xudong and Tan, Yang and Zeng, Changyu and Antsiferova, Anastasia and Shen, Xiaoyu},
  title = {From Data to Model: A Survey of the Compression Lifecycle in {MLLMs}},
  year = {2026},
  howpublished = {TechRxiv preprint},
  doi = {10.36227/techrxiv.177220375.55495124/v1},
  url = {https://doi.org/10.36227/techrxiv.177220375.55495124/v1}
}

@inproceedings{zhaiSigmoidLossLanguage2023,
  author =        {Zhai, Xiaohua and Mustafa, Basil and Kolesnikov, Alexander and Beyer, Lucas},
  title =         {Sigmoid {{Loss}} for {{Language Image Pre-Training}}},
  booktitle =     {Proc. ICCV},
  year =          {2023},
}

\begin{IEEEbiography}[{\includegraphics[width=1in,height=1.25in,clip,keepaspectratio]{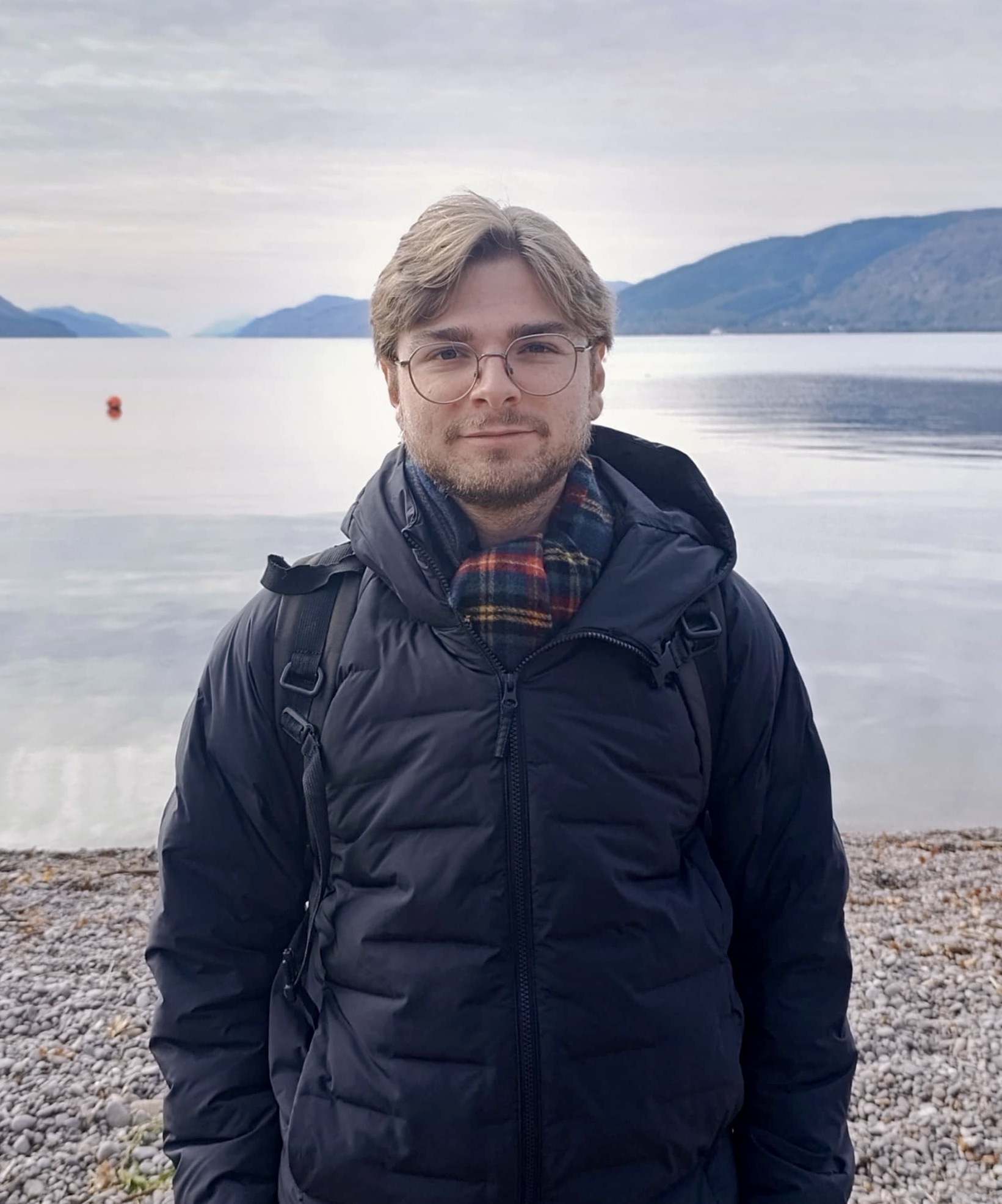}}]{Killian Steunou}
received the master's degree in mathematics, vision, and learning (MVA) from ENS Paris-Saclay, in 2025. He is currently working toward the PhD degree with the SAMOVAR Laboratory, T\'el\'ecom SudParis, Institut Polytechnique de Paris, Palaiseau, France, and is a research scientist with Moments Lab, Paris, France. His current research interests include efficient video--language models, temporal frame selection, and video understanding.
\end{IEEEbiography}

\begin{IEEEbiography}[{\includegraphics[width=1in,height=1.25in,clip,keepaspectratio]{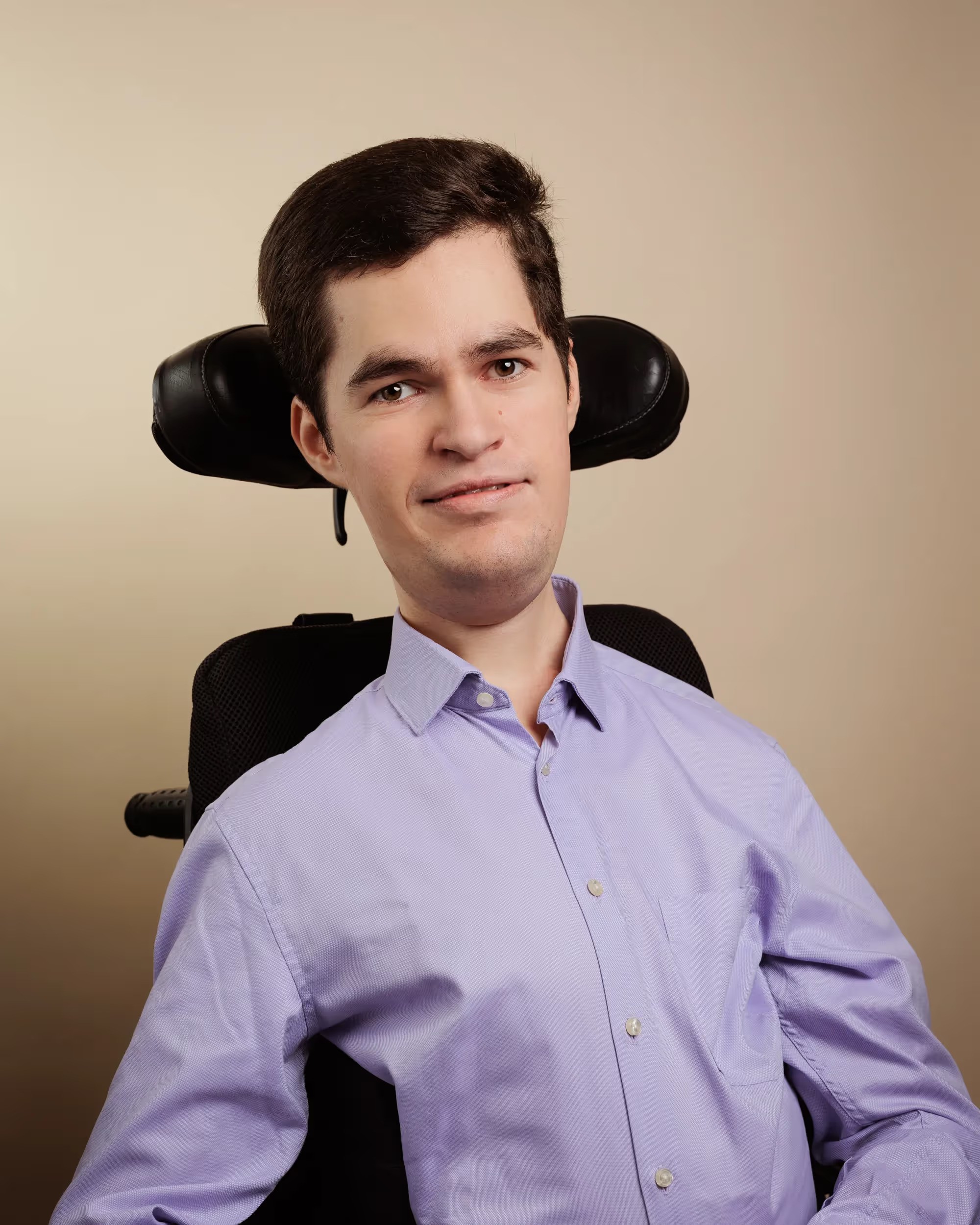}}]{Yannis Tevissen} (Member, IEEE)
received the PhD degree for his work in multimodal speaker diarization at Institut Polytechnique de Paris, in 2023, advised by J\'er\^ome Boudy and G\'erard Chollet. He is currently the head of research at Moments Lab, Paris, France. His current research interests include video understanding, multimodal retrieval, long-context efficiency, and fairness in vision--language models.
\end{IEEEbiography}

\begin{IEEEbiography}[{\includegraphics[width=1in,height=1.25in,clip,keepaspectratio]{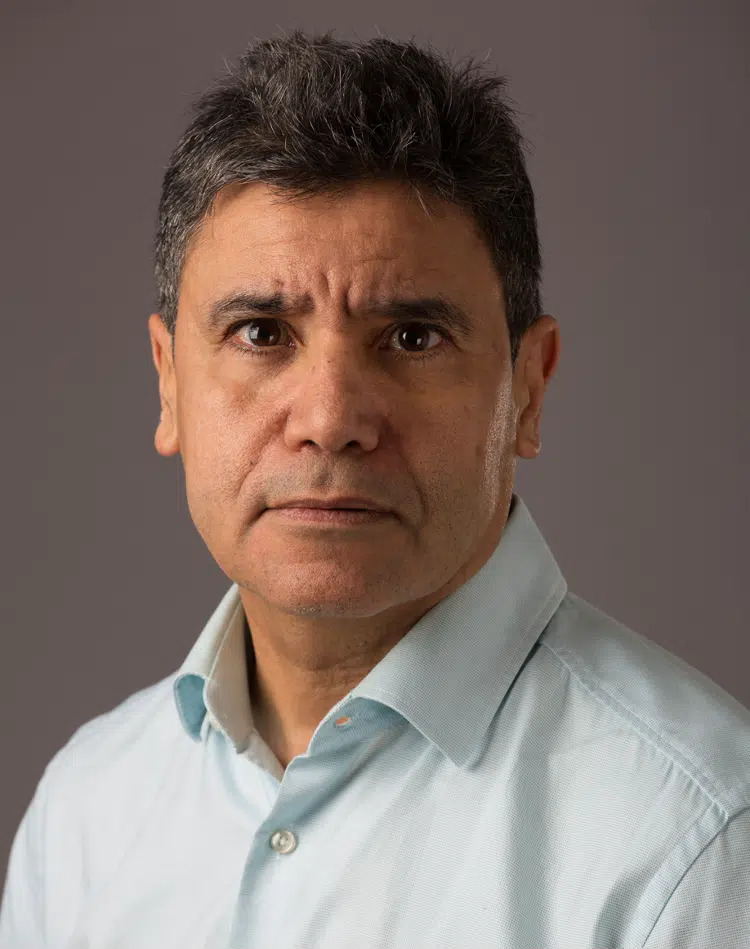}}]{Moun\^im A. El-Yacoubi}
(Senior Member, IEEE) received the PhD degree from the University of Rennes, France, in 1996, and the Habilitation à Diriger des Recherches from Université Paris-Saclay, in 2014. After research positions with La Poste (SRTP), France, CENPARMI, Concordia University, Canada, the Pontifical Catholic University of Paraná, Brazil, and Parascript, USA, he is currently a professor with the SAMOVAR Laboratory, Télécom SudParis, Institut Polytechnique de Paris, Palaiseau, France. His research interests include machine learning, deep learning, and pattern recognition, with an emphasis on modeling human behavioral data such as handwriting, gesture, and activity, and on e-health applications.
\end{IEEEbiography}

\vfill

\end{document}